\documentclass{article}

\usepackage{arxiv}

\usepackage[utf8]{inputenc} 
\usepackage[T1]{fontenc}    
\usepackage{hyperref}       
\usepackage{url}            
\usepackage{booktabs}       
\usepackage{amsfonts}       
\usepackage{nicefrac}       
\usepackage{microtype}      
\usepackage{lipsum}
\usepackage{fancyhdr}       
\usepackage{graphicx}       
\usepackage{subfigure}
\usepackage{xcolor}
\graphicspath{{media/}}     
\usepackage{textcomp}  
\usepackage{scalerel}  
\usepackage{framed} 
\usepackage{enumitem}

\usepackage{array,multirow,graphicx}
\usepackage{float}
\usepackage{mdframed}
\usepackage[T1]{fontenc}

\newcommand\icon{\raisebox{-3.7pt}{\includegraphics[width=1em]{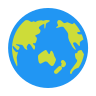}}}

\def\geolm{\textsc{GeoGalactica}}

\title{Ge\icon Galactica: A Scientific Large Language Model \\ in Geoscience}

\author{Zhouhan Lin$^{1,*}$, Cheng Deng$^1$, Le Zhou$^1$, Tianhang Zhang$^1$, Yi Xu$^1$, Yutong Xu$^1$, \\ 
\textbf{Zhongmou He$^{1,2}$, Yuanyuan Shi$^1$, Beiya Dai$^1$, Yunchong Song$^1$, Boyi Zeng$^1$, Qiyuan Chen$^1$, }\\ 
\textbf{Yuxun Miao$^1$, Bo Xue$^1$, Shu Wang$^3$, Luoyi Fu$^1$, Weinan Zhang$^1$, Junxian He$^4$,}\\
\textbf{Yunqiang Zhu$^3$, Xinbing Wang$^1$, Chenghu Zhou$^{1,3}$}
\\
$^1$Shanghai Jiao Tong University \quad $^2$University of Michigan \\
$^3$Institute of Geographical Science and Natural Resources Research, CAS \\
$^4$The Hong Kong University of Science and Technology \\
\texttt{lin.zhouhan@gmail.com, davendw@sjtu.edu.cn, junxianh@cse.ust.hk, zhouch@lreis.ac.cn} \\
}

\begin{document}
\begin{sloppypar}
\maketitle

\renewcommand{\thefootnote}{\fnsymbol{footnote}}
    \footnotetext[1]{Zhouhan Lin is the corresponding author (\href{mailto:lin.zhouhan@gmail.com}{lin.zhouhan@gmail.com}).}
    \footnotetext[2]{Version: v2 (major update on Apirl 10, 2024).}
    \footnotetext[3]{For detailed author contributions, please refer to Appendix~\ref{contribution}.}
    \footnotetext[4]{For all the checkpoints during the $3/4$ pre-training can be accessed on \href{https://huggingface.co/geobrain-ai/geogalactica-ckpt}{geobrain-ai/geogalactica-ckpt}. One can apply for the download links for further research and investigation, we will have a strict verification on the usage of the checkpoints for responsible AI principles.}
    
\renewcommand{\thefootnote}{\arabic{footnote}}

\begin{abstract}
Large language models (LLMs) have achieved huge success for their general knowledge and ability to solve a wide spectrum of tasks in natural language processing (NLP). Due to their impressive abilities, LLMs have shed light on potential inter-discipline applications to foster scientific discoveries of a specific domain by using artificial intelligence (AI for science, AI4S). 
In the meantime, utilizing NLP techniques in geoscience research and practice is wide and convoluted, contributing from knowledge extraction and document classification to question answering and knowledge discovery. 
In this work, we take the initial step to leverage LLM for science, through a rather straightforward approach. We try to specialize an open-sourced LLM into geoscience, by further pre-training the model with a vast amount of texts in geoscience, as well as supervised fine-tuning (SFT) the resulting model with our custom collected instruction tuning dataset.
These efforts result in a model \textbf{\geolm{}} consisting of \textbf{30 billion parameters}. To our best knowledge, it is the largest language model for the geoscience domain.
More specifically, \geolm{} is from further pre-training of Galactica -- a top-performing LLM trained with a large number of scientific documents. 
We train \geolm{} over a geoscience-related scientific text corpus containing 65 billion tokens, preserving as the largest geoscience-specific text corpus. Then we fine-tune the model with 1 million pairs of instruction-tuning data consisting of questions that demand professional geoscience knowledge to answer.
We validate \geolm{} on various geoscience examinations and geoscience-related open-domain questions evaluated by a group of senior geoscientists.
\geolm{} demonstrates the state-of-the-art performance in a diverse range of NLP tasks in geoscience, as well as revealing the potential of using geoscience-related tools.
In this technical report, we will illustrate in detail all aspects of \geolm{}, including data collection, data cleaning, base model selection, pre-training, SFT, and evaluation. 
We open-source our data curation tools and the checkpoints of \geolm{} during the first $3/4$ of pre-training in \url{https://github.com/geobrain-ai/geogalactica}$^{\S}$.

\end{abstract}
\newpage

\tableofcontents

\begin{figure}[h]
    \centering
    \includegraphics[width=\linewidth]{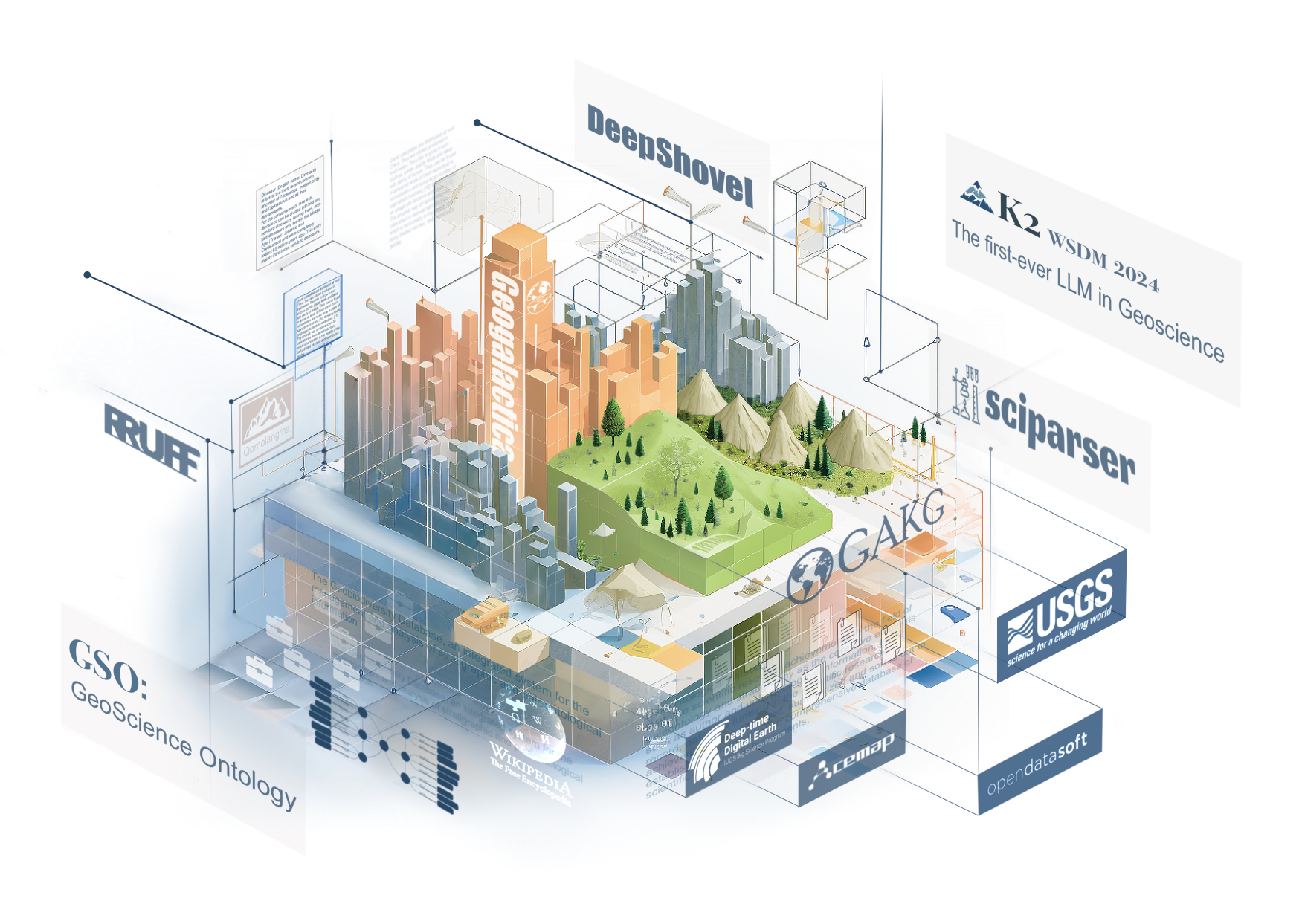}
    \caption{The overview of the processing, construction, components, and applications of \geolm{}.}
    \label{fig:overview}
\end{figure}

\newpage

\section{Introduction}
The rapid advancement of Large Language Models (LLMs) has ushered in a transformative era in natural language processing (NLP), where these models have exhibited remarkable capabilities across a wide spectrum of tasks and domains. These advanced AI models have demonstrated their prowess in handling diverse natural language tasks, including reading comprehension, open-ended question answering, code generation, etc. Their ability to harness vast amounts of general knowledge and apply it to solve specific challenges has sparked interest in exploring their potential applications in various scientific disciplines. In this context, the intersection of artificial intelligence (AI) and science, often referred to as AI for Science (AI4S), has emerged as a promising frontier for driving scientific discoveries and innovations.

Within the realm of AI4S, one particularly intriguing avenue is the integration of NLP techniques into geoscience research and practice. Geoscience is a comprehensive discipline encompassing fields such as geophysics, geology, meteorology, environmental science, etc., with a primary focus on unraveling the complexities of natural processes and phenomena on Earth. Traditionally, geoscientists have relied on theoretical and empirical approaches to advance their understanding of the Earth's systems. However, the sheer volume of data generated in contemporary geoscience research necessitates new strategies and tools for knowledge discovery. The integration of computer science methodologies and AI technologies into geoscience has thus emerged as a transformative paradigm, offering the potential to accelerate scientific progress and address pressing global challenges effectively. In an era characterized by global challenges such as climate change and natural disaster mitigation, the need for efficient data acquisition, information sharing, and knowledge dissemination in geoscience has never been more critical.

In the field of geoscience, domain-specific geoscientific knowledge is usually presented in various forms of text data, such as scientific literature, textbooks, patents, industry standards, etc., which traditionally require the utilization of knowledge systems~\cite{Wang2022AUF}, knowledge graphs\cite{Deng2021GAKGAM}, or semantic models~\cite{Ramachandran2022LanguageMF} to extract a structured form of these knowledge. More broadly, applying NLP techniques for geoscience use cases has been widely accepted~\cite{zhang2023geoscience}, ranging from less complex tasks such as document classification~\cite{qiu2019geoscience}, topic modeling~\cite{lawley2023applications}, and entity recognition\cite{qiu2020automatic,qiu2018dgeosegmenter}, to more complex tasks such as knowledge graph construction~\cite{wang2018information}, question answering~\cite{deng2023pk} and summarization~\cite{ma2022article}. 

While general domain LLMs like Galactica~\cite{Taylor2022GalacticaAL}, LLaMA~\cite{Touvron2023LLaMAOA}, and GLM~\cite{Zeng2022GLM130BAO} have achieved impressive performance across various NLP tasks, they lack the domain-specific knowledge required for geoscience applications. These models have been trained on general datasets that lack authoritative geoscience-related data, limiting their adequacy in addressing the unique challenges posed by the geoscience domain. 
Although our recent attempt to adapt the LLaMA-7B model for geoscience using geoscience-specific data, i.e. the K2\cite{Deng2023LearningAF} model, has shown promising results, this primitive attempt is constrained by its model size and data scale, which consequently may not fully capture the complexity of geoscientific terminology and concepts. However, training a larger LLM comes with new technical challenges, since many aspects of the process become fundamentally different as the model scales up. For example, the stability of training will become more vulnerable, and the training data needs to be scaled up accordingly, resulting in a more systematic way of managing different data sources, etc. 

Therefore, tailoring a general, larger LLM for the scientific domain of geoscience with a more systematically designed dataset and training pipeline is imperative in this era of LLMs. 
In response to these necessities, this work presents a significant step forward in the development of the model as well as the set of toolchains around it. 

Leveraging the vast amount of resources of scientific literature's meta-data, particularly the data resources collected for the OpenAlex~\footnote{\url{https://openalex.org/}}, Web of Science~\footnote{\url{https://www.webofscience.com/wos/}}, Semantic Scholar~\footnote{\url{https://www.semanticscholar.org/}}, and Acemap~\footnote{\url{https://www.acemap.info}}, we can create, organize, and manage a large and comprehensive geoscience dataset targeted for all stages in large language model training. In particular, we have introduced GAKG~\cite{Deng2021GAKGAM}, Deep Literature\footnote{https://idea.acemap.info/}, GSO\footnote{https://gso.acemap.info/}, and other platforms as carriers and repositories of geoscience text knowledge. These concerted efforts have not only allowed us to accumulate a comprehensive geoscience data archive but also have served as foundations for constructing an extensive instruction-tuning dataset for geoscience-related questions, \textbf{GeoSignal-v2}, which has been employed in supervised fine-tuning (SFT). In addition, we have developed and customized a series of data-cleaning tools that allow us to automatically convert various forms of raw data, such as PDF files, forms, equations, knowledge graphs, etc., into clean texts suited as training corpus for large language models. To our best knowledge, our collected corpus has become the largest geoscience dataset.

we have then successfully further pre-trained a language model with \textbf{30B} parameters, with Galactica-30B~\cite{Taylor2022GalacticaAL} as its base model. The resulting model is thus named as \geolm{}, empowering various academic tasks in the geoscience field. With its 30 billion parameters, this model represents the culmination of further pre-training and supervised fine-tuning, making it the largest language model dedicated to the geoscience domain. Our experimental findings demonstrate that, compared to models of equivalent scale, \geolm{} exhibits exceptional performance on GeoBenchmark~\cite{Deng2023LearningAF}. Regarding human evaluation, our model showcases impressive competence in geoscience-related tasks when compared with 5 general language models (\textit{ChatGPT\footnote{\url{https://chat.openai.com/}}, Yiyan\footnote{\url{https://yiyan.baidu.com/}}, Qianwen\footnote{\url{https://qianwen.aliyun.com/}}, MOSS\footnote{\url{https://moss.fastnlp.top/}}, ChatGLM\footnote{\url{https://chatglm.cn/}}}). 

Moreover, since our \geolm{} model provides a unified representation space and computational approach for diverse geological data described in various textual formats, it holds tremendous potential in narrowing the technological gap between different earth science tasks. 


In the subsequent sections of this technical report, we will provide a detailed description of the data collection and cleaning processes, base model selection, pre-training, supervised fine-tuning, and extensive evaluations in the creation of \textbf{\geolm{}}. Additionally, we are committed to promoting open science by making our data curation tools and pre-training checkpoints available to the research community through our GitHub repositories\footnote{The list of related tools, data, and codes can be found in \url{https://github.com/geobrain-ai/geogalactica}}.

\textbf{Broad Contribution}

In addition to establishing the academic mega-model in geoscience, our goal is to contribute to a broader research community. Specifically, the experiences documented in this paper provide evidence for further community understanding of several open questions in the literature. \textcolor{red}{Warning: The model in this manuscript might produce hallucinations and reader discretion is recommended.}

\begin{itemize}[leftmargin=1.2em]
    \item[1.] \textbf{A Domain-specific LLM:} Our construction of \geolm{}, following in the footsteps of our previous work K2~\cite{Deng2023LearningAF}, represents a geoscience LLM that focuses on interacting with humans and generating contents on highly professional academic topics.
    \item[2.] \textbf{A Toolchain for Data Cleaning:} A high-quality training dataset is crucial for successfully training large language models. Therefore, our contribution to the community includes developing an efficient academic data preprocessing toolchain to construct a clean training corpus from PDF documents~\footnote{The toolchain is open-sourced on Github repos: \url{https://github.com/Acemap/pdf_parser} and \url{https://github.com/davendw49/sciparser}. In addition, an online demo of this toolchain can be found at \url{https://sciparser.acemap.info/}.} 
    \item[3.] \textbf{Primitive Explorations to Use Tools:} 
    As for training \geolm{} to use tools, we also construct a set of supervised data \emph{Geotools} for training \geolm{} to use tools. We also open-source the codes and data on Github.\footnote{\url{https://github.com/zthang/geotools}}
    \item[4.] \textbf{Training Details and pre-training Checkpoints:} We conducted model training on the accelerator hardware provided by the Advanced Computing East China Sub-center. We will describe in detail the pre-training and SFT processes in the remainder of this paper. In addition, we are releasing the training checkpoints during the first $3/4$ of the pre-training process on Hugging Face.\footnote{\url{https://huggingface.co/geobrain-ai/geogalactica}}
    \item[5.] \textbf{Model and data analysis process:} In building a domain-specific LLM, the model and the data should be effectively evaluated and analyzed. We provide a set of analysis and visualization methods for the SFT data and the weights of the \geolm{}, open-sourced on Github.\footnote{\url{https://github.com/dbylynn/GeoGalactica_Analysis}}
\end{itemize}

In summary, we aim to contribute to the research community by developing the \geolm{} model and providing insights and tools related to data construction, training processes, and evaluation strategies. The organization of the paper can be seen in the contents section listed above.
\section{Related Work}
\subsection{Machine Learning in Geoscience}

With the advancement of artificial intelligence, utilizing machine learning, natural language processing, and recent large-scale model techniques to tackle critical problems in geoscience has become a crucial direction. Various subtasks in geoscience involve significant data collected from sensors, making them suitable for end-to-end learning using machine learning approaches. Some studies model multiple aspects of seismic signals using deep learning models to extract information relevant to earthquake prediction. Among them, \cite{Meng2023TowardEE} uses supervised learning with end-to-end training, while \cite{Fei2023UnsupervisedPS,kong2021deep} employs self-supervised learning to obtain models applied to downstream tasks. \cite{kumar2023machine,hussain2022application} utilize machine learning to explore the latent correlations among different rock properties for rock type prediction. Beyond relatively straightforward classification tasks, there are numerous works applying machine learning to address more complex scenarios in geoscience, such as calculating wellhead flow rate \cite{azim2022new}, capturing and storing carbon \cite{yao2023application}, and predicting the condition of SPBM tunnels \cite{xu2023prediction}. Additionally, machine learning is introduced to evaluate the real-world environment: \cite{chen2022high} explores the use of Few-Shot Learning (FSL) methods to enhance the accuracy of high-resolution pre-stack seismic inversion, and \cite{saha2022integrating} employs various machine learning techniques and ensemble methods to improve landslide hazard prediction, demonstrating their high practical value.
Machine learning is also being used to aid geoscience exploration, \cite{Bergen_2019} attempts to use machine learning to do data-driven modeling of solid earth science, \cite{Hulbert_2018} attempts to use machine learning to reveal the link between fast and slow earthquakes, \cite{Chen_2022} uses machine learning to reveal the impact of aerosols on climate impact.

\subsection{Natural Language Processing in Geoscience}

In addition to the diverse and heterogeneous data collected from various sensors, the field of geoscience also encompasses a significant amount of text data with standardized formats. The application of natural language processing (NLP) in earth science has witnessed remarkable progress. \cite{Liu2021GeoBERTPM,lawley2023applications} embed different sources of textual information into a unified space, \cite{Liu2021GeoBERTPM} employs joint training of language models with text and points of interest (POI) for POI retrieval, while \cite{lawley2023applications} integrates geological attribute information into the textual representation space to enable better knowledge extraction. \cite{qiu2023construction,wang2022understanding} enhance language models with knowledge graph techniques, where \cite{qiu2023construction} constructs a knowledge graph on geological text to discover ore-forming environments, and \cite{wang2022understanding} proposes an automatic entity and relation extraction approach via three-level extraction to build a geological knowledge graph from extracted information in geological reports. \cite{Denli2021GeoscienceLP} combines retrieval techniques with language models creates an integrated solution incorporating contextual retrieval and the GeoBERT model. \cite{qiuneurospe} focuses on various language irregularities encountered in natural language texts, introducing the NeuroSPE model for spatial extraction using neural networks. NLP techniques provide a unified representation space and computational approach for diverse geological data described in various textual formats, narrowing the technological gap between different earth science tasks.

\subsection{Domain-specific Large Language Model}

The recent emergence of large-scale language models marks a significant step towards unified information processing in geoscience. These models are pre-trained on vast amounts of text data and efficiently compress all input data. Currently, in addition to earth science, various domains have seen the development of domain-specific pre-trained models trained on domain-specific corpora. \cite{Beltagy2019SciBERTAP,gu2021domain,Wu2023BloombergGPTAL,Taylor2022GalacticaAL,Luo2022BioGPTGP,Xie2023DARWINSD} performs large-scale pre-training on domain-specific texts and has resulted in foundational models equipped with domain knowledge, while \cite{Lee2019BioBERTAP,Huang2019ClinicalBERTMC,Chalkidis2020LEGALBERTTM} fine-tuning these base models using domain-specific data, achieving models tailored to specific downstream tasks at a lower cost. These works have made significant strides in developing domain-specific LLMs through dedicated data integration and model training efforts.
Recently, \cite{zhang2023geogpt,ma2023impressiongpt,peng2023soft} explored the use of prompt engineering to unlock the potential of models without additional training, offering the possibility of unifying various geoscience tasks and further reducing the cost of employing large models in domain applications. In the field of geoscience, the exploration of large models is still in its early stages. \cite{Deng2023LearningAF} collected a substantial amount of high-quality data from earth science Wikipedia and research papers, and further fine-tuned the base model, leading to impressive scientific competence and knowledge in earth science. For the first time, our work utilizes a large corpus of earth science papers and textbooks, which were cleaned using a dedicated toolchain for constructing large-scale earth science models, ensuring data quality. Furthermore, our work completes the entire process of \textit{``further pre-training, supervised fine-tuning, augmented learning''} for large foundation models for geoscience, bringing the largest scale and highest quality proprietary language models to the geoscience field. This will unlock tremendous possibilities for future research conducted by earth science researchers.

We have outlined the progression of geoscience research with the use of cutting-edge AI techniques, including neural network (NN), K-nearest neighbor (KNN), recurrent neural network (RNN), convolutional neural network (CNN), backpropagation (BP), reinforcement learning (RL), support vector machine (SVM), long-short term memory (LSTM), graph convolutional neural network (GCN), Transformers, BERT, ChatGPT, and large language model (LLM).~\cite{wikiai}
The investigation reveals that the time intervals between AI technology advancements and their application in geoscience have significantly shortened, indicating an increasing reliance on advanced AI technology in the field of geoscience. The illustration is presented in Figure \ref{fig:progression}, and detailed information about the progression is shown in Appendix \ref{app:rel}.

\begin{figure}[ht]
    \centering
    \includegraphics[width=\linewidth]{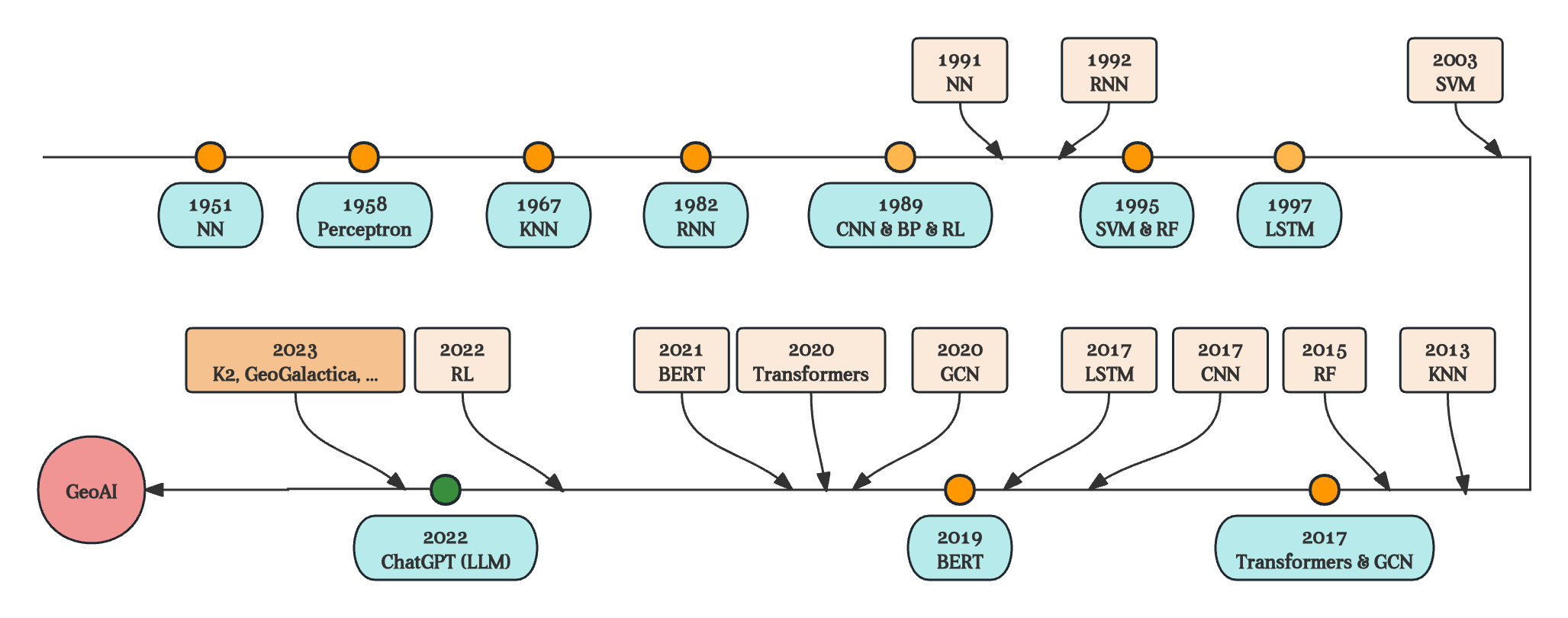}
    \caption{The progression illustration of geoscience research with the use of cutting-edge AI techniques. The textboxes in \textit{PaleTurquoise} show the techniques from computer science, The textboxes, in \textit{Bisque} show the research that probably the first time geoscientists used the techniques.}
    \label{fig:progression}
\end{figure}

\section{Preliminary and Vocabulary} 

To facilitate understanding of our work and the overview of our model, here are some key terms and their corresponding explanations that will be widely used in the narrative of this article.

\begin{table*}[h]
\resizebox{\textwidth}{!}{%
\begin{tabular}{@{}ll@{}}
\toprule
\textbf{Vocab}                  & \textbf{Stage and illustration}                                                                                  \\ \midrule
\textbf{Galactica-30B}          & The vanilla Galactica model                                                                                      \\
\textbf{GeoGalactica-FP}        & Checkpoint after pre-training over geoscience data (Further Pre-train)  \\
\textbf{GeoGalactica-Alpaca}    & Applying supervised fine-tuning with Alpaca data on top of GeoGalactica-FP \\
\textbf{GeoGalactica-GeoSignal} & Applying supervised fine-tuning with GeoSignal data on top of the first checkpoint (GeoGalactica-FP)             \\
\textbf{GeoGalactica}           & Applying supervised fine-tuning following training recipe of K2 on top of the first checkpoint (GeoGalactica-FT) \\
\textbf{GeoCorpus}              & Geoscience text corpus for pre-training\\
\textbf{GeoSignal}              & Supervised fine-tuning data for geoscience                                                                       \\
\textbf{GeoBench}           & Benchmarks for evaluating the performance of the geoscience LLM                                                  \\ 
\bottomrule
\end{tabular}%
}
\caption{Vocabulary for this technical report.}
\label{tab:vocab}
\end{table*}
Here we list the terms widely used in this report:
\begin{itemize}[leftmargin=1.2em]
    \item \textbf{Sciparser}. A PDF parsing toolkit for preparing text corpus to transfer PDF to Markdown. 
    \item \textbf{GeoTools}. A set of supervised instruction data for training \geolm{} to use tools.
    \item \textbf{K2}. The first-ever geoscience large language model trained by firstly further pre-training LLaMA on collected and cleaned geoscience literature, including geoscience open-access papers and Wikipedia pages, and secondly fine-tuning with knowledge-intensive instruction tuning data (GeoSignal).
    \item \textbf{Deep Literature}. Deep Literature is a literature platform, aiming to construct a knowledge information system for geoscience scholars, which, step-by-step goes through knowledge arrangement, knowledge mining and knowledge discovery.
    \item \textbf{GAKG}. GAKG~\cite{Deng2021GAKGAM} is a multimodal Geoscience Academic Knowledge Graph (GAKG) framework by fusing papers' illustrations, text, and bibliometric data.
    \item \textbf{DeepShovel}. DeepShovel~\cite{zhang2022deepshovel} is an Online Collaborative Platform for Data Extraction in Geoscience Literature with AI Assistance.
    \item \textbf{GSO}. Similar to WordNet, \textbf{G}eo\textbf{S}cience Ontology, (GSO) is a hierarchical tree of geological terms contains a vast amount of synonyms and word explanations, providing valuable geoscience connections between terms. 
    \item \textbf{Acemap}. AceMap is a platform displaying relationships among academic publications and a search engine for academic literature reviews.
    \item \textbf{DataExpo}. A one-stop dataset service for geoscience research.
\end{itemize}

Finally, we share the model card in~\autoref{modelcard}.
\section{Data Collection and Cleaning}

The training corpus of Galactica primarily consists of literature related to computer science and biochemistry rather than earth science. This suggests that Galactica may lack sufficient knowledge in the field of geoscience. To address this, we have collected approximately \textbf{six million} research documents specifically focused on earth science. These papers were carefully selected by professional experts in the field. Furthermore, we have expanded the GeoSignal dataset based on K2 to better support natural language processing tasks in earth science research. This expanded dataset was used for fine-tuning the model after further pre-training. In the following sections, we will provide a detailed explanation of how our dataset was constructed.

\subsection{The Customized Pre-training dataset: GeoCoprus}

According to our long-term collection efforts on geoscience papers, with the research fields subfields of geology and geography, through Acemap, we have accumulated a total of \emph{5,980,293} papers.

During this process, we commenced our data collection in early 2020 by gathering a list of journals in geoscience from LetPub~\footnote{\url{https://letpub.com.cn/}}. We identified the corresponding publishers' websites using the journal names and ISSN to collect open-access data through web page parsing. This process continued towards 2023 when we collaborated with experts in various sub-disciplines of geoscience, we collect paper from high-quality SCI journals in the mathematical geosciences, metamorphic petrology, geochronology,geomagnetism and paleomagnetism, geomorphology, tectonics, stratigraphy, hydrogeology, geophysical, geothermics, igneous and geochemistry, surficial geochemistry, geological mapping, sedimentological, petroleum geology, paleontology, paleogeography, and mineralogy. In total, we integrated a list of \textbf{849} geoscience-related journals(\autoref{appendix:geocorpus} shows the distribution of the collected papers in geoscience).

We employed the journal name list to search for journal information and their publishers' websites. Through web page scraping, we collected HTML pages and subsequently conducted data parsing to extract metadata from papers. For open-access articles, we matched them with the parsed DOI and the corresponding PDF from Sci-Hub\footnote{\url{https://sci-hub.se/}}. If no PDF was available, we downloaded it based on the URL.

Throughout this process, we adhered to the network conventions of information acquisition. When faced with obstacles such as anti-scraping measures like 5-second shields, JavaScript encryption, IP proxy bans, and account logins, we constrained our actions to ensure the compliance of our data. Moreover, data security remained our utmost priority during this process; thus, we refrained from publicly disclosing the data obtained during this stage.

In conclusion, we obtained a total of \emph{5,980,293} papers. Our data collection system operated through a distributed information fusion mechanism, utilizing an 8-workstation k8s cluster. Data collection was conducted using Scrapy-Redis~\footnote{\url{https://github.com/rmax/scrapy-redis}} framework. Additionally, we implemented compression techniques for HTML data to address challenges related to large-scale data storage.

Furthermore, we have leveraged the copyrights obtained from the publishers we have been collaborating with over the years to parse and anonymize the PDFs of these articles, creating a dataset of textual data. Additionally, referring to~\cite{suzgun2022challenging,zhang2022automatic}, we have reason to believe that the inclusion of program code in the model’s pre-training, alongside the text, can significantly enhance the reasoning capabilities of the LLM model.

Therefore, after collecting datasets from Acemap and ArXiv, we incorporated the training dataset from Codedata. Finally, our overall training corpus is detailed in~\autoref{tab:geocorpus}, totaling \emph{78B}. The data from a specific source is concatenated into a single record. After tokenization, we then split it according to a block size of 2048, with each instance ending with the \emph{tokenizer.eos} token. For each training batch, the proportion of geoscience papers to the other two datasets is $8:1:1$.

\textcolor{red}{
Notice: All the data employed in this manuscript is derived from web pages that are publicly available, thereby introducing a potential bias in the reliability of the data. Users are cautioned to be mindful of potential hallucination problems that may occur when utilizing large-scale models. Furthermore, this paper adheres to all copyright concerns. Should any issues arise, stakeholders are encouraged to notify the authors.}

\begin{table}[h]
\centering
\begin{tabular}{@{}llllll@{}}
\toprule
\textbf{Dataset}   & \textbf{\#blockNum} & \textbf{\#tokenNum} & \textbf{\#itemNum} & \textbf{\#tokenSize} & \textbf{\#batchRatio} \\ \midrule
\textit{GeoCorpus} & 25,743,070          & 52,721,798,004      & 5,548,479          & 98.21G               & 80\%                  \\
\textit{ArXiv}     & 6,691,886           & 13,704,981,558      & 742,835            & 25.53G               & 10\%                  \\
\textit{Codedata}  & 6,066,725           & 12,424,652,670      & 3,456,887          & 23.14G               & 10\%                  \\
\textbf{Total}     & 38,501,681          & 78,851,432,232      & 9,748,201          & 146.88G              & -             \\ \bottomrule
\end{tabular}%
\caption{Data distribution of the corpus used for training \geolm{}}
\label{tab:geocorpus}
\end{table}

We utilized data processing and enhancement tools based on DeepShovel and K2 during this process. With the help of Grobid~\cite{GROBID} and pdffigure2~\cite{clark2016pdffigures}, we provided a comprehensive parsing solution for extracting text, images, tables, formulas, and other data types from research documents. This was further enhanced by DeepShovel for parsing tables and formulas, resulting in the development of the SciParser tool. We plan to open-source this tool and share it on \href{https://github.com/davendw49/sciparser}{GitHub}.

Within PDF documents, there are various types of data, including text, images, tables, and formulas, all organized within the articles’ hierarchical structure and page layout. Data preprocessing is necessary to extract and ensure the readability of such content. It entails utilizing a PDF parsing tool to perform an initial parsing of the PDF document, resulting in a parsing file that contains various information from the document. However, the readability of this file is often poor, and it may have a significant amount of redundant information. Subsequently, the parsing file needs to undergo data cleansing, extracting the desired text, images, tables, formulas, and other data, and converting it into Markdown format for further processing, analysis, or display purposes.

Currently, we are utilizing Grobid as our PDF parsing tool. Grobid can accurately extract text from PDF documents and perform structured extraction of articles. It provides an outline of the text, forming an XML structure that allows for restoring the original PDF layout. Additionally, Grobid enables precise localization of images, tables, formulas, and other data types. With the provided bounding boxes, we can obtain the corresponding images using the PyMuPDF tool~\footnote{\url{https://github.com/pymupdf/PyMuPDF}}. Further leveraging the OCR recognition integrated into DeepShovel~\cite{zhang2022deepshovel}, we can convert tables, formulas, and other elements into Markdown format. The parsing process is completed by writing all the parsed content into a markdown file for further use. Throughout the entire process, for tables, we utilize the DeepShovel PDF Table Parser~\footnote{\url{https://github.com/ShaoZhang0115/Table-Extraction-for-Geoscience-Literature}}. This tool ensures the completeness and accuracy of the table content while preserving the table structure, making it convenient to reconstruct tables using Markdown. As for formulas, we employ an improved version of Latex-OCR~\footnote{\url{https://github.com/lukas-blecher/LaTeX-OCR}} for the recognition, converting the parsing results into the string format. We open-source our PDF parsing solution on GitHub~\footnote{\url{https://github.com/Acemap/pdf_parser}}. 

Tokenization is a crucial component of text corpus construction. To aid language models in comprehending academic papers, we utilize dedicated tokens for different types of special data. Finally, we use special tokens similar to the original Galactica paper \cite{Taylor2022GalacticaAL} to unify various forms of text extracted from various sources into one standard protocol. Below is an explanation of our special tokens.

\begin{itemize}[leftmargin=1.2em]
    \item Figures: We use the special tokens [START\_FIGURE] and [END\_FIGURE] to mark the captions of figures in the paper.
    \item Tables: The special tokens [START\_TABLE] and [END\_TABLE] are employed to identify the position of tables within paragraphs. During this process, we convert tables from the PDF into Markdown format.
    \item References: We use the special tokens [START\_REF] and [END\_REF] to annotate citations. The title of the article is placed within these special tokens to enhance readability.
    \item Formulas: For mathematical content or formulas, we employ regular expressions and rule-based methods to filter and clean irregular formulas parsed from the PDF. Additionally, we use the special tokens [START\_FORMULA] and [END\_FORMULA] to capture them.
\end{itemize}

\begin{figure}[h]
    \centering
    \includegraphics[width=0.8\linewidth]{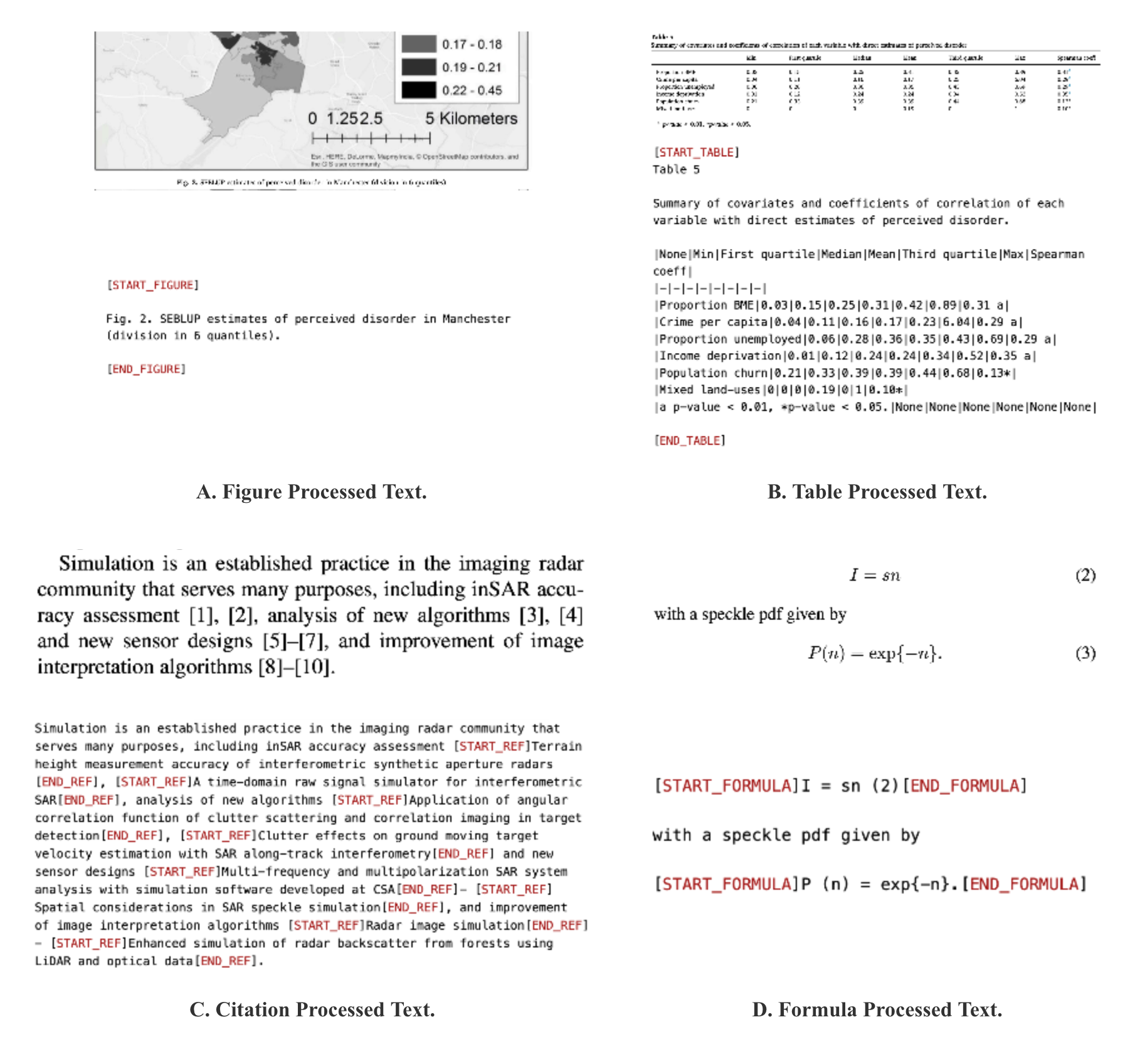}
    \caption{Tokenization processed text. A. shows an example of a figure marker, we only choose to preserve the captions; B. shows an example of a table marker, we transfer the tables into the form of Markdown; C. shows the tokenization of the citations, we replace the reference numbers into reference papers’ title to preserve the readability of the text corpus; D. shows an example of the special tokens for formulas.}
    \label{fig:prepro}
\end{figure}

And the dedicated tokens for these different types of special data are shown in~\autoref{fig:prepro} (We use the one in~\cite{Deng2023LearningAF})

\subsection{The Customized SFT dataset: GeoSignal Version 2}

Through extensive investigation and research, we have thoroughly explored natural language processing tasks specifically tailored to geoscience. In this endeavor, we have identified a set of tasks that cater to the unique requirements of geoscience applications. However, during this process, we have observed numerous unsupervised signals within these tasks that have yet to be fully harnessed and summarized.

\begin{itemize}[leftmargin=1.2em]
    \item Geoscience Knowledge Graph: Named entity recognition (NER) for temporal scales, rock types, etc., relation extraction (RE) for linking knowledge points, text-to-graph transformation, and knowledge discovery through reasoning
    \item Academic Applications: Keyword extraction, summarization, and information retrieval.
    \item General Applications: Question and Answering (Q\&A), conversations related to geoscience education, and text classification.
    \item Geographical Applications: Point of Interest (POI) queries and multimodal Q\&A.
\end{itemize}

However, the supervised signals for these tasks can be reconstructed using professional geoscience data websites. Based on the data scheme provided by K2, we further elaborate on the entire data construction process. In this process, we have built three categories of data:

\begin{itemize}[leftmargin=1.2em]
\item[1.] Literature-related data can be used to construct general natural language instructions, enabling the model to possess basic semantic reasoning abilities.

\item[2.] Geoscience-related data, which is used to build a knowledge-intensive instruction dataset, allowing the model to understand and comprehend the specific forms of natural language tasks in the field of geoscience.

\item[3.] Self-instruction-related data, following the examples of Alpaca~\cite{alpaca} and Baize~\cite{Xu2023BaizeAO}, we have distilled geoscience-related data from ChatGPT and invited geoscience experts to annotate it. This data is used to construct high-quality geoscience question-answering datasets.
\end{itemize}

\subsubsection{Domain General Natural Language Instruction}
For the general instruction learning data, we have integrated four platforms constructed by Acemap, and reconstructed the data accordingly. 

\begin{figure}[h]
    \centering
    \includegraphics[width=0.95\linewidth]{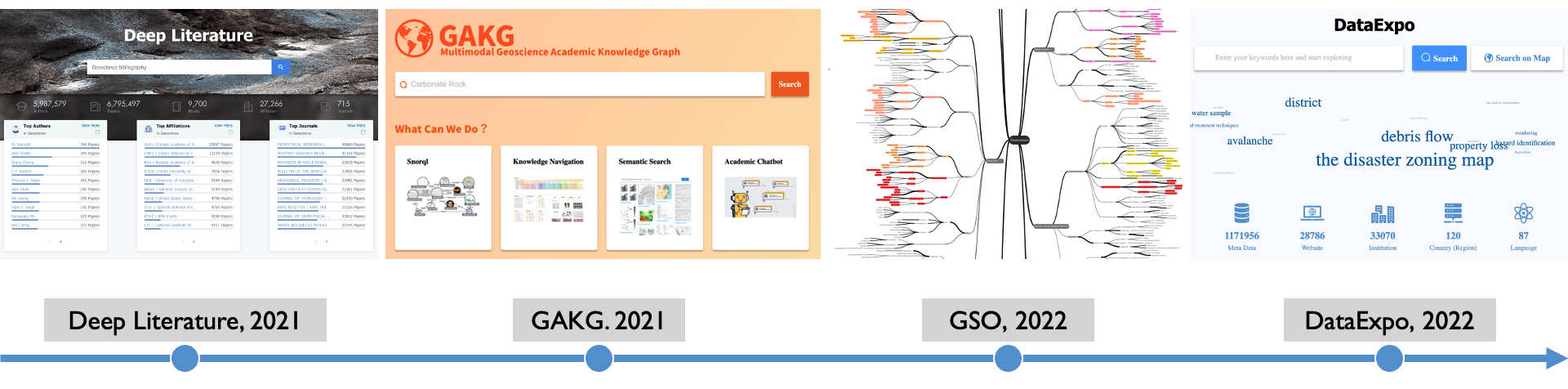}
    \caption{Four platforms that contribute most to our GeoSignal.}
    \label{fig:dde}
\end{figure}

Referring to RST~\cite{Yuan2022reStructuredP} and K2~\cite{Deng2023LearningAF}, we restructure the signals from various geoscience-related platforms. The following paragraphs will provide a detailed explanation for each platform and the illustrations for restructured domain-general natural language instruction.

\paragraph{Deep Literature and DataExpo.} This two platforms can be understood as collections of papers and datasets. Therefore, the Related Paper (with abstract) and Reference resolution of Deep Literature, as well as the Reference resolution of DataExpo, serve as excellent datasets for establishing referential relationships.

Using the text processing tool mentioned earlier, we explicitly employ a multi-threaded Grobid to process all documents and convert them into an intermediate XML format. Within the converted XML, we identify the bibl\_id of in-text citations and then locate the corresponding reference paper titles in the XML’s reference section.

\paragraph{GSO.} Similar to WordNet, the hierarchical tree of geological terms contains a vast amount of synonyms and word explanations, providing valuable supervised signals. As a result, we traverse all the nouns in GSO, extract all the synonyms for each term, and combine them with the term itself to create a set. We then construct all possible pairs of (term, synonym\_term) and add them to a list of results.

For the word description, we traverse all the nouns of GSO, extract the definition of the respective noun to serve as the description and create signal pairs (word, description). Additionally, there is also a specialized geology dictionary, which includes a dataset of categorized geology terms. The original data is in PDF format, and we convert it into JSON format through PDF parsing. In this process, we first use a parsing tool to convert the PDF into a docx format, and then use a data processing script to convert its content into JSON format. Subsequently, we proceed with content processing, removing hyphens at line breaks, and merging multiple definitions of a single term.
GSO use two geoscience dictionary. For the geology dictionary, each entry consists of a "name" and "description". For the geography knowledge dictionary, includes one more "attribute" field.

\paragraph{GAKG.} GAKG is rich in images, tables, and other elements from geology papers. Meanwhile, the text describing these images and tables, as well as their captions, can serve as excellent sequence-to-sequence (seq2seq) supervised data. Regarding the papers and their graphical information, four types of binary pairs can be generated. During this process, we transform the original text of the paper, tables, and illustrations in PNG format along with their corresponding captions, including table numbers and contents, into the target data format: \textit{(illustration caption, illustration content), (illustration caption, referring sentence), (table caption, table content), (table caption, referring sentence)}. For detailed information regarding this specific aspect, please refer to the Appendix.

Our approach to handling this is as follows:
\begin{itemize}[leftmargin=1.2em]
\item[1.] The captions and contents of tables and illustrations are stored in separate JSON files within their respective folders and can be extracted from there.
\item[2.] The referring sentences, on the other hand, need to be retrieved from the original text of the paper by referencing the table/illustration numbers mentioned in their captions.
\end{itemize}
Specifically, we search for the keywords “fig” (or variations like “Fig” and “FIG”) and “table” (or “Table” and “TABLE”) in the original text and identify the associated numbers (i.e., “i”) immediately following them. We then search for complete sentences between two periods preceding and following these numbers.

Our program handles some unexpected scenarios, such as excluding cases like “Fig11” or “Fig12” when searching for “Fig1,” and partially excluding cases where the confusion in numbering arises from referring to tables/illustrations from other papers. We also consider disorders caused by the dot used in English sentences and abbreviations, among other cases.

However, there are still a few limitations to this method:

\begin{itemize}[leftmargin=1.2em]
\item[1.] When the keywords "fig" or "table" appear at the end of a sentence, our program includes both that sentence and the subsequent one as the corresponding referring sentence.
\item[2.] There might be instances where figures/tables from other papers are referenced. Our program can identify such cases if:
    \begin{itemize}[leftmargin=1.2em]
        \item[1.] The figure/table numbers are more significant than the current paper's total number of figures/tables.
        \item[2.] The word "of" appears close after "Fig" in the text.
    \end{itemize}
\end{itemize}

In scenarios where it is difficult to discern whether a referenced figure/table belongs to another paper, we prioritize data quality. If we encounter any unmatched or garbled text, or the text is concise, we will discard that particular supervisory signal.

\paragraph{Wikipedia.} Wikipedia contains a lot of crowd-sourcing information for geoscience. Consequently, we have also incorporated geoscience data from the Wikipedia page. To retrieve the information, we utilized web scraping techniques and relevant libraries.

For the article's title, we used the Wikipedia library in Python~\footnote{\url{https://github.com/goldsmith/Wikipedia}}, which supports accessing sections of a Wikipedia page. Each section's title, text, and sub-sections form a nested structure. By recursively traversing each page, we obtained a list of triplets comprising each section's level, title, and paragraph. The triplets are structured as (level, title, paragraph), where the level indicates the depth of nesting, the title represents the section's title, and the paragraph contains the corresponding text content.

To retrieve the ``Summary \& Abstract'' of the article, we utilize the Wikipedia library in Python to access the abstract of the corresponding Wikipedia page directly. We then concatenate the paragraphs from the abovementioned sections to form the full text. Finally, we output the tuple (full text, abstract).

To extract the Entity mentioned in the article, we use the requests library and the BeautifulSoup~\footnote{\url{https://www.crummy.com/software/BeautifulSoup/}} library to scrape the Wikipedia page directly. We retrieve the text from all tags labeled "p" and "ul" and treat them as paragraphs. Next, within these paragraph tags, we search for tags labeled "a" with a href attribute starting with \textit{"/wiki/"}. These represent the highlighted blue hyperlinked sections within the text. We collect these entities and output the tuple \textit{(paragraph, entities)}.

\subsubsection{Restructured Knowledge-intensive Instruction}

In our work of building restructured knowledge-intensive instruction data, we begin by searching for authoritative websites related to paleontology, dinosaurs, fossils, rocks, and other fields within geoscience. We then filter these websites, specifically selecting those with structured data available for extraction.

\begin{table}[h]
\centering
\resizebox{\linewidth}{!}{%
\begin{tabular}{@{}lll@{}}
\toprule
Disciplines & Websites & Websites Intro. \\ \midrule
Dinosaur & https://dinoanimals.com/dinosaurdatabase/ & A comprehensive Dinosaur Database, offering a detailed catalog of dinosaurs. \\
Fossil & https://fossilcalibrations.org/ & A specialized resource offering a curated collection of fossil calibrations.  \\
Fossil & http://fossil-ontology.com/ & A multi-dimensional scientific database of fossil specimens. \\
Mineral & https://rruff.info/ & A dedicated resource for the study and identification of minerals. \\
Mineral & https://zh.mindat.org/ & A comprehensive online mineralogical database. \\
Sedimentary & https://mrdata.usgs.gov/ & A system with interactive maps and data for analyzing mineral resources on a regional and global scale. \\
Earthquake & https://www.usgs.gov/ & A website collecting all the earthquake world wide. \\
Hazard & https://public.opendatasoft.com/explore/ & A platform for exploring various datasets sorted by their modification date.  \\ \bottomrule
\end{tabular}%
}
\caption{Knowledge Intensive Data Sources.}
\label{tab:geodatasource}
\end{table}

For the websites that can be structured, we perform corresponding restructured processing like K2~\cite{Deng2023LearningAF}. Taking the provided image as an example, we match the structured data on the website using Key-Value pairs and create natural Instruction and Response pairs.

\begin{figure}[h]
    \centering
    \includegraphics[width=0.95\linewidth]{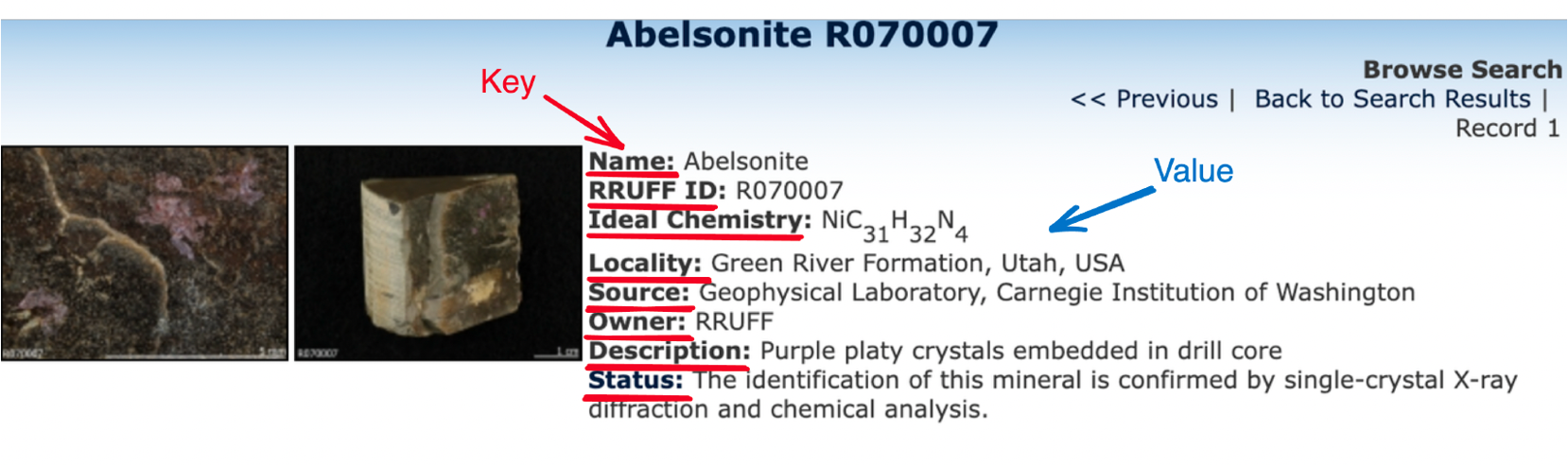}
    \caption{An example for illustrating the construction of restructured knowledge-intensive instruction data.}
    \label{fig:rst}
\end{figure}

\subsubsection{Self-Instruct}

According to Alpaca~\cite{alpaca} and Baize~\cite{Xu2023BaizeAO}, using problem seeds to generate answer from ChatGPT~\footnote{In the data curation process and experiments throughout this paper, we use the 2023 March version of ChatGPT and 2023 March version of GPT-4 unless otherwise specified} is an appropriate way to build instruction tuning data. In geoscience scenarios, we generate 1000 questions per subject under the geoscience, and we put the problem seeds on \geolm{}'s Github Repo.

In terms of overall data collection, the total amount is as follows. And we select a certain proportion of data to be included in our supervised fine-tuning process. In the final version, after further manual verification and cleaning, we choose to use a dataset of \textbf{100K} samples as GeoSignal Version 2 for instructional data during the supervised fine-tuning. The detailed statistic of the instruction tuning data is shown in~\autoref{tab:geosignal_full}.

\begin{table}[]
\renewcommand\arraystretch{1.5}
\resizebox{\textwidth}{!}{%
\begin{tabular}{cc|l|l|c}
\hline
\multicolumn{2}{c|}{} & \textbf{Signals} & \textbf{tuples} & \textbf{\#NumofSamples} \\ \hline
\multicolumn{2}{c|}{\multirow{6}{*}{\textbf{Scholar}}} & Title (with Abstract) & (abstract; title) & 2,690,569 \\ \cline{3-5} 
\multicolumn{2}{c|}{} & Abstract (with Publications Fulltext) & (fulltext; abstract) & 2,601,879 \\ \cline{3-5} 
\multicolumn{2}{c|}{} & Category (with abstract) & (abstract; category) & 12,321,212 \\ \cline{3-5} 
\multicolumn{2}{c|}{} & Related Paper (with abstract) & (source abstract; target abstract; reference sentence) & 40,047,777 \\ \cline{3-5} 
\multicolumn{2}{c|}{} & One Sentence Summary (with abstract) & (abstract; question; answer) & 2,690,569 \\ \cline{3-5} 
\multicolumn{2}{c|}{} & Reference resolution & (sentence; pronoun.; reference item) {[}including citation{]} & 2,329,820 \\ \hline
\multicolumn{2}{c|}{\multirow{2}{*}{\textbf{DataExpo}}} & Title & (abstract; title) & 216,036 \\ \cline{3-5} 
\multicolumn{2}{c|}{} & Summary \& Abstract & (fulltext; abstract) & 216,036 \\ \hline
\multicolumn{1}{c|}{\multirow{11}{*}{\textbf{GAKG}}} & \multirow{6}{*}{\textbf{GAKG}} & Principal Concepts & (sentence; entity; types) & 3,892,102 \\ \cline{3-5} 
\multicolumn{1}{c|}{} &  & Relations & (abstract; sentence; head entity; relation; tail entity) & 30,123 \\ \cline{3-5} 
\multicolumn{1}{c|}{} &  & Paper table caption & (table caption; refering sentence) & 2,772,166 \\ \cline{3-5} 
\multicolumn{1}{c|}{} &  & Paper illustration caption & (illustration caption; refering sentence) & 9,128,604 \\ \cline{3-5} 
\multicolumn{1}{c|}{} &  & Paper table content & (table caption; table content) & 2,772,166 \\ \cline{3-5} 
\multicolumn{1}{c|}{} &  & Paper illustration content & (illustration caption; illustration content) & 9,128,604 \\ \cline{2-5} 
\multicolumn{1}{c|}{} & \multirow{4}{*}{\textbf{GSO}} & Factual knowledge & (sentence; facts; improper statement) & 114,392 \\ \cline{3-5} 
\multicolumn{1}{c|}{} &  & Taxonomy & (upper term; term) & 112,298 \\ \cline{3-5} 
\multicolumn{1}{c|}{} &  & Synonyms & (term; synonym term) & 23,018 \\ \cline{3-5} 
\multicolumn{1}{c|}{} &  & Word description & (word; description; source) & 110,209 \\ \cline{2-5} 
\multicolumn{1}{c|}{} & \textbf{GA-Dialogue} & Future content and Previous content & (corrupted text; corrupted positions; target spans) & 5,434 \\ \hline
\multicolumn{1}{c|}{\multirow{8}{*}{\textbf{GeoOpenData}}} & \textbf{dinosaur} & Factual knowledge & (property; property value) & 11,348 \\ \cline{2-5} 
\multicolumn{1}{c|}{} & \textbf{fossilcalibrations} & Factual knowledge & (property; property value) & 1,749 \\ \cline{2-5} 
\multicolumn{1}{c|}{} & \textbf{fossilontology} & Factual knowledge & (property; property value) & 3,210 \\ \cline{2-5} 
\multicolumn{1}{c|}{} & \textbf{mindat} & Factual knowledge & (property; property value) & 51,291 \\ \cline{2-5} 
\multicolumn{1}{c|}{} & \textbf{ngdb} & Factual knowledge & (property; property value) & 148,212 \\ \cline{2-5} 
\multicolumn{1}{c|}{} & \textbf{opendatasoft} & Factual knowledge & (property; property value) & 37,823 \\ \cline{2-5} 
\multicolumn{1}{c|}{} & \textbf{rruff} & Factual knowledge & (property; property value) & 32,778 \\ \cline{2-5} 
\multicolumn{1}{c|}{} & \textbf{usgsearthquake} & Factual knowledge & (property; property value) & 37,284 \\ \hline
\multicolumn{2}{c|}{\multirow{2}{*}{\textbf{WordNet}}} & Synonyms & (term; synonym term) & 6,408 \\ \cline{3-5} 
\multicolumn{2}{c|}{} & Word description & (word; description; source) & 27,123 \\ \hline
\multicolumn{2}{c|}{\multirow{4}{*}{\textbf{Wikipedia}}} & Title & (term; abstract) & 3,033,595 \\ \cline{3-5} 
\multicolumn{2}{c|}{} & Summary \& Abstract & (fulltext; abstract) & 753,920 \\ \cline{3-5} 
\multicolumn{2}{c|}{} & Entity mentions & (paragraph; entities) & 3,688,926 \\ \cline{3-5} 
\multicolumn{2}{c|}{} & Relation & (text; subject; property; object) & 630,210 \\ \hline
\multicolumn{2}{c|}{\multirow{2}{*}{\textbf{IODP}}} & Title & (abstract; title) & 2,839 \\ \cline{3-5} 
\multicolumn{2}{c|}{} & Summary \& Abstract & (fulltext; abstract) & 2,638 \\ \hline
\end{tabular}%
}
\caption{GeoSignal Statistics Table.}
\label{tab:geosignal_full}
\end{table}
\definecolor{baby_blue}{RGB}{173, 216, 230}
\definecolor{dark_blue}{RGB}{25, 86, 145}
\definecolor{light_grey}{RGB}{211, 211, 211}

\section{Training}

Taking the lessons from GLM-130B~\cite{Zeng2022GLM130BAO}, we design the frameworks and plans of the \geolm{}. The following are the details of our progress.

\subsection{Further Pre-training}

After the initial pre-training by Meta AI, the model Galactica can undergo additional training on a geoscience-specific dataset. We hope this fine-tunes the model's understanding and generation capabilities in particular domains or styles.

We utilize a supercomputing cluster based on the Hygon DCU architecture, combined with the Megatron-LM framework~\cite{shoeybi2019megatron}, to further pre-train our models. The computing cluster consists of 512 nodes, with each node equipped with a 32-core CPU, 128GB of memory, and 4 DCU acceleration cards, each with 16GB of memory, resulting in a total of 2048 acceleration cards, where each acceleration card is equivalent to approximately \emph{0.2} times the computing power of an NVIDIA A100 GPU. The Megatron-LM framework employs 3D parallelism strategies, including pipeline-parallel, model-parallel, and data-parallel, to maximize GPU performance while reducing communication overhead. Given the four acceleration cards per node, we set the model parallel size to 4 for optimal model-parallel efficiency. Additionally, in the case of a mini-batch size of 1, we set the pipeline-parallel size to 16 to fully utilize the memory resources.

We preprocess all training text data by performing tokenization. The tokenized results of each document are then concatenated using an \emph{end-of-sentence (eos)} marker. Subsequently, we crop the concatenated sequences into fixed lengths of \emph{2048}, resulting in \emph{30 million} training samples, corresponding to \emph{7324} training steps. Before formally starting the training, we conduct a preliminary experimental analysis of node failures and save checkpoints at intervals of \emph{100 steps}. We initiate the pre-training process after transforming the initial checkpoint format into the format Megatron-LM requires. Ultimately, after running for \emph{16 days}, the computing cluster completes the further pre-training of the model at a speed of \emph{3 minutes per step}. Due to the frequent occurrence of node failures, the actual training takes nearly a month to complete. After the pre-training, we convert the checkpoints into the Hugging Face format for subsequent applications.

\paragraph{Challenge in further pre-training}

\begin{itemize}[leftmargin=1.2em]
\item[1.] \textbf{Over-fitting:} Further pre-training may increase the risk of overfitting, especially when the training data is relatively limited compared to the original Galactica pre-training data (refer to Section 4).
\item[2.] \textbf{Catastrophic forgetting:} In Further pre-train, ensuring that the training on the initial pre-training data is not forgotten is crucial. Sudden increases in the loss of new data sources can lead to the loss of knowledge acquired from the Galactica pre-training. It is essential to address how to effectively transfer higher-level language abilities to specific tasks and prevent the loss of the model's generality obtained during the initial pre-training during the fine-tuning process.
\item[3.] \textbf{Stability and Convergence:} Further pre-training models may be more prone to training instability and convergence difficulties. During the training process, more sophisticated optimization techniques and strategies may be required to ensure that the model converges smoothly to an appropriate state.
\end{itemize}

\paragraph{Parameters transformation from Galactica to Megatron GPT-2}

Since Galactica belongs to the OPT model, we referred to and modified the code available on Hugging Face for converting HF GPT-2 to Megatron GPT-2. The conversion parameters can be adjusted based on the actual scale of pipeline parallelism (PP), model parallelism (MP), and data parallelism (DP) during runtime.

\paragraph{Training detail}

\begin{itemize}[leftmargin=1.2em]
\item[*] \textbf{Training Setup:} In this study, we utilized a supercomputing cluster based on the hygon DCU architecture and combined it with the Megatron-LM framework for further pre-training of the model. The computing cluster consisted of 512 nodes, with each node equipped with a 32-core CPU, 128GB of memory, and 4 DCU accelerator cards with 16GB of VRAM, totaling 2048 accelerator cards, each of which is equivalent to approximately 0.2 times the computational power of an NVIDIA A100.
\item[*] \textbf{Parallel Configuration:} The Megatron-LM framework employed 3D parallelism techniques, including pipeline parallelism, model parallelism, and data parallelism, to maximize GPU performance and minimize communication overhead. Since each node had 4 accelerator cards, we set the model parallel size to 4 to achieve optimal parallel efficiency. Additionally, in cases where the mini-batch size was 1, we set the pipeline-parallel size to 16 to fully utilize the VRAM resources.
\item[*] \textbf{Data Preprocessing:} We performed tokenization on all training text data, and the tokenized results of each document were concatenated using the \textit{<eos>} marker. Subsequently, we cropped the concatenated tokens into fixed lengths of 2048, resulting in 30 million training samples, corresponding to 7324 training steps.
\item[*] \textbf{Checkpoints:} Before formally starting the training process, we analyzed the node failure patterns through preliminary experiments and saved checkpoints at intervals of 100 steps.
\item[*] \textbf{Hyperparameter Selection:} We conducted extensive experiments for hyperparameter selection in Further pre-train. Regarding learning rate scheduling, initial experiments showed that directly adopting a maximum learning rate of $1e-4$ from the Galactica-30B model led to a rapid increase in loss after a certain number of steps, resulting in training failure. Hence, we observed the gradient norm curve during the training warm-up phase and selected the learning rate corresponding to the minimum gradient norm, which was $1e-5$, as the actual maximum learning rate for training, which remained constant throughout the entire training process. For the training warm-up, we employed a linear training warm-up strategy and tested different training warm-up steps, and the optimum result was achieved with 100 training warm-up steps. Regarding other hyperparameters, we opted for the Adam optimizer with a $\beta_{1}$ of 0.9, a $\beta_{2}$ of 0.95, a weight decay rate of 0.1, and epsilon of $1e-8$. To balance effectiveness and efficiency, we set the global batch size to 4096 and utilized checkpoint activations to save VRAM. Additionally, we set the gradient clip threshold to 1.0 and a dropout rate of 0.1.
\end{itemize}

For a better understanding of our training, we list the hyperparameters of the model and the configured setting of the training in~\autoref{hparam}.

\begin{figure}[h]
    \centering {\label{fig:loss_curve}
    \includegraphics[width=1.0\linewidth]{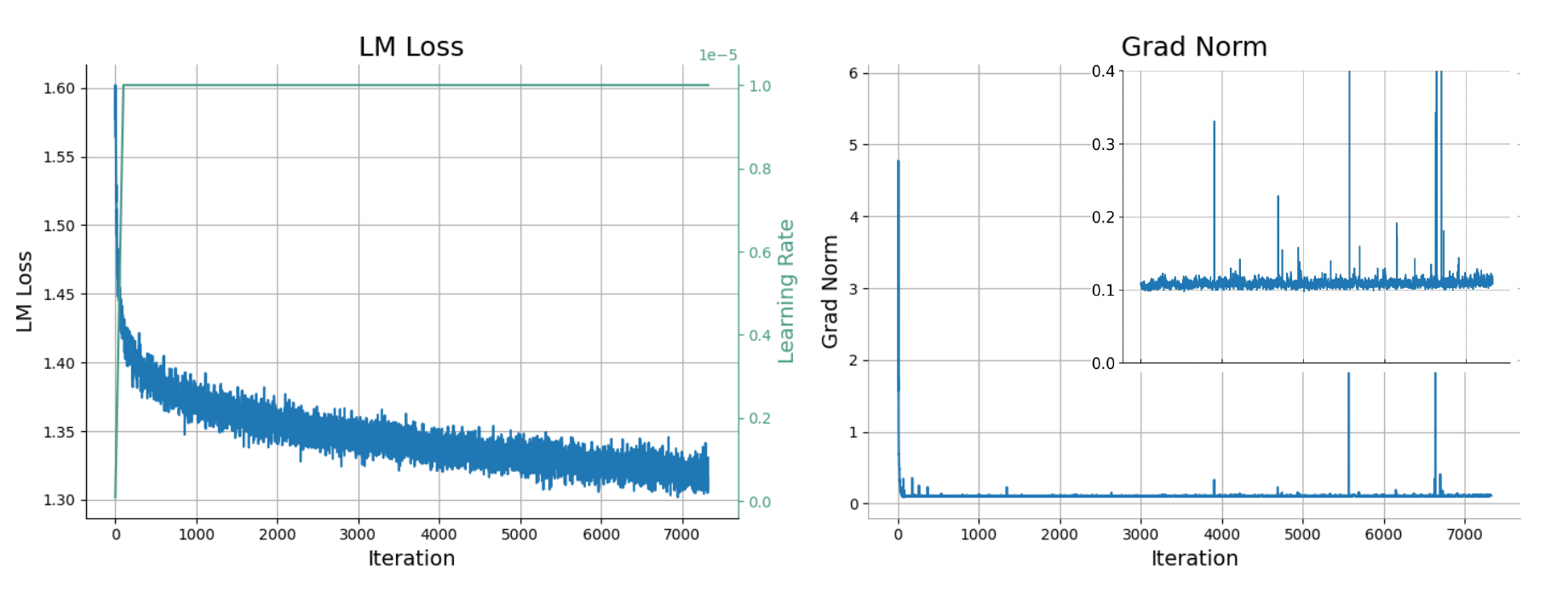}}
    \caption{Training curve during the further pre-training.}
    \label{fig:loss_grad_norm}
\end{figure}

\begin{figure}[h]
    \centering {\label{fig:local_gradnorm_loss}
    \includegraphics[width=0.5\linewidth]{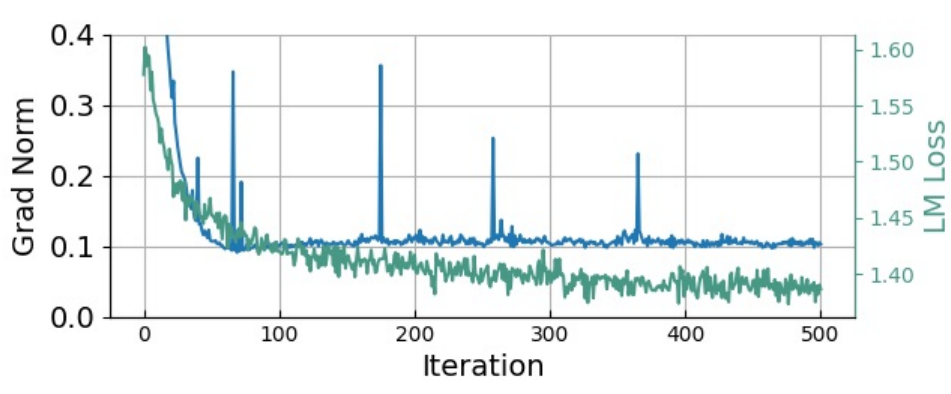}}
    \caption{Training curve of the first 500 steps during the further pre-training.}
    \label{fig:loss_grad_norm2}
\end{figure}

\paragraph{Training curves}
We share the curves of the training loss and gradient normalization as \autoref{fig:loss_grad_norm} and \autoref{fig:loss_grad_norm2}. We observed that the training loss quickly dropped from about 1.60 to 1.40 during the first 300 steps and then smoothly decreased from 1.40 to 1.32 in the subsequent steps. Although the gradient normalization showed several spikes, sharply increasing from 0.1 to approximately 0.3\textasciitilde4.8, the model exhibited no signs of saturation after further pre-training on 60 billion tokens. This demonstrates the stability of the entire further pre-training process.

\begin{figure}[h]
    \centering {
    \includegraphics[width=0.4\linewidth]{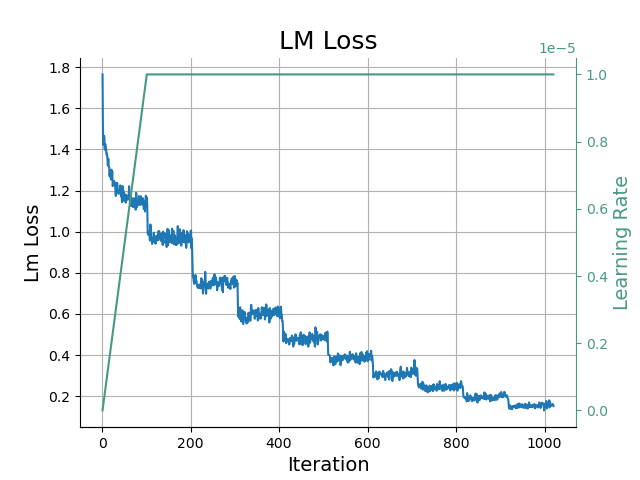}}
    \caption{Training curve during the SFT on dataset Alpaca.}
    \label{fig:alpaca_train}
\end{figure}

\begin{figure}[h]
    \centering {
    \includegraphics[width=0.85\linewidth]{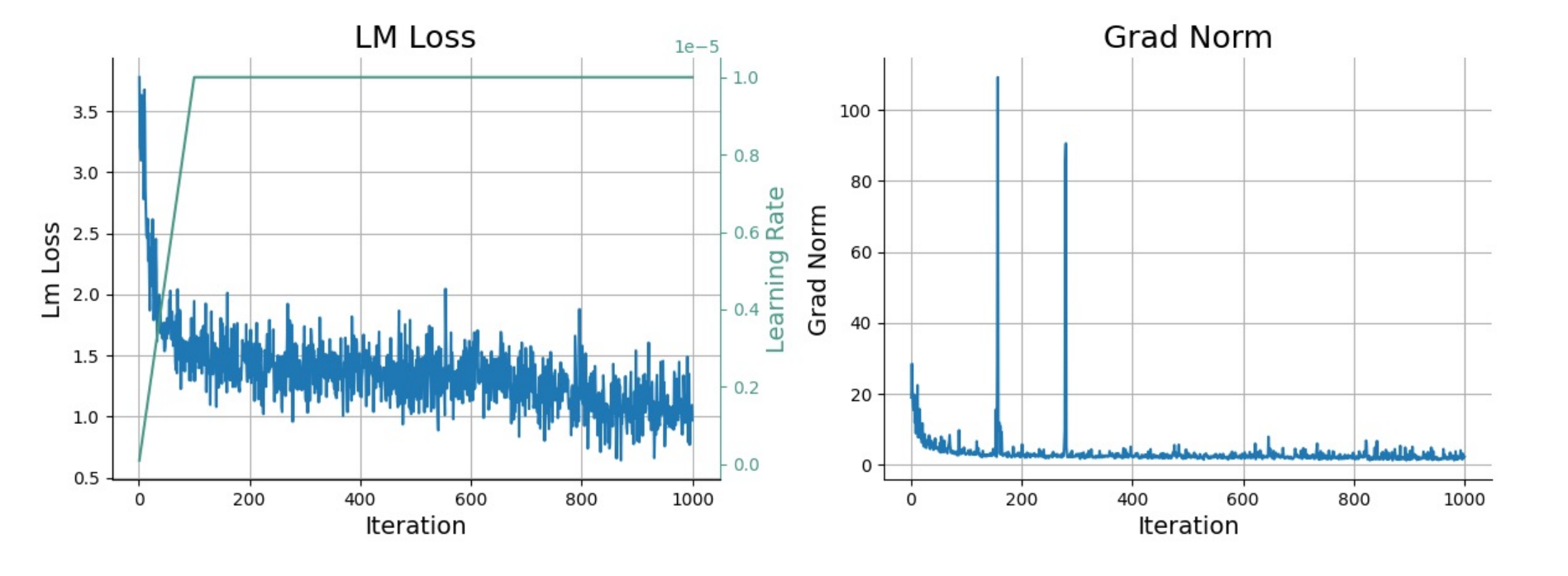}}
    \caption{Training curve during the SFT on Geosignal.}
    \label{fig:geosignal_train}
\end{figure}

\begin{figure}[h]
    \centering {
    \includegraphics[width=0.85\linewidth]{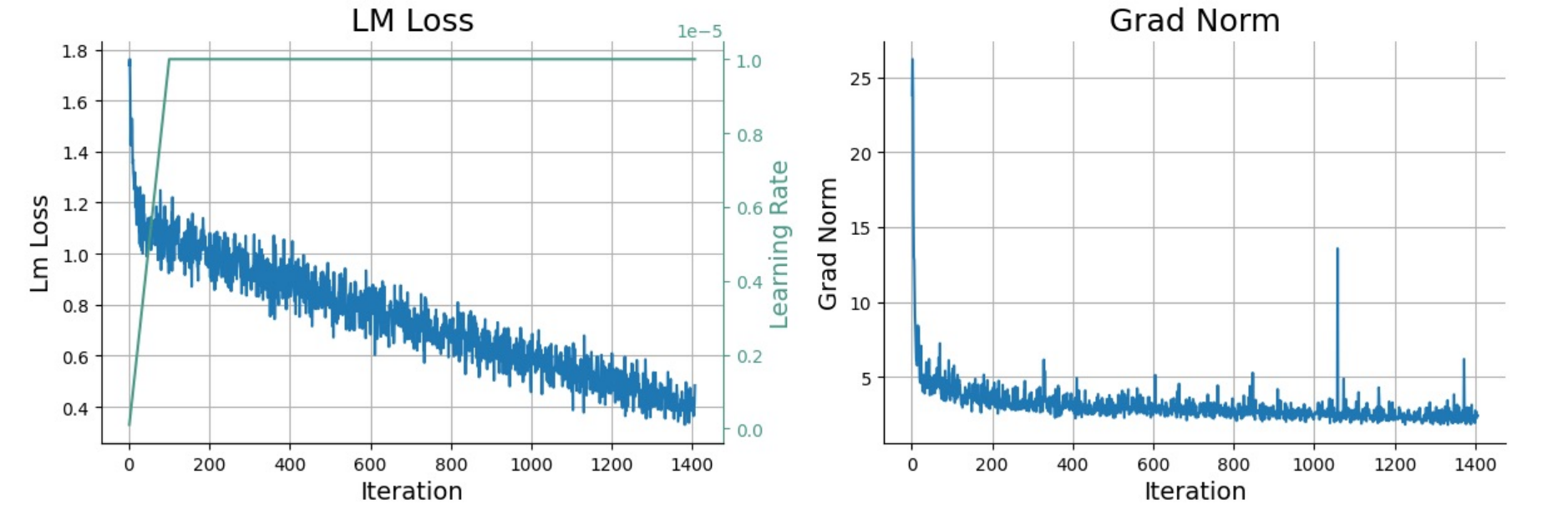}}
    \caption{Training curve during the tools SFT.}
    \label{fig:tools_train}
\end{figure}

\paragraph{The bottleneck of the training}

\begin{itemize}[leftmargin=1.2em]
\item The embedding layer is not treated as a stand-alone component in the training process. Instead, it is combined with the first transformer layer. As a result, the VRAM usage on certain cards is 60\% higher than on others, leading to decreased training efficiency. This is because a larger PP value is required to accommodate the entire model, which increases communication overhead.
\item Due to some bugs in Megatron, the "continue pre-train" function cannot utilize distributed optimizers. This results in each DP group or model replica having a complete copy of the optimizer state, significantly increasing VRAM usage.
\end{itemize}

\subsection{Supervised Fine-Tuning}

After pre-training, LLMs can be supervised fine-tuning (SFT) on a smaller, more targeted dataset under human supervision. This process adapts the model to specific tasks or improves its performance in certain areas.

We employed SFT to enhance the geoscientific reasoning performance of our large-scale models on specific geoscientific tasks. This process is essential to effectively transfer advanced language capabilities to geoscientific-specific tasks and preserve the model’s generalization acquired during pre-training.

We utilized two major frameworks, Huggingface and DeepSpeed, during this stage to facilitate our training work. This aimed to accomplish instruction fine-tuning and model prediction tasks. In the training process, the Hygon DCU cluster remained our primary resource. Compared to the pre-training stage, SFT truncation only took advantage of 128 nodes and their accompanying 512 DCUs. We continued to employ the learning rate schedule used during pre-training, where the maximum learning rate was set to $1e-5$, combined with a linear warm-up consisting of 100 warm-up steps. For the optimizer, we still selected the Adam optimizer, with $\beta_{1}$ and $\beta_{2}$ set to 0.9 and 0.999, respectively. Additionally, a weight decay of 0.05 and $\epsilon$ value of 1e-8 was chosen to better adapt to the required fine-tuning tasks.

Considering the enormous scale of the model, we utilized the DeepSpeed ZeRO3 technique for memory optimization, along with the gradient checkpoint method, to further reduce memory pressure. The maximum input sequence length was limited to 512 in this process to avoid unnecessary computational overhead. However, due to the limitations of DeepSpeed, the global batch size had to be no smaller than the number of accelerator cards. Therefore, we opted for a larger global batch size of 512. Regarding the settings of other parameters, we followed the default values of the Huggingface trainer framework. For the subsequent training, we used the Alpaca dataset and conducted training for three epochs, which only took about one day to obtain the final SFT model. This training process, supported by Megatron-LM, supported our research work.

Following the recipe proposed by K2~\cite{Deng2023LearningAF} and a similar experience in RadiologyGPT~\cite{liu2023radiology}, we did the SFT in two stages. For the first stage, we aligned the model with humans via Alpaca instruction tuning data, while using the GeoSignal v2 in the second stage. The training curve of SFT on Alpaca is \autoref{fig:alpaca_train} whole the SFT on GeoSignal is \autoref{fig:geosignal_train}

\begin{figure*}[!t]
    \centering
    \subfigure[]{\label{mapa}
    \includegraphics[width=0.44\linewidth]{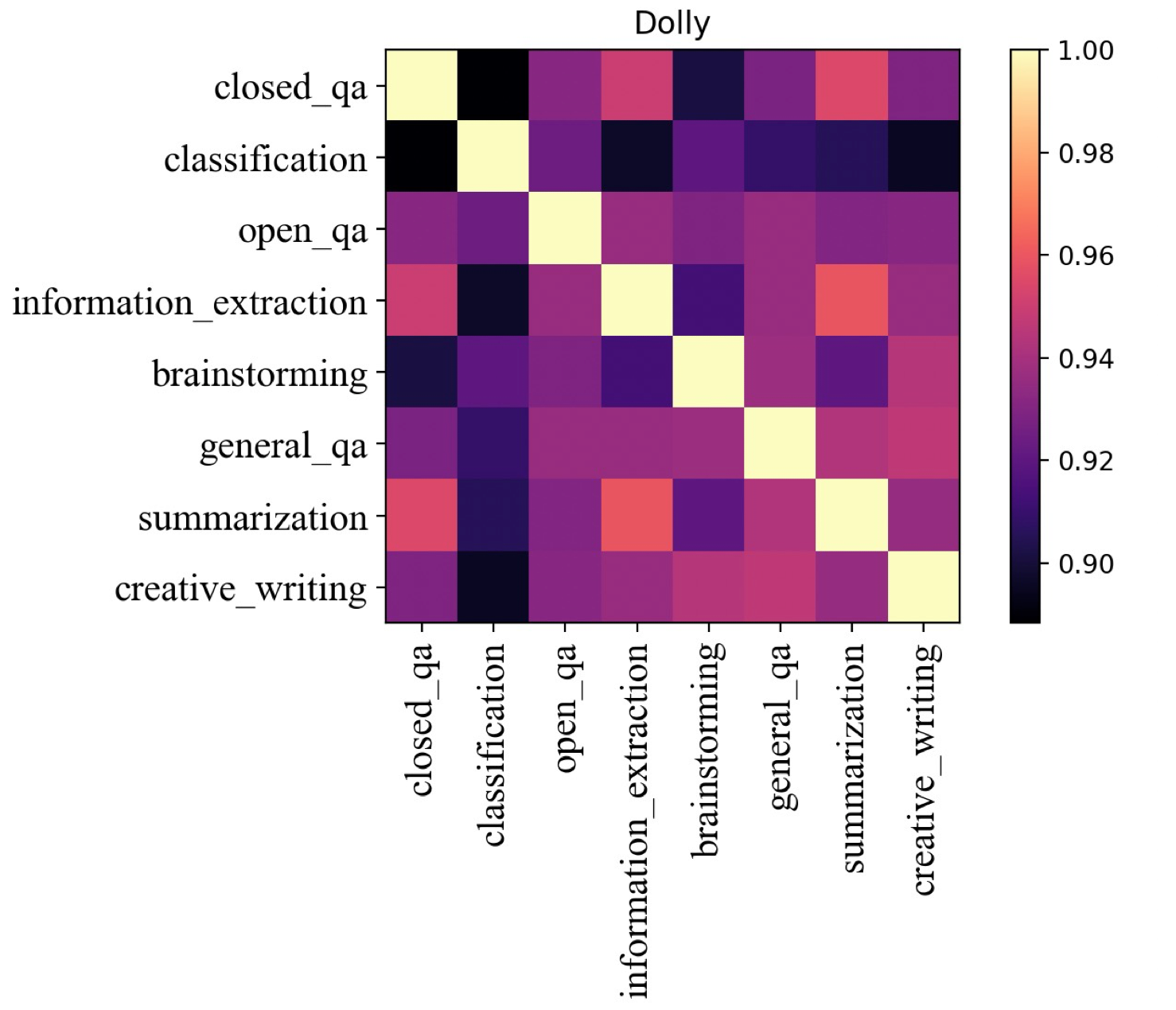}}
    \subfigure[]{\label{mapb}
    \includegraphics[width=0.44\linewidth]{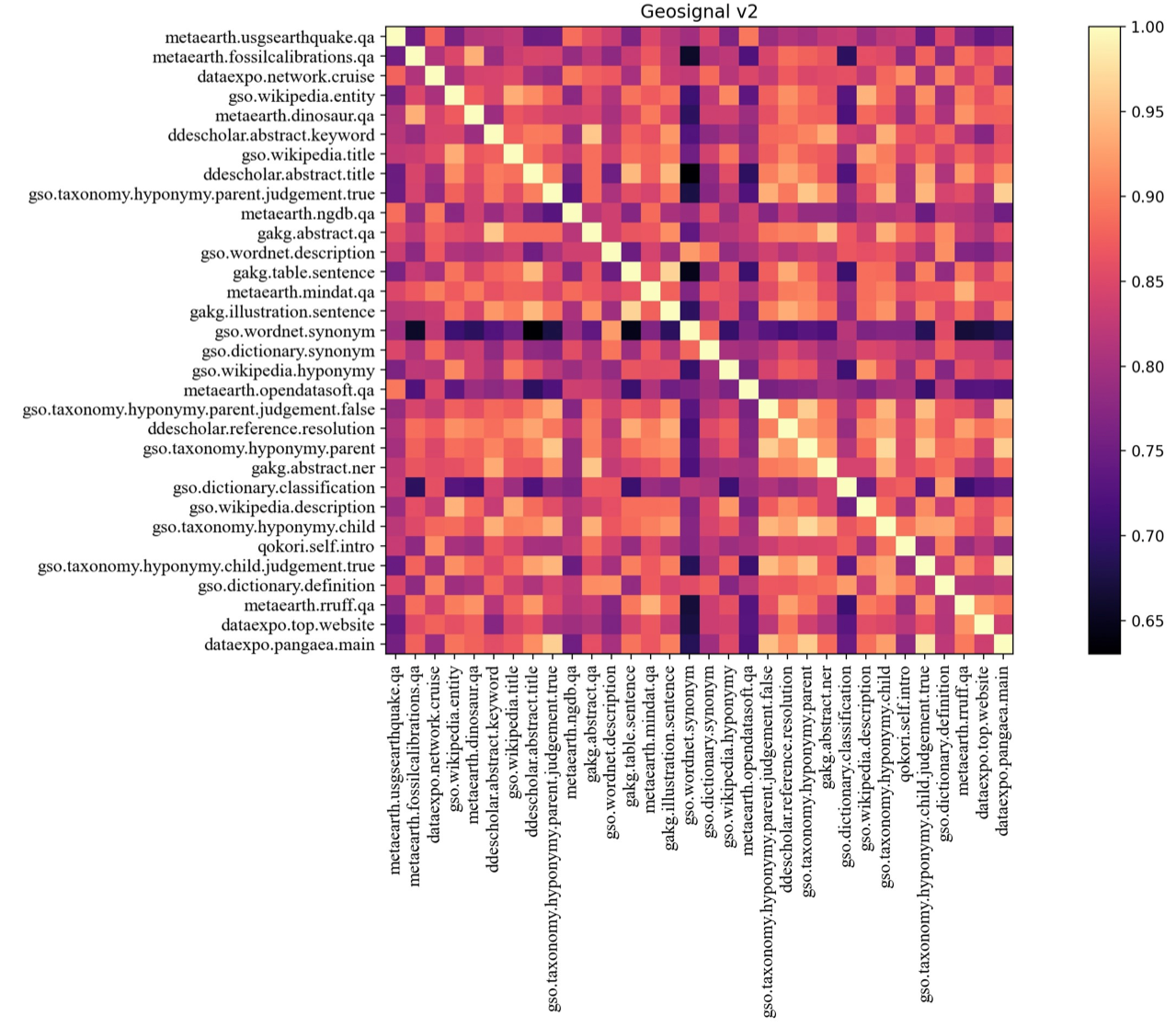}}
    \caption{Variety of the instruction tuning data in Dolly and GeoSignal.}
    \label{map}
\end{figure*}

Moreover, we compare the variety of the instruction tuning data of Dolly and GeoSignal in~\autoref{map}, showing that the general domain instructions dataset has less variety than the knowledge-intensive instructions dataset.

\subsection{Tool Learning}

In addition, LLMs can be designed to interact with and learn from various tools, such as browsers, databases, or other software interfaces. This allows the model to perform more complex tasks that require external information or specific functionalities.

We leveraged the ToolBench dataset~\cite{Qin2023ToolLLMFL}, an open-source resource, to enable geoscientific large-scale models to leverage tool API capabilities. We sampled five types of tool QA data from ToolBench, namely \textit{arxiv, bing\_search, database, weather, and wolframalpha}, and supplemented it with our collected \textbf{geo\_search} data, resulting in approximately 10k training samples. We open-source this dataset on \href{GitHub}{https://github.com/zthang/geotools}.

During the SFT stage, we trained the models together with training data such as alpaca. As the training samples from the tool data tend to be longer, we set the max\_length to be 2048. During training, we only calculate the loss and backpropagate the gradients for the part of the API call, specifically the thought, action, action input, and the corresponding tokens for the final answer.

Once the model is trained, we specify the tool’s description and corresponding API parameter instructions in the prompt. For a given question, we first let the model output the related API call (thought, action, action input) to obtain the results returned by the external tool. These results are then used as observations and fed back into the model, generating a new set of thought, action, and action input for the next iteration (if further tool calls are required). This process continues until the model gathers enough information and outputs the final answer.

Here are two naive examples of how the Galactica-30b models use the tool. The detailed examples are shown in the Appendix:

\begin{mdframed}[linecolor=baby_blue, linewidth=2pt, roundcorner=10pt, fontcolor=black, 
                shadow=true, shadowsize=2pt, shadowcolor=light_grey]
\vspace{1ex}
\begin{itemize}[leftmargin=*]
\item {\bf Example 1: }

Question: "What is the weather in New York 3M years ago?"\\
Thought: "weather"\\
Action: "geo\_search"\\
Action Input: "New York, Weather, 3M years"\\

\item {\bf Example 2}

Question: "What is the definition of plate tectonics?"\\
Thought: "arxiv"\\
Action: "search"\\
Action Input: "query: plate tectonics"\\
\end{itemize}
\centering
\end{mdframed}

\section{Evaluation}
Once we have completed the model's training, we proceed to examine its grasp of scientific and geoscientific knowledge. We have divided the evaluation into two parts.

\begin{itemize}[leftmargin=1.2em]
    \item The first part involves automated evaluation using the GeoBench provided by K2. This enables us to assess the model's performance in handling geoscientific tasks. Additionally, to examine if the newly learned knowledge has affected the pre-existing ability, we conducted MMLU (Minimal Meaningful Learning Units) tests. These tests are compared against the original Galactica model.
    \item The second part encompasses manual evaluation, where we carefully selected several subtasks from geoscience. For this evaluation, we invited 10 researchers specializing in geoscience to participate in voting and scoring. Ultimately, we compare the model's performance with five other large-scale platforms in open testing.
\end{itemize}

By conducting these evaluations, we aim to comprehensively assess the model's abilities and compare its performance against automated benchmarks and human assessments, ensuring its competence in scientific and geoscientific domains.

\subsection{Automatic Evaluation}

\subsubsection{GeoBench}
GeoBench, proposed by~\cite{Deng2023LearningAF} is a benchmarking tool specifically designed to evaluate and test the geoscientific understanding and capabilities of LLMs. It focuses on assessing how well LLMs can process and generate responses involving geographic and geological information. 

\begin{table}[h]
\centering
\begin{tabular}{@{}lcc@{}}
\toprule
\textbf{Baselines} & \textbf{NPEE} & \textbf{APTest}\\ \midrule
Random             & 27.1               & 20.0      \\
Gal-6.7B           & 25.7               & 29.9      \\
LLaMA-7B           & 21.6               & 27.6      \\
K2-7B              & 39.9               & 29.3      \\ 
ChatGPT            & \textbf{48.8}      & 20.0      \\ 
Gal-30B            & 41.2               & \underline{38.5}      \\ 
GalAlp-30B         & 42.6               & \textbf{44.1}      \\ 
\geolm{}-30B       & \underline{46.6}   & 36.9      \\ \bottomrule
\end{tabular}
\caption{comparison among baselines on Objective tasks in GeoBench.}
\label{tab:baselines}
\end{table}

Through testing our models on GeoBench, we have observed that larger and more academic models outperform benchmarks like NPEE, which are inclined toward academic research. However, they do not perform well in benchmarks like AP Study, which lean more towards foundational education. This difference may be caused by training materials that guide the model to contemplate more advanced knowledge. The training data consists of academic research achievements, namely papers, which may result in a deviation from and lack of basic knowledge. This is an area we intend to focus on for improvement in the future.

It is worth noting that Galactica, with 30 billion parameters, often fails to outperform Llama, with 7 billion parameters, in general benchmark tasks. However, in our GeoBench, we have successfully developed \geolm{}, which builds upon Galactica, surpassing K2, built upon Llama.

\subsubsection{MMLU}

The MMLU has been divided into math and non-math sections by Galactica, and we have been following their reports closely. From the results (Shown in \autoref{tab:mmlu-math}), it is evident that after processing 6 million geoscience-related literature documents, specific skills of the model, such as algebra, biology, chemistry, and mathematics, have shown improvement. This phenomenon appears to be linked to papers focusing on mathematical geology, biological geoscience, and chemical geology, highlighting the interdisciplinary nature of geoscience.
Surprisingly, machine learning has experienced significant enhancement, likely due to the inclusion of GitHub code in our corpus. In summary, subjects closely related to geoscience, including those logically connected to geology and its subfields, have shown notable progress. However, disciplines like physics indicate that the original Galactica outperforms our \geolm{} and subjects unrelated to geosciences, such as medical genetics, medicine, and electrical engineering, have shown a decline in performance. 
It is noteworthy that \geolm{} and the original Galactica are generally at a similar stage regarding average performance in math-related subjects within the MMLU.

\begin{table}[h]
\centering
\begin{tabular}{@{}lccc@{}}
\toprule
\textbf{Subject} & \textbf{\geolm{}30B} & \textbf{GAL 30B} & \textbf{GalAlp 30B} \\ \midrule
Abstract Algebra & \textbf{0.300} & 0.250 & 0.320 \\
Astronomy & 0.461 & \textbf{0.500} & 0.474 \\
College Biology & \textbf{0.576} & \textbf{0.576} & 0.514 \\
College Chemistry & \textbf{0.370} & 0.320 & 0.350 \\
College Computer Science & 0.400 & \textbf{0.410} & 0.370 \\
College Mathematics & 0.320 & \textbf{0.350} & \textbf{0.350} \\
College Medicine & 0.480 & \textbf{0.520} & 0.445 \\
College Physics & 0.284 & \textbf{0.333} & 0.294 \\
Econometrics & \textbf{0.377} & 0.368 & 0.368 \\
Electrical Engineering & 0.538 & \textbf{0.579} & 0.503 \\
Elementary Mathematics & \textbf{0.328} & 0.310 & 0.288 \\
Formal Logic & \textbf{0.302} & 0.270 & 0.278 \\
High School Biology & \textbf{0.565} & 0.561 & 0.535 \\
High School Chemistry & 0.360 & \textbf{0.399} & 0.355 \\
High School Computer Science & 0.500 & 0.480 & \textbf{0.510} \\
High School Mathematics & \textbf{0.311} & 0.256 & 0.304 \\
High School Physics & 0.298 & \textbf{0.364} & 0.325 \\
High School Statistics & 0.333 & \textbf{0.352} & 0.319 \\
Machine Learning & \textbf{0.411} & 0.339 & 0.366 \\
Medical Genetics & 0.550 & \textbf{0.580} & 0.520 \\ \midrule
\textbf{Average} & \textbf{0.4032} & \textbf{0.40585} & \textbf{0.3894} \\ \bottomrule
\end{tabular}%
\caption{We report the results of the three models in math.}
\label{tab:mmlu-math}
\end{table}

\begin{table}[h]
\centering
\begin{tabular}{@{}lccc@{}}
\toprule
\textbf{Subject} & \textbf{GeoGal 30B} & \textbf{Gal 30B} & \textbf{GalAlp 30B} \\ \midrule
Anatomy & 0.496 & \textbf{0.541} & 0.533 \\
Business Ethics & \textbf{0.430} & 0.420 & 0.470 \\
Clinical Knowledge & 0.532 & \textbf{0.555} & 0.491 \\
Computer Security & 0.600 & \textbf{0.650} & 0.620 \\
Conceptual Physics & \textbf{0.481} & 0.434 & 0.417 \\
Global Facts & \textbf{0.390} & 0.300 & 0.340 \\
High School European History & 0.533 & \textbf{0.606} & 0.491 \\
High School Geography & \textbf{0.581} & 0.540 & 0.515 \\
High: School Gov \& Politis & 0.534 & \textbf{0.565} & 0.461 \\
High School Macroeconomics & \textbf{0.408} & 0.405 & 0.367 \\
High School Microeconomics & 0.424 & \textbf{0.458} & 0.424 \\
High School Psychology & 0.613 & \textbf{0.628} & 0.556 \\
High School US History & \textbf{0.436} & 0.352 & 0.319 \\
High School World History & \textbf{0.620} & 0.456 & 0.446 \\
Human Aging & \textbf{0.552} & \textbf{0.552} & 0.511 \\
Human Sexuality & 0.511 & \textbf{0.565} & 0.481 \\
International Law & 0.612 & \textbf{0.644} & 0.554 \\
Jurisprudence & \textbf{0.491} & 0.472 & 0.444 \\
Logical Fallacies & 0.423 & \textbf{0.472} & 0.442 \\
Management & 0.573 & \textbf{0.602} & 0.515 \\
Marketing & 0.641 & \textbf{0.705} & 0.607 \\
Miscellaneous & \textbf{0.522} & 0.501 & 0.470 \\
Moral Disputes & \textbf{0.480} & 0.462 & 0.468 \\
Moral Scenarios & 0.238 & 0.244 & \textbf{0.245} \\
Nutrition & \textbf{0.536} & 0.520 & 0.448 \\
Philosophy & 0.444 & \textbf{0.492} & 0.431 \\
Prehistory & 0.503 & \textbf{0.522} & 0.435 \\
Professional Accounting & \textbf{0.344} & 0.312 & 0.319 \\
Professional Iaw & 0.326 & 0.326 & \textbf{0.327} \\
Professional Medicine & 0.438 & \textbf{0.449} & 0.379 \\
Professional Psychology & 0.472 & \textbf{0.505} & 0.449 \\
Public Relations & \textbf{0.473} & 0.445 & 0.455 \\
Security Studies & \textbf{0.424} & 0.408 & 0.322 \\
Sociology & 0.537 & \textbf{0.547} & 0.483 \\
US Foreign Policy & \textbf{0.550} & 0.510 & 0.540 \\
Virology & \textbf{0.434} & 0.422 & 0.410 \\
World Religion & 0.421 & \textbf{0.427} & 0.380 \\ \midrule
\textbf{Average} & \textbf{0.487} & \textbf{0.486} & \textbf{0.448} \\ \bottomrule
\end{tabular}%
\caption{We report the results of the three models in social sciences.}
\label{tab:mmlu-social}
\vspace{-2em}
\end{table}

After assessing the mathematical subject, we examined the results of the subjects that were excluded. Overall, \geolm{} performs slightly better than the original Galactica in the average of non-math-related subjects in MMLU. Interestingly, subjects like global facts, US History, and World History have significantly improved compared to the original Galactica. This phenomenon can be attributed to the fact that many aspects of history, such as significant discoveries and political knowledge, are closely intertwined with geoscience. This underscores the significance of geoscience, which can profoundly influence global progress.

Furthermore, in conceptual physics, learning from geoscience papers has led to a better understanding of the model. This suggests that several concepts in geoscience do not align with the knowledge taught in colleges and high schools. Consequently, models struggle to apply this related knowledge when solving problems at the college and high school levels.

\paragraph{Observation on ablation}
Fortunately, we came across Galpaca-30B on Hugging Face~\footnote{\url{https://huggingface.co/GeorgiaTechResearchInstitute/galpaca-30b}}, which significantly reduced the carbon emissions from our finetuning experiments. This model utilized Alpaca's instructions to learn from the dataset and was applied to SFT on Galactica-30B. Upon horizontal comparison, Galpaca-30B performed notably worse than the original Galactica and \geolm{} in the majority of disciplines. This indicates that instruction learning in the general domain can significantly impact the performance of specialized domain models during practical evaluations.

\subsection{Human Evaluation}
In this part, we have selected five open models to evaluate together with our \geolm{} model. These models include:
\begin{itemize}[leftmargin=1.2em]
\item[1.] MOSS, an open-source tool-augmented conversational language model, was released by Qiu Xipeng's team from the School of Computer Science at Fudan University as a ChatGPT-like model. 
\item[2.] Qwen is a chatbot developed by Alibaba Cloud, a technology company under the Alibaba Group. Alibaba announced its intention to open Tongyi Qianwen to the public, indicating its readiness for the market and reflecting China's growing focus on AI technology. 
\item[3.] ChatGPT is an AI language model developed by OpenAI, known for its ability to generate human-like text based on prompts, facilitate engaging conversations, answer questions, and perform a wide range of language-related tasks. \footnote{We use the 2023 March version of ChatGPT.}
\item[4.] Yiyan, also known as Ernie Bot, is an AI chatbot service product developed by Baidu. It has been under development since 2019 and is based on a large language model named "Ernie 4.0", which was announced on October 17, 2023
\item[5.] ChatGLM is an open bilingual language model developed by Tsinghua University. It is optimized for Chinese conversation and is based on the General Language Model architecture.
\end{itemize}

For the selected projects, our evaluation is designed as follows:

\begin{table}[]
\resizebox{\textwidth}{!}{%
\begin{tabular}{@{}llll@{}}
\toprule
\textbf{Category}   & \textbf{Problem}                                                                      & \textbf{Prompt}                                                                                                                                                                                                     & \textbf{Skills}        \\ \midrule
\textbf{Open-ended} & Noun Definition                                                                       & What is carbonate rock?                                                                                                                                                                                             & Knowledge              \\
\textbf{Open-ended} & Beginner Level Q\&A                                                                   & How many continents are there in the world?                                                                                                                                                                         & Knowledge              \\
\textbf{Open-ended} & Intermediate Level Q\&A                                                               & \begin{tabular}[c]{@{}l@{}}Would the Ohio train derailment leading to vinyl chloride leakage \\ affect the ecological environment around the Great Lakes based on \\ ocean currents or air dispersion?\end{tabular} & Analysis               \\
\textbf{Open-ended} & Advanced Level Q\&A                                                                   & How did dinosaurs become extinct?                                                                                                                                                                                   & Discovery              \\
\textbf{Functional} & \begin{tabular}[c]{@{}l@{}}Confirmation of Geoscience\\ Knowledge System\end{tabular} & Is carbonate rock a type of limestone?                                                                                                                                                                              & Judgment               \\
\textbf{Functional} & Geoscience Paper Titling                                                          & \begin{tabular}[c]{@{}l@{}}This is the abstract of my paper. \\ Can you help me come up with a title?\end{tabular}                                                                                                  & Summarization          \\
Functional          & Paper Summary                                                                         & \begin{tabular}[c]{@{}l@{}}This is my passage.\\ Can you help me summarize the passage?\end{tabular}                                                                                                                  & Summarization          \\
\textbf{Functional} & Speech Writing                                                                        & Please help me write a speech based on my topic.                                                                                                                                                                    & Writing                \\
\textbf{Functional} & \begin{tabular}[c]{@{}l@{}}Pre-requisite \\ Knowledge Recommendations\end{tabular}    & \begin{tabular}[c]{@{}l@{}}This is my article. \\ Can you recommend some prerequisite knowledge points?\end{tabular}                                                                                                & Information extraction \\ \bottomrule
\end{tabular}%
}
\caption{Tasks we designed in human evaluation parts.}
\label{tab:humaneval}
\end{table}

We refer to K2's Human Evaluation and define the evaluation metrics for open-ended questions: scientificity, correctness, and coherence (score range is [1, 2, 3]). The specific explanations are as follows:

\begin{itemize}[leftmargin=1.2em]
\item Scientificity: It represents whether the generated content appears as something that a geoscience professional would say. A score of 1 indicates not good, 2 indicates acceptable, and 3 indicates very good.
\item Correctness: From the perspective of a geoscience expert, whether the model convinces you and if the information obtained is correct. A score of 1 indicates incorrect, 2 indicates possibly right, and 3 shows correct.
\item Coherence: This metric is used to evaluate the consistency and coherence of the model, i.e., whether the text consistently discusses a specific topic and reads smoothly. A score of 1 indicates not good, 2 indicates acceptable, and 3 indicates very good.
\end{itemize}

Based on this, the cumulative score can be calculated. Additionally, for the functional questions of the large model, the evaluation metric is relative ranking. Participants in the evaluation will receive replies from all six models on the same input, and our expert judges will rate these models in the order of 1, 2, 3, 4, 5, and 6. Finally, the total ranking of each model will be calculated. In this part, we invite \textbf{10} geoscience practical people, including \emph{6} students and \emph{4} teachers. (The contribution of the human evaluation is shown in~\autoref{contribution}).

\paragraph{Open-ended Tasks}

For open-ended questions, the general large-scale model uses the interface output provided by ChatALL~\footnote{\url{https://github.com/sunner/ChatALL}} for consistency. Our large model interacts through our UI interface, where higher metric scores are preferred.

\subsubsection{Noun Definition}
In our geoscience entrance exam question set, we randomly selected 20 geoscience vocabulary terms to evaluate the model's understanding of domain-specific terminology, the whole terms are in \autoref{appendix:evaluation:noun}.

\begin{table}[h]
\centering
\begin{tabular}{lccc}
\toprule
 & Scientificity & Correctness & Coherence \\ \midrule
MOSS & 291 & 302 & 351 \\
Qianwen & 419 & 435 & 435 \\
ChatGPT & 337 & 351 & 357 \\
Yiyan & 236 & 276 & 305 \\
ChatGLM & 278 & 291 & 347 \\
\geolm{} & 339 & 361 & 393 \\ \bottomrule
\end{tabular}%
\caption{Comparison on Noun Definition tasks.}
\label{tab:noun}
\end{table}
In this task, our model has demonstrated remarkable vocabulary proficiency, but what indeed astonishes us is its exceptional ability to handle scientific questions and professional respond beyond what other models can achieve. One example is shown in~\autoref{gen_example:1}.

\subsubsection{Beginner Level Q\&A}

We selected 10 easy questions from the high school geoscience Olympiad in China and had them translated into English by professional translators. The questions are shown in \autoref{appendix:evaluation:bqa}. One example is shown in~\autoref{gen_example:2}.

\begin{table}[h]
\centering
\begin{tabular}{lccc}
\toprule
 & Scientificity & Correctness & Coherence \\ \midrule
MOSS & 116 & 120 & 147 \\
Qianwen & 191 & 177 & 207 \\
ChatGPT & 219 & 214 & 225 \\
Yiyan & 176 & 174 & 187 \\
ChatGLM & 160 & 156 & 184 \\
\geolm{} & 176 & 173 & 202 \\ \bottomrule
\end{tabular}%
\caption{Comparison on Beginner Level Q\&A.}
\label{tab:qaeasy}
\end{table}

Overall, our model ranks third when considering all aspects, but ChatGPT outperforms other models significantly in this category of questions.

\subsubsection{Intermediate Level Q\&A}

We have selected 10 moderately difficult questions from Chegg and SaveMyExam, which require a certain level of geoscience knowledge training. The questions are shown in ~\autoref{appendix:evaluation:iqa}. One example is shown in~\autoref{gen_example:3}.

\begin{table}[h]
\centering
\begin{tabular}{lccc}
\toprule
 & Scientificity & Correctness & Coherence \\ \midrule
MOSS & 143 & 154 & 178 \\
Qianwen & 178 & 180 & 193 \\
ChatGPT & 210 & 206 & 207 \\
Yiyan & 180 & 186 & 189 \\
ChatGLM & 161 & 163 & 179 \\
\geolm{} & 162 & 169 & 171 \\ \bottomrule
\end{tabular}%
\caption{Comparison on Intermediate Level Q\&A.}
\label{tab:qamed}
\end{table}

Overall, our model ranks tied for fourth place, but ChatGPT outperforms other models significantly in this category of questions. The results are shown in \autoref{tab:qamed}.

\subsubsection{Advanced Level Q\&A}

We have selected 9 highly difficult questions from the urgent geoscience problems proposed by the Institute of Geography, Chinese Academy of Sciences. These questions require extensive training in geoscience knowledge as well as the ability to reason through scientific research. The questions are shown in ~\autoref{appendix:evaluation:aqa}. One example is shown in~\autoref{gen_example:4}.

\begin{table}[h]
\centering
\begin{tabular}{lccc}
\toprule
 & Scientificity & Correctness & Coherence \\ \midrule
MOSS & 166 & 173 & 194 \\
Qianwen & 202 & 199 & 209 \\
ChatGPT & 137 & 133 & 181 \\
Yiyan & 190 & 192 & 200 \\
ChatGLM & 172 & 171 & 194 \\
\geolm{} & 185 & 187 & 206 \\  \bottomrule
\end{tabular}%
\caption{Comparison on Advanced Level Q\&A.}
\label{tab:qahard}
\end{table}

Overall, according to~\autoref{tab:qahard} our model ranks third, but ChatGPT seems to lack sufficient capability to handle these types of questions.

\paragraph{Functional Tasks}

When it comes to the evaluation of functional questions, we have chosen to apply \geolm{} to scientific research literature. \geolm{} is dedicated to facilitating the comprehension and interpretation of scientific research literature. When external information input is not required, we utilize the consistent output provided by the interface of ChatALL. In terms of overall evaluation, since it involves ranking, lower scores are preferred.

\subsubsection{Knowledge-based associative judgment question}

To determine the presence or absence of knowledge system relationships, the questions are derived from the Knowledge trees in GSO. The questions are shown in ~\autoref{appendix:evaluation:kqa}. One example is shown in~\autoref{gen_example:5}.

\begin{table}[h]
\centering
\begin{tabular}{lc}
\toprule
\textbf{Models} & \textbf{Sum of Rank} \\ \midrule
MOSS & 579 \\
Qianwen & 557 \\
ChatGPT & 600 \\
Yiyan & 570 \\
ChatGLM & 752 \\
\geolm{} & 725 \\ \bottomrule
\end{tabular}%
\caption{Comparison on knowledge-based associative judgment question.}
\label{tab:gsoeval}
\end{table}

Overall, our model ranks fifth, indicating that there is still significant room for improvement in handling these logical questions. Further advancements can be achieved by constructing CoT-type data and injecting more expertise into the model.

\subsubsection{Research Paper Titling Task}

In this phase, we randomly selected abstracts from 20 geoscience research papers and inputted them into the model, asking it to generate a title. This task showcases the model’s understanding of knowledge points and familiarity with the field. The questions are shown in ~\autoref{appendix:evaluation:rqa}. One example is shown in~\autoref{gen_example:6}.

\begin{table}[h]
\centering
\begin{tabular}{lc}
\toprule
\textbf{Models} & \textbf{Sum of Rank} \\ \midrule
MOSS & 805 \\
Qianwen & 426 \\
ChatGPT & 326 \\
Yiyan & 561 \\
ChatGLM & 440 \\
\geolm{} & 451 \\ \bottomrule
\end{tabular}%
\caption{Comparison on research paper titling task.}
\label{tab:titleeval}
\end{table}

Overall, our model ranks fourth, with no clear distinction between the ChatGLM, Qianwen, and our model in terms of performance on this task.

\subsubsection{Geoscience Research Functionality}

To ensure fairness when incorporating external research papers for evaluation, we employ our own PDF parsing solution to interpret the papers. We then use the consistent output provided by the ChatALL interface. As for our \geolm{}, we utilize our UI interface for interactions and obtain outputs accordingly.

\begin{figure}[h]
    \centering
    \includegraphics[width=0.95\linewidth]{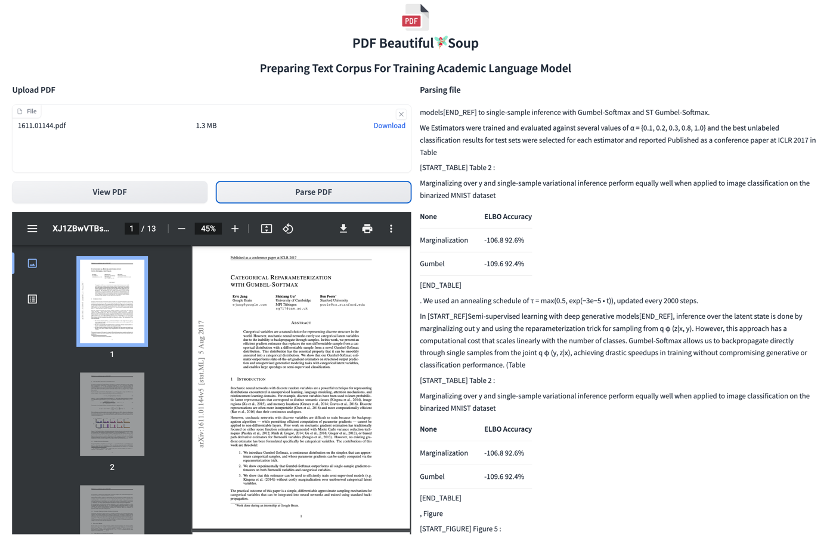}
    \caption{The user interface for evaluating the \geolm{}.}
    \label{fig:ui}
\end{figure}

More specifically, when it comes to interpreting scientific literature, we often inquire about the following aspects:
\begin{itemize}[leftmargin=1.2em]
\item[1.] Can you help me write a speech based on the content of the article?
\item[2.] Can you help me summarize the article?
\item[3.] Could you please recommend some prerequisite knowledge points? 
\end{itemize}

These three scenarios are all closely related to us as researchers. We have assessed five papers, which cover various domains of Earth sciences and are written in different styles. The papers are listed in \autoref{appendix:evaluation:gqa}. One example is shown in~\autoref{gen_example:7}.
After the evaluation, our ranking is as follows:

\begin{table}[h]
\centering
\begin{tabular}{lccc}
\toprule
 & Writing & Summary & Extraction \\ \midrule
MOSS & 114 & 164 & 115 \\
Qianwen & 135 & 185 & 232 \\
ChatGPT & 62 & 86 & 51 \\
Yiyan & 178 & 139 & 160 \\
ChatGLM & 106 & 168 & 169 \\
\geolm{} & 135 & 100 & 212 \\ \bottomrule
\end{tabular}%
\caption{Comparison on geoscience research functionality.}
\label{tab:sciaide}
\vspace{-2em}
\end{table}

Overall, considering the summation of the three scores, ChatGPT and MOSS are better than \geolm{}, and \geolm{} is tied third with ChatGLM. Our model demonstrates excellent summarization skills, thanks to its comprehensive incorporation of knowledge in the field of Earth sciences. The same principle applies to the task of extracting key information, for which we still need to gather a certain amount of expert thinking data.
\section{discussion}

\subsection{The Necessity of Pre-training}

Initiating training for a domain-specific language model from scratch in the field of geoscience is a complex decision that requires careful consideration of multiple factors. Here, we discuss some thoughts on why we did not consider training from scratch:
\begin{itemize}[leftmargin=1.2em]
\item[1.] Geoscience data is relatively limited, and training a high-quality model from scratch requires sufficient data support. The availability of geoscience data is constrained, potentially leading to the issue of data scarcity when training from scratch.
\item[2.] Training a large-scale language model from scratch demands substantial computational resources and time. As we lack sufficient resources and time, employing pre-trained models and conducting transfer learning may yield more cost-effective results.
\item[3.]Using pre-trained models and conducting transfer learning has shown promising results within a relatively short timeframe. Thus, further training can be practical. Refer to K2.
Therefore, while training from scratch would enable models to comprehend and capture domain-specific terminology, concepts, and relationships in geoscience, as well as train from the outset on geoscience data to better adapt to domain-specific language and knowledge, we opted for a strategy that involves further pre-training due to cost, time, and data considerations. 
\end{itemize}

\subsection{The Necessity of Further Pre-training}

Thus, in our perspective, employing more general-purpose models for transfer learning and further pre-training in the field of geoscience can be meaningful because:

\begin{itemize}[leftmargin=1.2em]
\item[1.] Geoscience encompasses various specialized domains, such as geology, meteorology, and environmental science. Further pre-training can enhance the model's understanding and capture of these domain-specific concepts, terms, and relationships through training on geoscience-related textual data. Additionally, geoscience often involves many domain-specific terms and contextual information that may not be commonly found in everyday language. The model can better comprehend and contextualize these terms through further pre-training, thereby improving performance in geoscience texts.
\item[2.] Geoscience texts may rely on specific geographical backgrounds, spatial and temporal relationships, and regional information. Further pre-training can assist the model in better understanding these contextual dependencies, leading to more accurate information processing and generation. Further pre-training can enhance model performance for tasks such as text classification, information extraction, and generating geological reports. The model can learn more task-specific feature representations by training on relevant domain-specific data.
\item[3.] During this process, we observed the alleviation of data scarcity in geoscience. In certain domains within geoscience, scarce data may limit training samples. However, further pre-training enables the model to learn general language abilities from a larger-scale dataset and subsequently fine-tune on a smaller amount of domain-specific data, mitigating the impact of data scarcity.
\end{itemize}

In conclusion, further pre-training can enable large language models to better adapt to the characteristics and requirements of the geoscience field, leading to enhanced performance in geoscience text processing and task execution.

\subsection{Carbon Emissions}

During our cumulative training, \textbf{1,488,137.26 DCU hours} were consumed, resulting in cumulative carbon emissions of 212 $tCO_{2}eq$ calculated by \autoref{carbon_exp}. Our work provides a foundational model for subsequent geoscience researchers to fine-tune their smaller models, potentially reducing carbon emissions in their future work.
\begin{equation}
    \label{carbon_exp}
    C_{Emission} (kg) = DCU\ hours * TDP\ (kW) * C_{Intensity}\ (kg/kWh)
\end{equation}

\subsection{Towards Unified Foundation Model in Geoscience}

The application of artificial intelligence in geosciences demonstrates vast prospects. In terms of geoscientific literature analysis, AGI systems especially the unified foundation model can assist researchers in identifying the frequency of specific vocabulary while addressing any ambiguities, thereby enhancing the accuracy of literature comprehension. Furthermore, AGI can integrate dispersed geoscientific knowledge by analyzing extensive literature uncovering novel correlations and trends, thus providing new perspectives and directions for geoscience research. Additionally, AGI systems can aid in geoscience education, offering personalized content and teaching methods to facilitate students' ease of learning and understanding of geoscientific knowledge.

On the other hand, the application of a unified foundation model in the geoscience domain extends beyond academic research to practical uses such as geological hazard warnings, resource exploration, and environmental protection. Regarding geological hazard warnings, AGI can utilize big data and models to provide accurate predictions and assessments, helping to mitigate the damages caused by natural disasters. Concurrently, a unified foundation model plays a crucial role in underground resource exploration, improving efficiency and accuracy in exploration endeavors. In the realm of environmental protection, a unified foundation model aids in the real-time monitoring of environmental conditions through the analysis of remote sensing data, supporting decision-making processes for environmental conservation.

In the future, with sufficient abundant data and computing powers, and other feasibility of achieving unified foundation models in the field of geoscience, the future of AGI in Geoscience can be expected. In the future, the unified foundation model will continue to play a role in advancing frontiers in geoscience research. It can assist scientists in conducting large-scale data analysis, unraveling complex phenomena such as internal Earth structures, plate tectonics, and crustal evolution, thereby providing deeper scientific comprehension. AGI also contributes to environmental monitoring and protection, aeromagnetic data interpretation, water resource management, carbon capture, and other domains. By analyzing hydrological data and geological information, a unified foundation model can predict groundwater resources' distribution and sustainable utilization, providing scientific foundations for water resource management. In carbon capture, a unified foundation model can assist researchers in selecting suitable geological storage layers and sealing rocks, thereby driving the development of carbon reduction technologies. Overall, AGI accelerates the accumulation of scientific knowledge in geosciences and offers unprecedented support in addressing global challenges, providing robust intelligent assistance for humanity's future sustainable development.
\section{Conclusion}
In conclusion, the utility of NLP in geoscience research and practice is vast, and large language models (LLMs) have shown great success in various NLP domains. However, specialized LLMs in geoscience are scarce. We introduce \geolm{}, a 30B parameters language model designed explicitly for geoscience applications. Through training on a comprehensive geoscience academic dataset and fine-tuning with geoscience-knowledge intensive instruction pairs, \geolm{} outperforms existing models in geoscience NLP tasks. Our validation with senior geoscientists confirms its effectiveness. The release of \geolm{} and our training experience aims to contribute to the advancement of unified foundation models in geoscience.

\section*{Acknowledgement}
The computation resource was supported by the Advanced Computing East China Sub-center. 
This work is supported by NSF China (No.62020106005, 61960206002, 42050105, 62061146002, 62106143), National Key Technologies R\&D Program (No. 2022YFB3904201), Shanghai Pilot Program for Basic Research - Shanghai Jiao Tong University. 
The second author would like to thank Wu Wen Jun Honorary Doctoral Scholarship, AI Institute, Shanghai Jiao Tong University.

\newpage
\bibliographystyle{unsrt}
\bibliography{ref}  

\newpage

\appendix
\section{Appendix: Progression of geoscience with AI}
\label{app:rel}
Here we show the progression of geoscience research with the use of cutting-edge AI techniques summarized by ~\cite{wikiai}.
\begin{table}[h]
\resizebox{\textwidth}{!}{%
\begin{tabular}{@{}llll@{}}
\toprule
\textbf{Methods} & \textbf{In CS} & \textbf{In Geoscience} & \textbf{Gap} \\ \midrule
NN & 1951 & \textbf{1991} Neural computing in geophysics~\cite{McCormack1991NeuralCI}& 40 \\
Perceptron & 1958 & - &  \\
KNN & 1967 & \textbf{2013} Methods to compute fault images, extract fault surfaces, and estimate fault throws from 3D seismic images~\cite{Hale2013MethodsTC} & 46 \\
CNN/BP & 1980/1989 & \textbf{2017} Salt classification using deep learning~\cite{Waldeland2017SaltCU}& 37 \\
RNN & 1982 & \textbf{1992} Adaptive minimum prediction-error deconvolution and source wavelet estimation using Hopfield neural networks~\cite{Wang1992AdaptiveMP} & 10 \\
BP & 1986 & - &  \\
RL & 1989 & \textbf{2022} Deep reinforcement learning for optimal well control in subsurface systems with uncertain geology~\cite{Nasir2022DeepRL} & 33 \\
Random Forest & 1995 & \textbf{2015} Supervised learning to detect salt body~\cite{Guilln2015SupervisedLT} & 20 \\
SVM & 1995 & \textbf{2003} A support vector machine for avo interpretation~\cite{Kuzma2003ASV} & 8 \\
LSTM & 1997 & \textbf{2017} Machine learning can extract the information needed for modelling and data analysing from unstructured documents~\cite{Blondelle2017MachineLC} & 20 \\
Transformer & 2017 & \textbf{2020} Using Transformer Networks and Knowledge Graphs in Earth Science Literature to Synthesize Mass Information for Transdisciplinary Research~\cite{2020AGUFMIN030..04Z} & 3 \\
GCN & 2017 & \textbf{2020} Graph Convolutional Networks for Hyperspectral Image Classification~\cite{Hong2020GraphCN} & 3 \\
BERT & 2018 & \textbf{2021} BERT-E: An Earth Science Specific Language Model for Domain-Specific Downstream Tasks~\cite{2021AGUFMIN15B..06K} & 3 \\
ChatGPT/LLM & 2022 & \textbf{2023} K2: A Foundation Language Model for Geoscience Knowledge Understanding and Utilization~\cite{Deng2023LearningAF} & 1 \\ \bottomrule
\end{tabular}%
}
\caption{The progression of geoscience research with the use of cutting-edge AI techniques.}
\label{tab:rel}
\end{table}
\section{Appendix: GeoCorpus}
\label{appendix:geocorpus}
Here we show the distribution of the collected papers from top-10 amounts journals in geoscience.

\begin{figure}[h]
    \centering
    \includegraphics[width=0.95\linewidth]{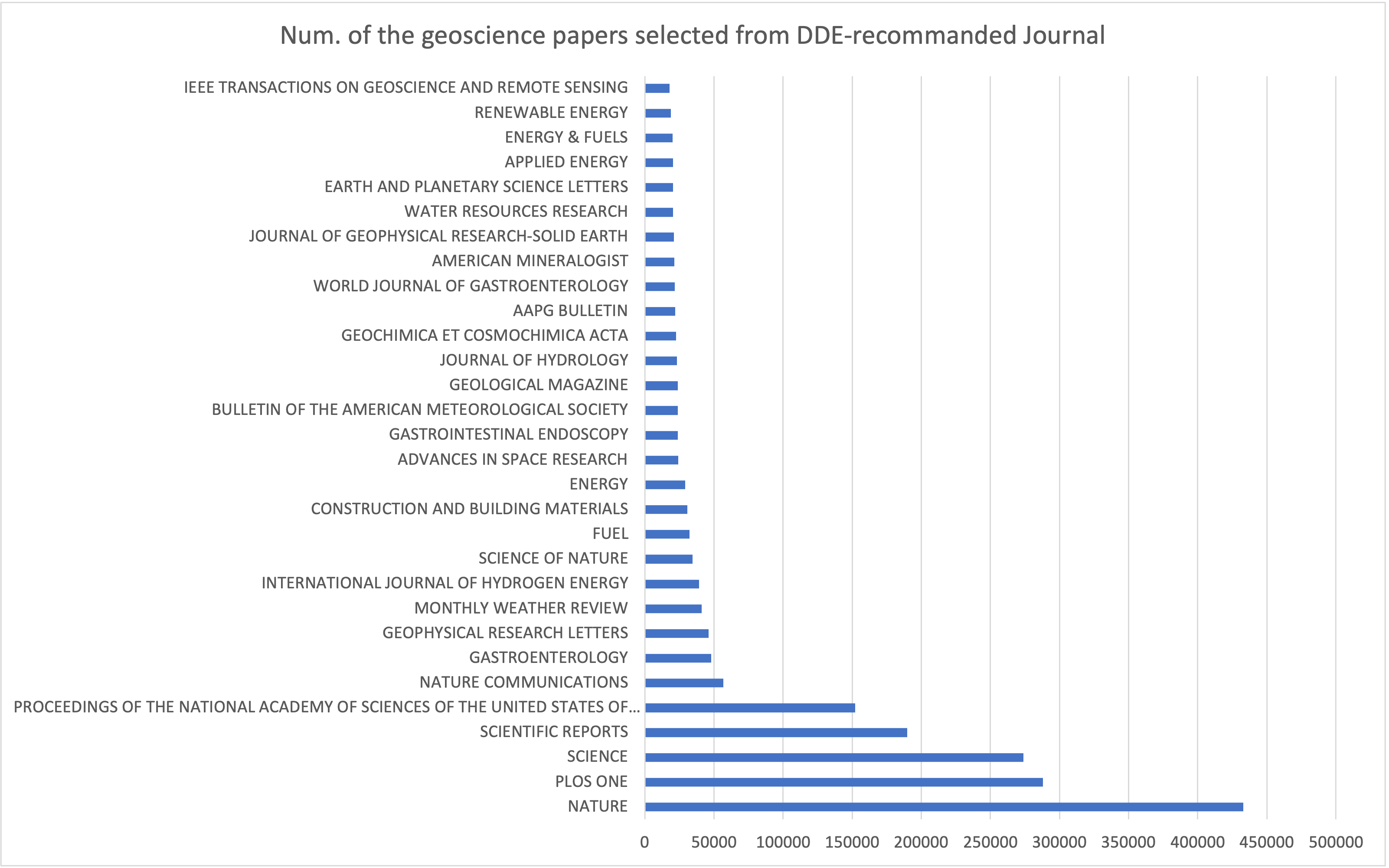}
    \caption{Distribution of the collected papers from top-10 amounts journals in geoscience.}
    \label{fig:papercount}
\end{figure}

\newpage

\section{Appendix: GeoSignal V2 Curation}
\label{appendix:geosignal}
Below, we will provide a detailed explanation of how we obtain useful supervision signals for geoscience tasks from websites like MinDat, USGS, NGDB, Fossil Ontology, and Fossil calibrations.

\subsection{MinDat}
\begin{figure}[h]
    \centering
    \includegraphics[width=0.95\linewidth]{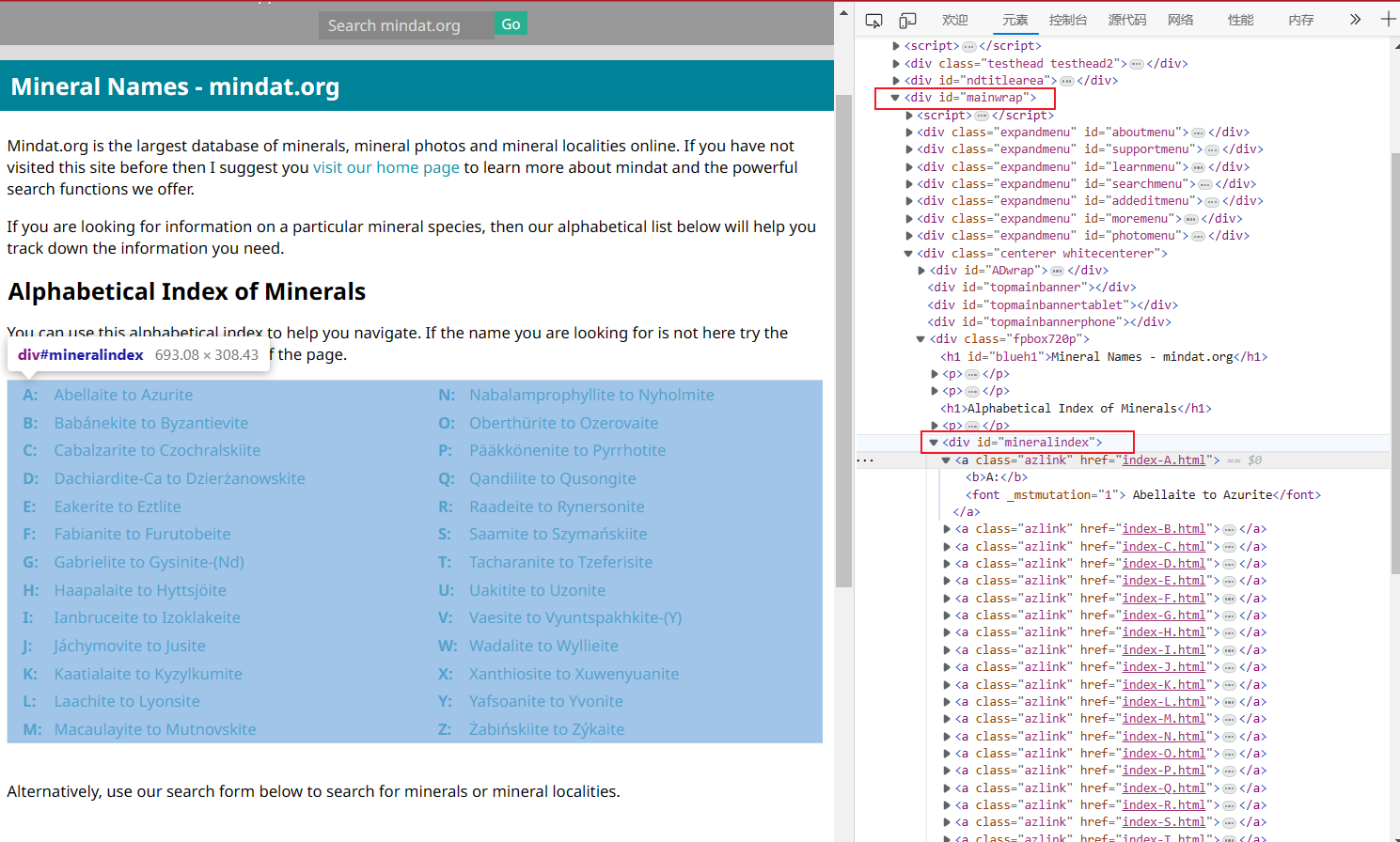}
    \caption{Mines in MinDat.}
    \label{fig:md1}
\end{figure}

\begin{figure}[h]
    \centering
    \includegraphics[width=0.95\linewidth]{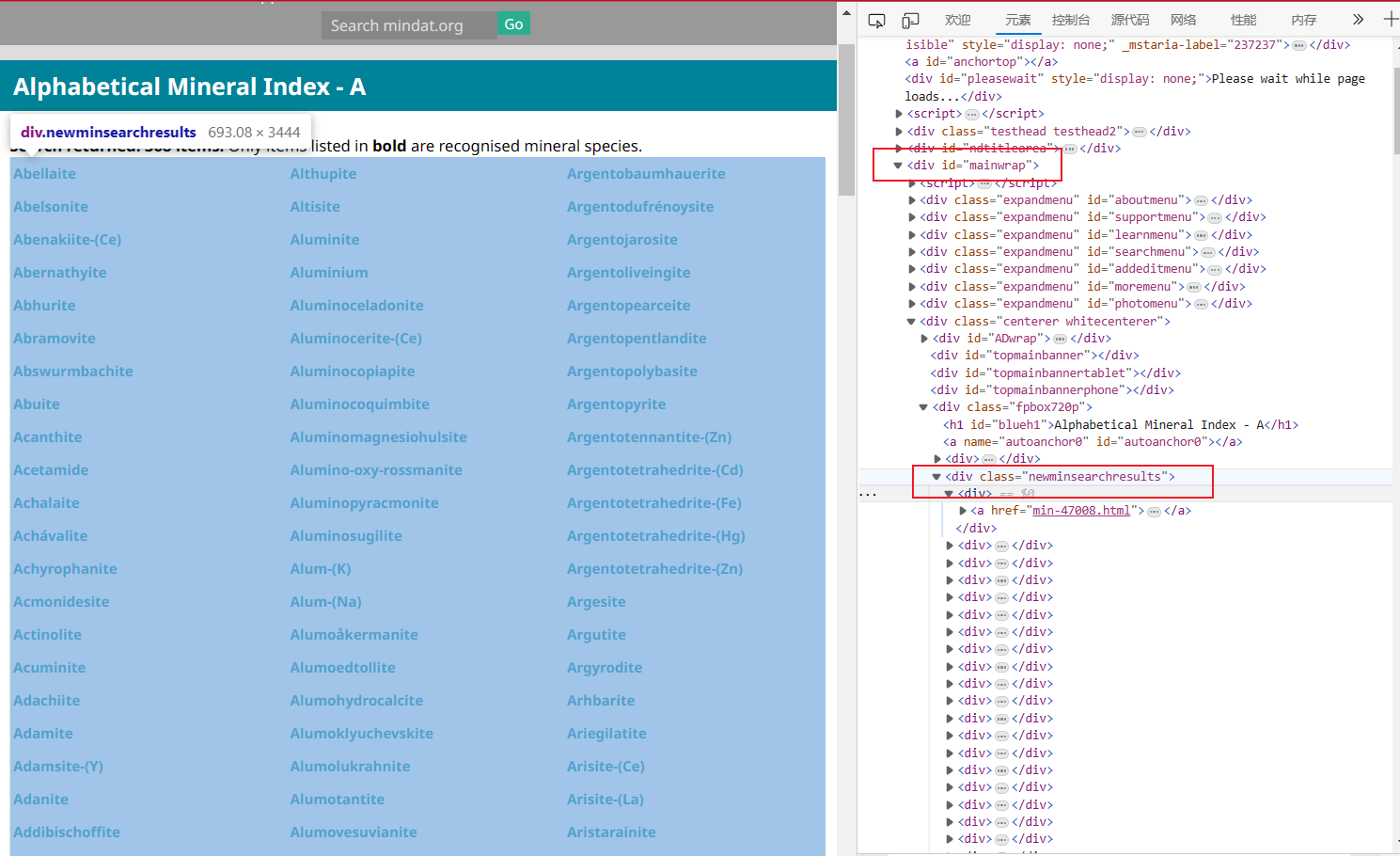}
    \caption{Mines in MinDat.}
    \label{fig:md2}
\end{figure}

\begin{figure}[h]
    \centering
    \includegraphics[width=0.95\linewidth]{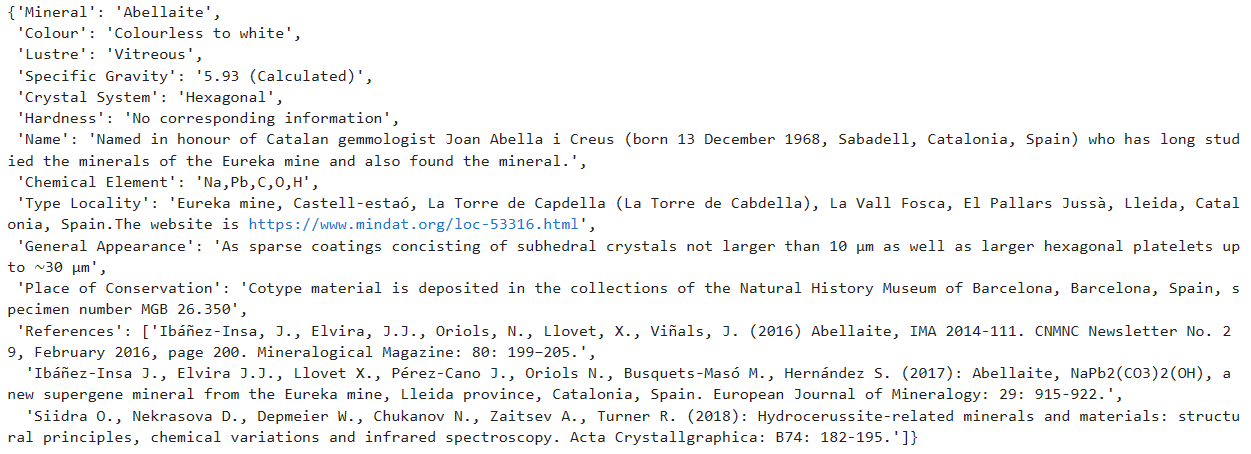}
    \caption{Signals in MinDat.}
    \label{fig:md3}
\end{figure}

This website provides a list of all minerals in alphabetical order, and it is possible to obtain the collection of all minerals starting with a certain letter from this page. (As shown in \autoref{fig:md1})
After scraping the HTML page, the following Python spider logic can be used to retrieve the URLs of all minerals starting with letters A-Z: (as shown in \autoref{fig:md2})
find(id="mainwrap") → find(id="mineralindex")

To extract the relevant information for each mineral, we can create a dictionary for each mineral containing the following information:

\begin{itemize}[leftmargin=1.2em]
    \item \textbf{Mineral Name.} Key: Mineral, Value: Name of Mineral.
    \item \textbf{Physical Properties.} Extract information on the physical properties of the mineral, including Colour, Lustre, Specific Gravity, Crystal System, Hardness, Name.
    \item \textbf{Chemical Properties.} Extract the chemical elements present in the mineral as Chemical Element and flatten them. For example, Abellaite should be Na, Pb, C, O, H.
    \item \textbf{Type and Occurrence.} Extract information on the Type and Occurrence of the mineral, including Type locality (which will be a separate URL in the format shown below), General Appearance, Place of Conservation.
    \item \textbf{References.} Extract and include the References in the dictionary, with value as a list.
    
\end{itemize}

Not all minerals will have all the above information. For minerals with missing information, we will still create a dictionary entry but the value will be "No corresponding information". For example, the dictionary for Abellaite will be illustrated in \autoref{fig:md3}.

\subsection{USGS}

\begin{figure}[H]
    \centering
    \includegraphics[width=0.95\linewidth]{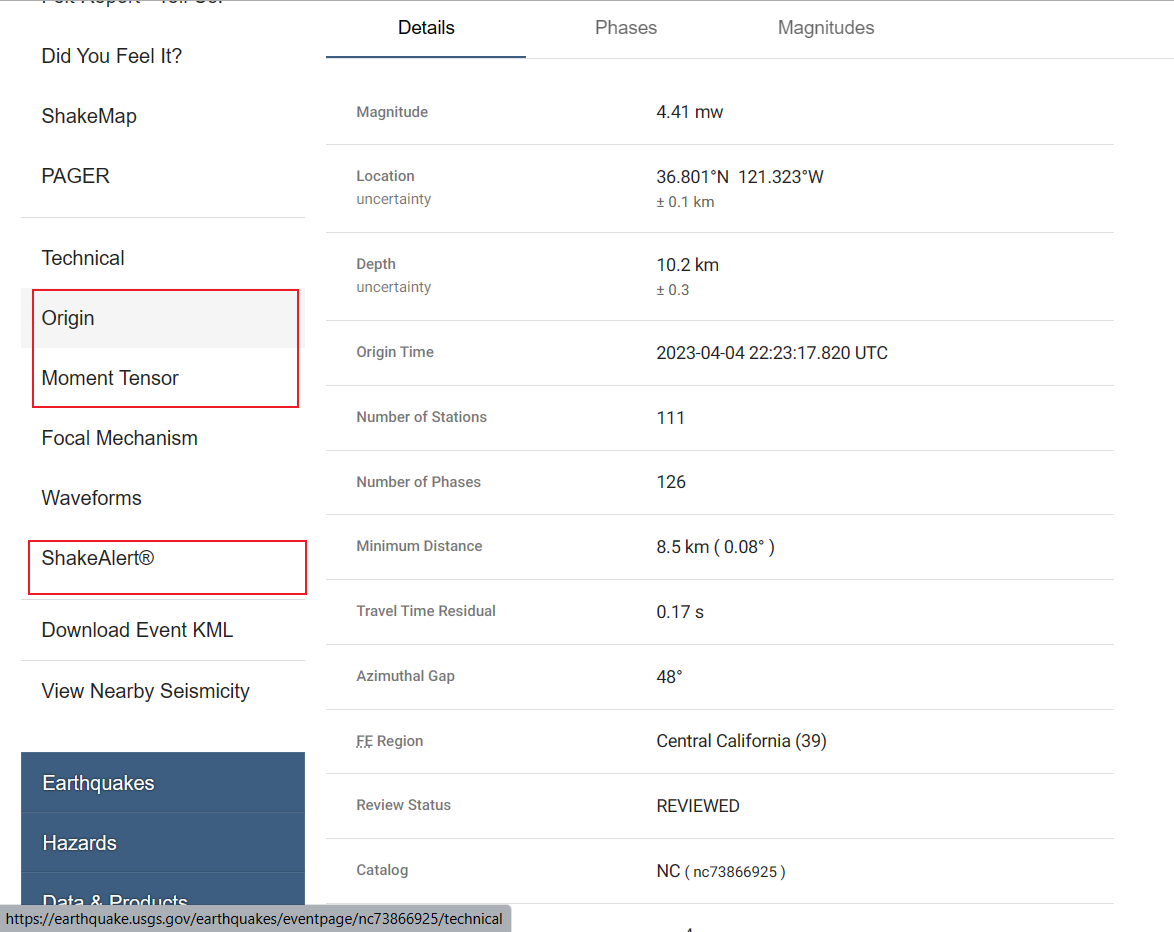}
    \caption{USGS collections.}
    \label{fig:usgs1}
\end{figure}

\begin{figure}[H]
    \centering
    \includegraphics[width=0.95\linewidth]{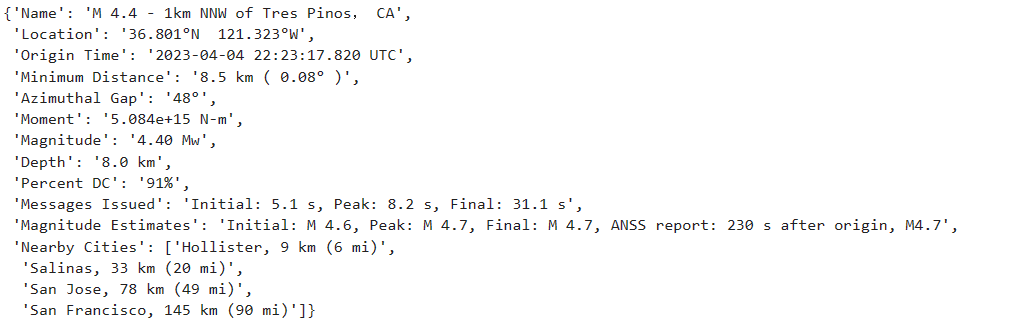}
    \caption{USGS signals.}
    \label{fig:usgs2}
\end{figure}

When the webpage is opened, it displays important earthquake information from 1900 to 2023 around the world, and the information can be organized by year. For a given year "x", the corresponding webpage for significant earthquakes in that year can be accessed through the URL: \url{https://earthquake.usgs.gov/earthquakes/browse/significant.php?year=x}

For example, the webpage for significant earthquakes in the year 2023 can be accessed through the URL:
\url{https://earthquake.usgs.gov/earthquakes/browse/significant.php?year=2023}

The webpage for the year 2002 can be accessed through the URL:
\url{https://earthquake.usgs.gov/earthquakes/browse/significant.php?year=2002}

Therefore, using this method, all significant earthquake information pages from 1900 to 2023 can be obtained.

For each specific webpage corresponding to a given year, the list of earthquake URLs can be obtained using the function "find(class='eqitem')".

\begin{itemize}[leftmargin=1.2em]
\item \textbf{Basic Information.} The basic information that needs to be obtained is the name, with the key value set as "Name" and the value as the name obtained from the webpage title.

\item \textbf{Origin.} The information needed for this section includes: Location, Origin Time, Minimum Distance, and Azimuthal Gap.

\item \textbf{Moment Tensor.} The information needed for this section includes: Moment, Magnitude, Depth, and Percent DC.

\item \textbf{Post ShakeAlert.} The information needed for this section includes: Messages Issued, Magnitude Estimates, and Nearby Cities. Since there may be multiple nearby cities, a list will be created to store them as the value for this specific dictionary entry.

\end{itemize}

Finally, we obtain the signals in USGS, specifying which signals are retrieved and providing details on how they are processed.

\subsection{NGDB}

\begin{figure}[btp]
    \centering
    \includegraphics[width=0.95\linewidth]{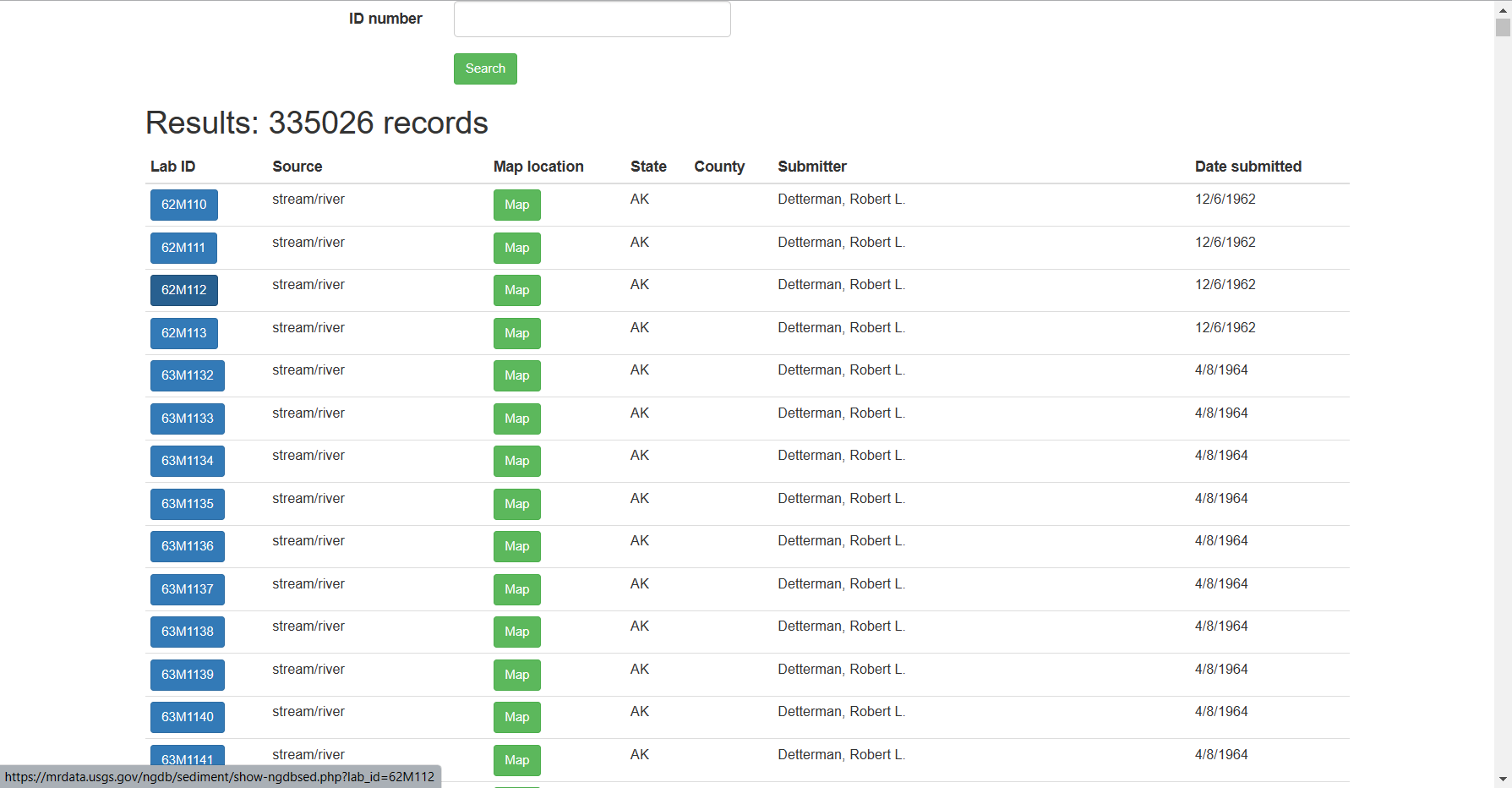}
    \caption{NGDB collections.}
    \label{fig:ngdb1}
\end{figure}

The webpage provides a list of URLs containing information about all the sedimentary rocks.~\autoref{fig:ngdb1}

\begin{figure}[H]
    \centering
    \includegraphics[width=0.7\linewidth]{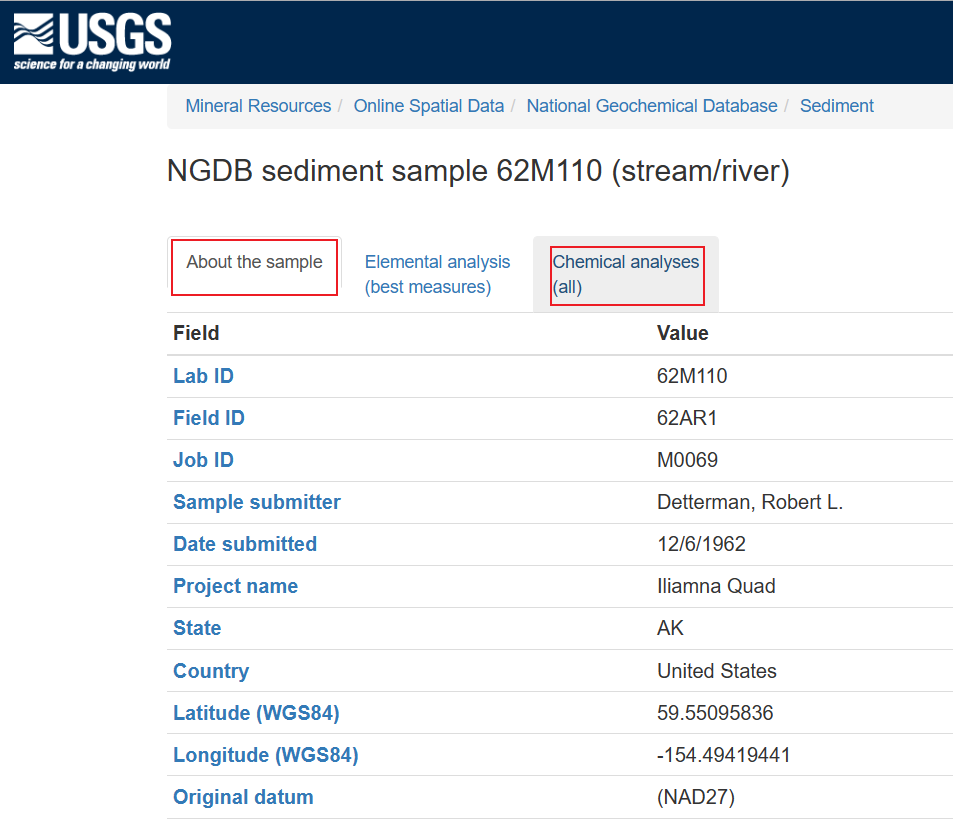}
    \caption{Mines in NGDB.}
    \label{fig:ngdb2}
\end{figure}

\begin{figure}[H]
    \centering
    \includegraphics[width=0.7\linewidth]{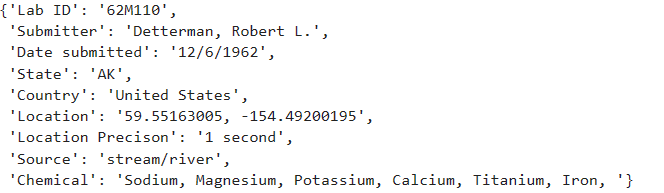}
    \caption{NGDB final signals.}
    \label{fig:ngdb3}
\end{figure}

{\bf Information Acquisition}:

For each type of sedimentary rock, corresponding information needs to be extracted. The extracted information for each type of sedimentary rock will be combined into one dictionary, and each dictionary will be a list for each type of sedimentary rock. In practice, it is possible that some sedimentary rocks may lack information. In this case, a corresponding dictionary should still be created, with the value set to "No corresponding information." The final data will be stored in a JSON file. The information that needs to be extracted for each type of sedimentary rock is as follows, taking 62M110 as an example.

The information needs to be extracted from "About the sample" and "Chemical analysis" sections (marked in red in the figure).

\begin{itemize}[leftmargin=1.2em]
    \item \textbf{About the sample.} The following information needs to be extracted from this section: Lab ID, Submitter, Date submitted, State, Country, Location (this part is obtained by synthesizing the data from Original Latitude and Original Longitude), Location Precision, and Source.
\end{itemize}

\begin{itemize}[leftmargin=1.2em]
\item \textbf{Chemical analysis} This section requires obtaining the major fossil elements of the sedimentary rock, by only recording chemical elements with a composition greater than or equal to 1\%.

\end{itemize}
Finally, we obtain the signals in NGDB, with details on the nature of these signals and the processing steps involved.

\subsection{Fossil Ontology}

\begin{figure}[H]
    \centering
    \includegraphics[width=0.99\linewidth]{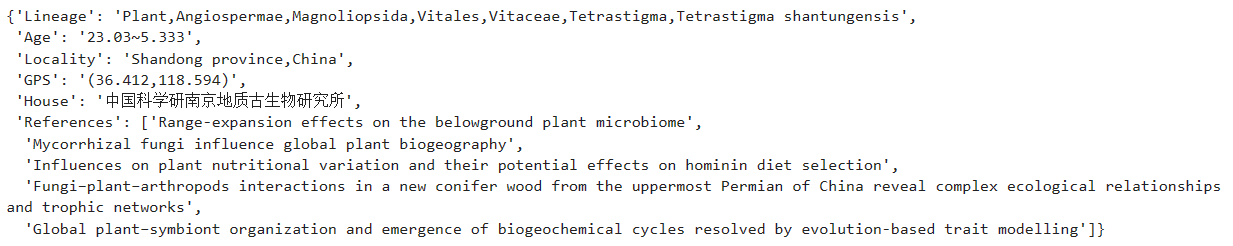}
    \caption{Fossil ontology signals.}
    \label{fig:fo1}
\end{figure}

The webpage displays a list of all fossils, so the list can be directly organized. The final data format will be a list, where each element in the list is a dictionary representing a type of fossil, and the data will be saved in a JSON file. Taking Tetrastigma shantungensis as an example, below shows the data structure:

\begin{itemize}[leftmargin=1.2em]
    \item \textbf{Basic Information of the Fossil.} Basic Information of the Fossil This part only requires the information of the biological lineage to which the fossil belongs, with key value "Lineage".
    \item \textbf{Geological Information of the Fossil.} This part requires information about the age of the fossil, the location of the fossil, and GPS information. The key values are "Age", "Locality", and "GPS", respectively.
    \item \textbf{Relevant Research Information of the Fossil.} This part requires information about the collection points and reference sources of the fossil, with key values "House" and "References", respectively. Among them, there are many References, which are organized as a list as the value.
\end{itemize}

Finally, we obtain the signals in Fossil Ontology.

\subsection{Fossil calibrations}

Starting from the aforementioned URLs, all the specific webpage information about 220 fossil types is included. After grabbing the HTML, please organize and save it accordingly.

To search, we use "find(class='listed-calibrations')", as shown in the following image:

\begin{figure}[H]
    \centering
    \includegraphics[width=0.95\linewidth]{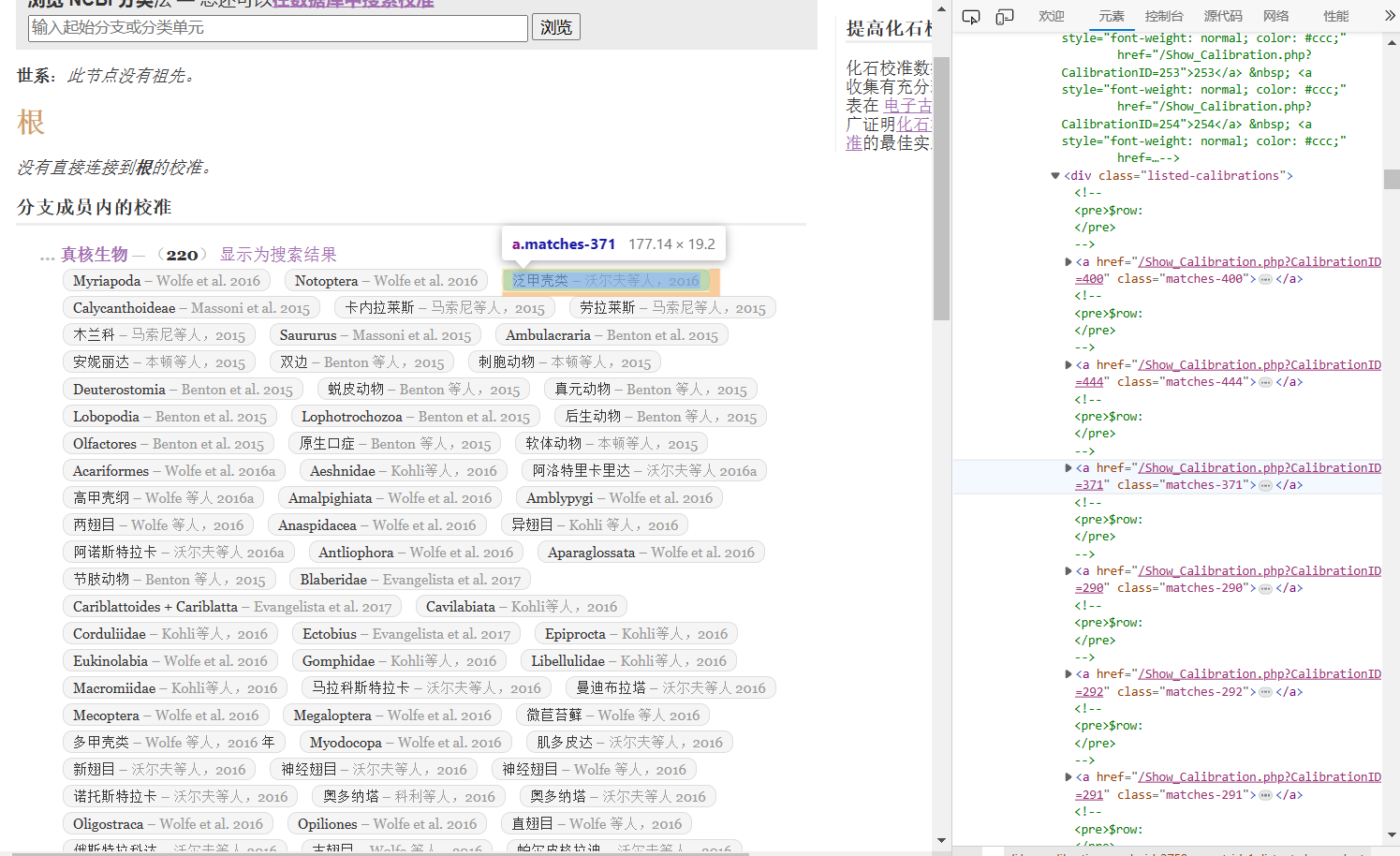}
    \caption{Fossil calibrations collections.}
    \label{fig:fc1}
\end{figure}

After performing the above search, it is possible to obtain the URLs for all the fossils, and then extract information about each of them one by one.

The final data format will be a list, where each element in the list is a dictionary representing a type of fossil, and the data will be saved in a JSON file. Taking the fossil Sessilia as an example, below shows the information to be extracted and the format:

\begin{itemize}[leftmargin=1.2em]
    \item \textbf{Basic Information.} This section contains two parts: the species to which the fossil belongs, and the biological lineage of the species. The key value for species name is "Name", and the value is the name of the species to which the fossil belongs; The key value for biological lineage is "Lineage", and the value is the biological lineage of the species.
    \item \textbf{Age Information.} This part includes the earliest and latest time of the fossil, with key values "Minimum age" and "Maximum age", respectively.
    \item \textbf{Key Fossil Information.} This part mainly concerns the key fossils used to determine the node period, including the location and geological age, with key values "Locality" and "Geological age" respectively.
    \item \textbf{References.} This part mainly includes the calibration and reference sources of the fossil information, with key values "Calibration" and "Reference", respectively.
\end{itemize}

Finally, we obtain the signals in Fossil Calibrations.

\begin{figure}[H]
    \centering
    \includegraphics[width=0.95\linewidth]{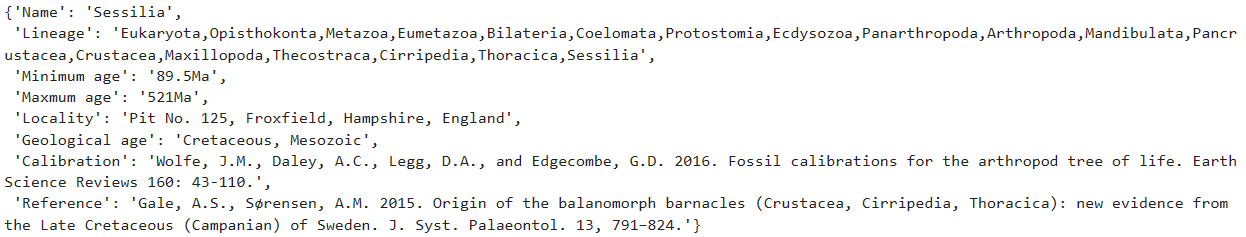}
    \caption{Fossil calibrations signals.}
    \label{fig:fc2}
\end{figure}
\newpage

\section{Appendix: Prompts}

Below, we will list the prompt words we use for constructing Instruction tuning data.

\definecolor{baby_blue}{RGB}{173, 216, 230}
\definecolor{baby_green}{RGB}{63, 135, 118}
\definecolor{dark_blue}{RGB}{31, 119, 180}
\definecolor{light_grey}{RGB}{211, 211, 211}

{\bf deepliterature.abstract.title} 
\begin{enumerate}
    \item {\bf \color{dark_blue} Input}: What is a suitable title for this geoscience paragraph: [input] \\
    {\bf \color{baby_green} Output}: The title can be '[output]'
    
    \item {\bf \color{dark_blue} Input}: Provide a fitting title for this geoscience text: [input]\\
    {\bf \color{baby_green} Output}: One possible title can be '[output]'
    
    \item {\bf \color{dark_blue} Input}: Suggest a title for this geoscience passage: [input]\\
    {\bf \color{baby_green} Output}: '[output]' can be a suitable title.
    
    \item {\bf \color{dark_blue} Input}: What would be an appropriate title for this geoscience paragraph: [input] \\
    {\bf \color{baby_green} Output}: Here is a title for you to consider: [output].

    \item {\bf \color{dark_blue} Input}: What should be the title of this geoscience text considering its content: [input]\\
    {\bf \color{baby_green} Output}: '[output]' is one option.
    
\end{enumerate}

{\bf deepliterature.abstract.keyword} 
\begin{enumerate}
    \item {\bf \color{dark_blue} Input}: Extract the main keywords from this geoscience paragraph: [input]\\
    {\bf \color{baby_green} Output}: The main keywords in the paragraph can be listed as follows: [output]
    
    \item {\bf \color{dark_blue} Input}: Identify the key terms in this geoscience passage: [input]\\
    {\bf \color{baby_green} Output}:The key terms in the passage can be identified as: [output]
    
    \item {\bf \color{dark_blue} Input}: What are the most important keywords in this geoscience text: [input]\\
    {\bf \color{baby_green} Output}: The most important keywords in the text are: [output]

    \item {\bf \color{dark_blue} Input}: What are the fields of study connected to this geoscience text: [input]\\
    {\bf \color{baby_green} Output}:The fields of study connected to this text include: [output]

    \item {\bf \color{dark_blue} Input}: Determine the fields of research related to this geoscience paragraph: [input]\\
    {\bf \color{baby_green} Output}: The related fields of research for this paragraph are: [output]
    
\end{enumerate}

{\bf deepliterature.reference.resolution} 
\begin{enumerate}
    \item {\bf \color{dark_blue} Input}: When you see the sentence like [input], can you recommend some related papers for me?\\
    {\bf \color{baby_green} Output}: Alright, I have found a relevant paper for you: {\color{light_grey} \textbackslash n}[output]
    
    \item {\bf \color{dark_blue} Input}: Based on the content of this sentence [input], could you suggest a few relevant papers?\\
    {\bf \color{baby_green} Output}: Okay, I've got the related paper right here: {\color{light_grey} \textbackslash n}[output]

    \item {\bf \color{dark_blue} Input}: Considering the information in this sentence [input], what papers would you recommend?\\
    {\bf \color{baby_green} Output}: Sure, here's the associated paper you're looking for: {\color{light_grey} \textbackslash n}[output]

    \item {\bf \color{dark_blue} Input}: Given the context of this sentence [input], can you provide a list of related papers?\\
    {\bf \color{baby_green} Output}: No problem, I have the corresponding paper for you: {\color{light_grey} \textbackslash n}[output]

    \item {\bf \color{dark_blue} Input}: After reading this sentence [input], what are some papers you would recommend?\\
    {\bf \color{baby_green} Output}: Certainly, I'm showing you the related paper now: {\color{light_grey} \textbackslash n}[output]
\end{enumerate}

{\bf gso.wikipedia.title} 
\begin{enumerate}
    \item {\bf \color{dark_blue} Input}: Can you explain the concept of [title] in relation to [keyword]?\\
    {\bf \color{baby_green} Output}: Certainly, when discussing [title] in the context of [keyword], [paragraph]
    
    \item {\bf \color{dark_blue} Input}: Please share information about [title] concerning [keyword].\\
    {\bf \color{baby_green} Output}: Of course, regarding [title] and its connection to [keyword], [paragraph]

    \item {\bf \color{dark_blue} Input}: What does the [title] section cover about the topic [keyword]?\\
    {\bf \color{baby_green} Output}: In the [title] section about [keyword], we have: [paragraph]

    \item {\bf \color{dark_blue} Input}: Please describe [title] as it pertains to [keyword].\\
    {\bf \color{baby_green} Output}: Certainly, in terms of [title] and its relationship to [keyword], [paragraph]

    \item {\bf \color{dark_blue} Input}: Can you elaborate on [title] in connection with [keyword]?\\
    {\bf \color{baby_green} Output}: Of course, when discussing [title] and its connection to [keyword], [paragraph]
\end{enumerate}

{\bf gso.wikipedia.entity} 
\begin{enumerate}
    \item {\bf \color{dark_blue} Input}: Can you identify the key terms related to earth science mentioned in this passage? The passage: [input]\\
    {\bf \color{baby_green} Output}: Sure, the terms related to earth science mentioned in the passage include: [output]
    
    \item {\bf \color{dark_blue} Input}: What are the terms connected to earth science present in this passage? The passage: [input]\\
    {\bf \color{baby_green} Output}: The terms connected to earth science in the passage are: [output]

    \item {\bf \color{dark_blue} Input}: Please list the terms associated with earth science found in the given passage. The passage: [input]\\
    {\bf \color{baby_green} Output}: The terms associated with earth science in the passage are as follows: [output]

    \item {\bf \color{dark_blue} Input}: Can you help me find the terms relevant to earth science in this passage? The passage: [input]\\
    {\bf \color{baby_green} Output}: Sure, the terms relevant to earth science in the passage are: [output]

    \item {\bf \color{dark_blue} Input}: Identify the terms in the passage that are related to earth science. The passage: [input]\\
    {\bf \color{baby_green} Output}: The terms in the passage related to earth science are: [output]
\end{enumerate}

{\bf gso.wordnet.description}
\begin{enumerate}
    \item {\bf \color{dark_blue} Input}: Can you provide a description for the word '[input]'?\\
    {\bf \color{baby_green} Output}: Sure, the description for the word '[input]' is: [output] 
    
    \item {\bf \color{dark_blue} Input}: What does the word '[input]' mean?\\
    {\bf \color{baby_green} Output}: The meaning of the word '[input]' is: [output]

    \item {\bf \color{dark_blue} Input}: Please explain the meaning of the term '[input]'.\\
    {\bf \color{baby_green} Output}: The term '[input]' means: [output]

    \item {\bf \color{dark_blue} Input}: What can you tell me about the meaning of the word '[input]'?\\
    {\bf \color{baby_green} Output}: The word '[input]' has the following meaning: [output]

    \item {\bf \color{dark_blue} Input}: Help me understand the definition of the term '[input]'.\\
    {\bf \color{baby_green} Output}: The definition of the term '[input]' is: [output]
\end{enumerate}

{\bf gso.wordnet.synonym}
\begin{enumerate}
    \item {\bf \color{dark_blue} Input}: Can you provide a synonym for the word '[input]'?\\
    {\bf \color{baby_green} Output}: Sure, a synonym for the word '[input]' is: [output]
    
    \item {\bf \color{dark_blue} Input}: What is a synonym of the word '[input]'?\\
    {\bf \color{baby_green} Output}: A synonym of the word '[input]' is: [output]

    \item {\bf \color{dark_blue} Input}: Please give me a synonym for the term '[input]'.\\
    {\bf \color{baby_green} Output}: A synonym for the term '[input]' is: [output]

    \item {\bf \color{dark_blue} Input}: What would be a synonym for the word '[input]'?\\
    {\bf \color{baby_green} Output}: A synonym for the word '[input]' is: [output]

    \item {\bf \color{dark_blue} Input}: Help me find a synonym for the term '[input]'.\\
    {\bf \color{baby_green} Output}: A synonym for the term '[input]' is: [output]
\end{enumerate}

{\bf gso.dictionary.definition}
\begin{enumerate}
    \item {\bf \color{dark_blue} Input}: Can you provide the specialized definition of the term '[input]'?\\
    {\bf \color{baby_green} Output}: Sure, in the context of geoscience, the term '[input]' is defined as: [output]
    
    \item {\bf \color{dark_blue} Input}: What does the term '[input]' mean in a specialized context?\\
    {\bf \color{baby_green} Output}: In the field of geoscience, the term '[input]' means: [output]

    \item {\bf \color{dark_blue} Input}: Please explain the meaning of the specialized term '[input]'.\\
    {\bf \color{baby_green} Output}: The specialized term '[input]', in the context of geoscience, means: [output]

    \item {\bf \color{dark_blue} Input}: Can you suggest the meaning of the term '[input]' in a specific field?\\
    {\bf \color{baby_green} Output}: In the field of geoscience, the meaning of the term '[input]' is: [output]

    \item {\bf \color{dark_blue} Input}: What would be the definition of the word '[input]' in a specialized context?\\
    {\bf \color{baby_green} Output}: In the context of geoscience, the definition of the word '[input]' is: [output]
\end{enumerate}

{\bf gso.dictionary.synonym}
\begin{enumerate}
    \item {\bf \color{dark_blue} Input}: Can you provide a synonym for the word '[input]'?\\
    {\bf \color{baby_green} Output}: Sure, you might [output]
    
    \item {\bf \color{dark_blue} Input}: What is a synonym of the word '[input]'?\\
    {\bf \color{baby_green} Output}: You can [output]

    \item {\bf \color{dark_blue} Input}: Please give me a synonym for the term '[input]'.\\
    {\bf \color{baby_green} Output}: Of course, you should [output]

    \item {\bf \color{dark_blue} Input}: What would be a synonym for the word '[input]'?\\
    {\bf \color{baby_green} Output}: One can [output]

    \item {\bf \color{dark_blue} Input}: What's a synonym for the word '[input]'?\\
    {\bf \color{baby_green} Output}: You can [output]
\end{enumerate}

{\bf gso.dictionary.classification}
\begin{enumerate}
    \item {\bf \color{dark_blue} Input}: Identify the scientific discipline associated with the term '[input]':\\
    {\bf \color{baby_green} Output}: [output]
    
    \item {\bf \color{dark_blue} Input}: In which scientific discipline would you find the term '[input]'?\\
    {\bf \color{baby_green} Output}: [output]

    \item {\bf \color{dark_blue} Input}: The term '[input]' is primarily associated with which of these disciplines?\\
    {\bf \color{baby_green} Output}: [output]

    \item {\bf \color{dark_blue} Input}: In the realm of Earth sciences, which discipline is most closely linked to the term '[input]'?\\
    {\bf \color{baby_green} Output}: [output]

    \item {\bf \color{dark_blue} Input}: Which scientific discipline is most relevant when discussing the term '[input]'?\\
    {\bf \color{baby_green} Output}: [output]
\end{enumerate}

{\bf gso.taxonomy.hyponymy.child}
\begin{enumerate}
    \item {\bf \color{dark_blue} Input}: What are some subfields or specific topics that fall under the broader concept of [parent]?\\
    {\bf \color{baby_green} Output}: Some subfields or specific topics that fall under the broader concept of [parent] in geoscience include [child].
    
    \item {\bf \color{dark_blue} Input}: What are some specific examples of [parent] within geoscience?\\
    {\bf \color{baby_green} Output}: Some specific examples of [parent] within geoscience include [child].

    \item {\bf \color{dark_blue} Input}: What are some narrower categories or subdisciplines within [parent]?\\
    {\bf \color{baby_green} Output}: Some narrower categories or subdisciplines within [parent] in geoscience include [child].

    \item {\bf \color{dark_blue} Input}: What are some specific types or varieties of [parent] in geoscience?\\
    {\bf \color{baby_green} Output}: Some specific types or varieties of [parent] in geoscience include [child].

    \item {\bf \color{dark_blue} Input}: What are some specific techniques or methodologies used to study [parent] in geoscience?\\
    {\bf \color{baby_green} Output}: Some specific techniques or methodologies used to study [parent] in geoscience include [child].
\end{enumerate}

{\bf gso.taxonomy.hyponymy.parent}
\begin{enumerate}
    \item {\bf \color{dark_blue} Input}: What is the overarching category that [child] belongs to?\\
    {\bf \color{baby_green} Output}: It can be classified under [parent], which is a broad field of study within geoscience.
    
    \item {\bf \color{dark_blue} Input}: What are the subfields that fall under [parent]?\\
    {\bf \color{baby_green} Output}: Some subfields that are part of [parent] include [child], among others.

    \item {\bf \color{dark_blue} Input}: What are some related concepts to [parent] in geoscience?\\
    {\bf \color{baby_green} Output}: Some related concepts to [parent] in geoscience are [child], among others.

    \item {\bf \color{dark_blue} Input}: What is the specific category that [child] belongs to within geoscience?\\
    {\bf \color{baby_green} Output}: The specific category that [child] belongs to within geoscience is [parent].

    \item {\bf \color{dark_blue} Input}: What are the different branches of geoscience that [child] is part of?\\
    {\bf \color{baby_green} Output}: Some branches of geoscience that [child] is part of include [parent], among others.
\end{enumerate}

{\bf gso.taxonomy.hyponymy.judgment.parent}
\begin{enumerate}
    \item {\bf \color{dark_blue} Input}: Is [child] a child concept of [parent] in geoscience?\\
    {\bf \color{baby_green} Output}: [Answer]
    
    \item {\bf \color{dark_blue} Input}: Does [child] fall under the broader concept of [parent] in geoscience?\\
    {\bf \color{baby_green} Output}: [Answer]

    \item {\bf \color{dark_blue} Input}: Is [child] a narrower category or subdiscipline within [parent] in geoscience?\\
    {\bf \color{baby_green} Output}: [Answer]

    \item {\bf \color{dark_blue} Input}: Are [child] specific types or varieties of [parent] in geoscience?\\
    {\bf \color{baby_green} Output}: [Answer]

    \item {\bf \color{dark_blue} Input}: Do scientists use [child] as a technique or methodology to study [parent] in geoscience?\\
    {\bf \color{baby_green} Output}: [Answer]
\end{enumerate}

{\bf gso.taxonomy.hyponymy.judgment.child}
\begin{enumerate}
    \item {\bf \color{dark_blue} Input}: Is [parent] the broader concept that encompasses [output] in geoscience?\\
    {\bf \color{baby_green} Output}: [Answer]
    
    \item {\bf \color{dark_blue} Input}: Does [parent] serve as the parent category for [output] in geoscience?\\
    {\bf \color{baby_green} Output}: [Answer]

    \item {\bf \color{dark_blue} Input}: Is [parent] a higher-level concept that includes [output] in geoscience?\\
    {\bf \color{baby_green} Output}: [Answer]

    \item {\bf \color{dark_blue} Input}: Does [parent] have a broader field of study than [child] within geoscience?\\
    {\bf \color{baby_green} Output}: [Answer]

    \item {\bf \color{dark_blue} Input}: Is [parent] a common parent concept of [child] within geoscience?\\
    {\bf \color{baby_green} Output}: [Answer]
\end{enumerate}

{\bf metaearth}
\begin{enumerate}
    \item {\bf \color{dark_blue} Input}: What's the [key] of the [object]?\\
    {\bf \color{baby_green} Output}: The [object]'s [key] is [value].
    
    \item {\bf \color{dark_blue} Input}: Can you tell me the [key] of the [object]?\\
    {\bf \color{baby_green} Output}: Sure, the [object]'s [key] is [value].

    \item {\bf \color{dark_blue} Input}: What is the [object]'s [key]?\\
    {\bf \color{baby_green} Output}: The [key] of the [object] is [value].

    \item {\bf \color{dark_blue} Input}: What would be the [key] of the [object]?\\
    {\bf \color{baby_green} Output}: The [object]'s [key] is [value].

    \item {\bf \color{dark_blue} Input}: I'd like to know the [key] of the [object].\\
    {\bf \color{baby_green} Output}: The [object] has a [key] of [value].
\end{enumerate}

{\bf gakg.qa}
\begin{enumerate}
    \item {\bf \color{dark_blue} Input}: What's the [key] of the paper according to the abstract: {\color{light_grey} \textbackslash n}[object]?\\
    {\bf \color{baby_green} Output}: The paper's [key] is [value].
    
    \item {\bf \color{dark_blue} Input}: Can you tell me the [key] of the paper according to the abstract: {\color{light_grey}\textbackslash n}[object]?\\
    {\bf \color{baby_green} Output}: Sure, the paper's [key] is [value].

    \item {\bf \color{dark_blue} Input}: What is the paper's [key]? According to the abstract: {\color{light_grey} \textbackslash n}[object]\\
    {\bf \color{baby_green} Output}: The [key] of the paper is [value].

    \item {\bf \color{dark_blue} Input}: Please inform me of the paper's [key]. According to the abstract: {\color{light_grey} \textbackslash n}[object]\\
    {\bf \color{baby_green} Output}: The [key] associated with the paper is [value].

    \item {\bf \color{dark_blue} Input}: I'd like to know the [key] of the paper According to the abstract: {\color{light_grey} \textbackslash n}[object].\\
    {\bf \color{baby_green} Output}: The paper has a [key] of [value].
\end{enumerate}

{\bf ner}
\begin{enumerate}
    \item {\bf \color{dark_blue} Input}: Please extract the named entities in: [input].\\
    {\bf \color{baby_green} Output}: The named entities in the passage can be listed as follows: {\color{light_grey} \textbackslash n}[output]
    
    \item {\bf \color{dark_blue} Input}: What are the named entities mentioned in: [input]?\\
    {\bf \color{baby_green} Output}: The named entities mentioned in the passage are: {\color{light_grey} \textbackslash n}[output]

    \item {\bf \color{dark_blue} Input}: Can you identify the named entities in: [input]?\\
    {\bf \color{baby_green} Output}: Sure, the named entities in the passage are: {\color{light_grey} \textbackslash n}[output]

    \item {\bf \color{dark_blue} Input}: What are the geological terms or concepts mentioned in: [input]?\\
    {\bf \color{baby_green} Output}: The geological terms or concepts mentioned in the passage are: {\color{light_grey} \textbackslash n} [output]

    \item {\bf \color{dark_blue} Input}: What are the names of any geological formations mentioned in: [input]?\\
    {\bf \color{baby_green} Output}: The names of the geological formations mentioned in the passage are: {\color{light_grey} \textbackslash n} [output]
\end{enumerate}

{\bf gakg.illustration}
\begin{enumerate}
    \item {\bf \color{dark_blue} Input}: What kind of illustration when you see these content? [input]\\
    {\bf \color{baby_green} Output}: [output]
    
    \item {\bf \color{dark_blue} Input}: What type of illustration comes to mind when you observe the following content? [input]\\
    {\bf \color{baby_green} Output}: [output]

    \item {\bf \color{dark_blue} Input}: "What sort of illustration is associated with this content? [input]\\
    {\bf \color{baby_green} Output}: [output]

    \item {\bf \color{dark_blue} Input}: What kind of illustration can you imagine when presented with this content? [input]\\
    {\bf \color{baby_green} Output}: [output]

    \item {\bf \color{dark_blue} Input}: When examining the following content, what type of illustration would you associate with it? [input]\\
    {\bf \color{baby_green} Output}: [output]
\end{enumerate}

{\bf gakg.table}
\begin{enumerate}
    \item {\bf \color{dark_blue} Input}: What kind of table when you see these elements? [input]\\
    {\bf \color{baby_green} Output}: [output]
    
    \item {\bf \color{dark_blue} Input}: What type of table comes to mind when you observe the following content? [input]\\
    {\bf \color{baby_green} Output}: [output]

    \item {\bf \color{dark_blue} Input}: What sort of table is associated with this elements? [input]\\
    {\bf \color{baby_green} Output}: [output]

    \item {\bf \color{dark_blue} Input}: What kind of table can you imagine when presented with this content? [input]\\
    {\bf \color{baby_green} Output}: [output]

    \item {\bf \color{dark_blue} Input}: Upon seeing the content, what kind of table do you think of? [input]\\
    {\bf \color{baby_green} Output}: [output]
\end{enumerate}
\newpage
\section{Appendix: Training setup}
\label{hparam}
\begin{itemize}[leftmargin=1.2em]
    \item Model setup (30B parameters)
    \begin{itemize}[leftmargin=1.2em]
        \item num layers: 48
        \item num attention heads: 56
        \item hidden size: 7168
        \item max position embeddings: 2048
        \item layernorm epsilon: $1e-5$
    \end{itemize}
    \item Regularization Setup
    \begin{itemize}[leftmargin=1.2em]
        \item optimizer: Adam
        \item attention dropout: 0.1
        \item hidden dropout: 0.1
        \item weight decay: 0.1
        \item clip-grad: 1.0
        \item adam $\beta_{1}$: 0.9
        \item adam $\beta_{2}$: 0.95
        \item adam $\epsilon$: $1e-8$
    \end{itemize}
    \item Training Setup
    \begin{itemize}[leftmargin=1.2em]
        \item micro-batch-size: 1
        \item global-batch-size: 4096
        \item recompute-activations: True (gradient checkpointing)
        \item train-samples: 30M (60B token)
        \item disable-bias-linear: True (turn off the bias of \textit{nn.linear})
        \item seed: 42
        \item save-interval: 100
    \end{itemize}
    \item Learning Rate Setup
    \begin{itemize}[leftmargin=1.2em]
        \item lr-decay-style: linear
        \item lr-warmup-steps: 100
        \item lr: $1e-5$
        \item min-lr: $1e-7$
    \end{itemize}
    \item Mixed Precision Setup
    \begin{itemize}[leftmargin=1.2em]
        \item FP16: False
        \item BF16: False
    \end{itemize}
    \item Parallel Configuration
    \begin{itemize}[leftmargin=1.2em]
        \item tensor-model-parallel-size: 4
        \item pipeline-model-parallel-size: 16
        \item distributed-backend: NCCL
        \item sequence-parallel: True
    \end{itemize}
\end{itemize}
\textbf{Notices:} With a model parallel size (TP) of 4 and a pipeline parallel size (PP) of 16, it can be considered that a 30B model with 48 layers is divided into 16 parts with 3 layers in each part,  and each of the 4 accelerator cards in a node is responsible for processing 3 continuous layer. 
\newpage
\section{Appendix: Model Card}
\label{modelcard}
Our model is based on Galactica, a standard GPT2 structure with 48 transformer blocks and 56 attention heads per layer, with a hidden dim of 7,168. The parameters and tokenizer of our model are initialized from the hugging face release checkpoint of Galactica-30B, and the maximum input length is 2048 and no bias settings are followed. Although Galactica-30B is an fp16 model, to ensure the stability of the training process, we used fp32 to train and save model parameters.

\begin{figure}[H]
\raggedright
\begin{framed}
\begin{center}{\large {\bf Model Card - \geolm{}}}\end{center} 
{\bf Model Details} 
\begin{itemize}[leftmargin=*]
\item Developed by: Shanghai Jiao Tong University and Deep-time Digital Earth Science Center.
\item Shared by: Shanghai Jiao Tong University and \href{https://www.geobrain-ai.com}{GeoBRAIN.ai}.
\item Model type: Further pre-train and Supervised Fine-tuning.
\item Language(s) (NLP): English.
\item License: Apache License 2.0.
\item Further pre-train from model: Galactica~\cite{Taylor2022GalacticaAL}.
\end{itemize}

{\bf Model Sources}
\begin{itemize}[leftmargin=*]
    \item Repository: \url{https://github.com/geobrain-ai/geogalactica}
    \item Paper: \geolm{}: A Scientific Large Language Model in Geoscience
\end{itemize}

{\bf Intended Use}
\begin{itemize}[leftmargin=*]
\item Research Assistance: Providing support in academic and industrial research by summarizing current scientific literature, suggesting hypotheses, and identifying gaps in existing research.
\item Educational Tool: Serving as an educational resource for students and professionals in geosciences, offering explanations of complex concepts and providing interactive learning experiences.
\item Collaboration and Communication: Facilitating collaboration among geoscientists by providing a platform for sharing data and insights, and helping in communicating complex geoscientific information to non-experts.
\end{itemize}

{\bf Ethical Considerations}
\begin{itemize}[leftmargin=*]
\item This model inherits from Galactica~\cite{Taylor2022GalacticaAL}, and in the training corpus, we have conducted sufficient data governance to ensure that the training data embodies geographical community, transparency, inclusiveness, respect for privacy, and topic neutrality.
\end{itemize}

{\bf Training Data} 
\begin{itemize}[leftmargin=*]
\item Further pre-train: A geoscience-related text corpus containing \textbf{65 billion tokens}, preserving as the largest geoscience-specific text corpus.
\item Supervised Fine-tuning: \href{https://huggingface.co/datasets/daven3/geosignal}{daven3/geosignal}.
\item Tool-Augmented Learning: \href{https://github.com/zthang/geotools}{zthang/geotools}.

\end{itemize}

{\bf Evaluation Data} 
\begin{itemize}[leftmargin=*]
\item MMLU: Massive Multitask Language Understanding, a large-scale research initiative aimed at improving language models' understanding and reasoning abilities across a diverse range of subjects and tasks.
\item GeoBench: The benchmark mentioned in K2~\cite{Deng2023LearningAF}. The data can be access on [daven3/geobench](https://huggingface.co/datasets/daven3/geobench).
\item Human Evaluation: Selected questions.
\end{itemize}

{\bf Model Card Contact} \\
\geolm{} is a research preview intended for non-commercial use only. Please contact us if you find any issues. For details, you can email via \href{mailto:davendw@sjtu.edu.cn}{davendw@sjtu.edu.cn}.

\end{framed}
\caption{Model Card for \geolm{}.}
\end{figure}

\newpage
\section{Appendix: Evaluation}
\label{appendix:evaluation}
\subsection{Open-ended Tasks}
\subsubsection{Noun Definition}
\label{appendix:evaluation:noun}
For a better understanding of the words, we added the definitions from the Geology Dictionary.
\begin{itemize}[leftmargin=1.2em]
\item[1] Physical weathering and chemical weathering are processes that break down rocks and minerals through mechanical and chemical means, respectively.
\item[2] Sedimentary differentiation (\textit{the sorting and separation of sediments during the formation of sedimentary rocks}).
\item[3] A continental margin (\textit{the boundary between a continent and an ocean, characterized by various geological features}).
\item[4] Seafloor spreading (\textit{the process where new oceanic crust is formed at mid-ocean ridges as tectonic plates move apart}).
\item[5] Stratum occurrence (\textit{relates to the presence and distribution of rock layers in a particular geological context}).
\item[6] Normal fault (\textit{a type of fault where the hanging wall moves downward relative to the footwall}).
\item[7] Plate tectonics (\textit{the theory that describes the movement and interaction of Earth's lithospheric plates}).
\item[8] Continental margin (\textit{the boundary between a continent and an ocean, characterized by various geological features}).
\item[9] The Yanshan Movement (\textit{refers to a tectonic event in China that resulted in significant geological changes}).
\item[10] Continental margin (\textit{the boundary between a continent and an ocean, characterized by various geological features}).
\item[11] A united paleocontinent (\textit{a reconstructed ancient landmass formed by merging the continents' positions from the distant past}).
\item[12] Earth resources (\textit{natural materials and substances that are valuable to humans for various purposes}).
\item[13] Reverse fault (\textit{a type of fault where the hanging wall moves upward relative to the footwall}).
\item[14] Ediacara fauna (\textit{refers to a group of early, soft-bodied, and mostly extinct organisms from the Ediacaran Period}).
\item[15] Fingerfacies fossil (\textit{represents a specific type of fossil that provides information about sedimentary environments and conditions}).
\item[16] Walther's Law (\textit{describes the vertical succession of sedimentary rock layers that were originally deposited in lateral proximity}).
\item[17] Vertical accumulation (\textit{is the process of sediment layers accumulating on top of each other over time}).
\item[18] The original levelness principle (\textit{suggests that sedimentary layers were initially deposited horizontally}).
\item[19] The stratum overlap principle (\textit{relates to the idea that younger sedimentary layers can cover or overlap older ones}).
\item[20] Lateral accumulation (\textit{refers to the process of sedimentary material accumulating horizontally, typically in a depositional environment}).
\end{itemize}

\subsubsection{Beginner Level Q\&A}
\label{appendix:evaluation:bqa}
\begin{itemize}[leftmargin=1.2em]
\item[1] How does a cloud fill up with water?
\item[2] How does diffraction make a tree's shadow blurry?
\item[3] How does trash in the ocean disappear?
\item[4] How does water dowsing work?
\item[5] How does wind create all the ocean currents?
\item[6] If I jump, will the entire earth move a little bit?
\item[7] If I were able to dig a hole from the U.S. through the center of the earth, what part of China would I end up in?
\item[8] Is a quadruple rainbow possible?
\item[9] What causes the water going down a drain to swirl clockwise in the northern hemisphere and counter-clockwise in the southern hemisphere?
\item[10] What keeps the continents floating on a sea of molten rock?
\end{itemize}

\subsubsection{Intermediate Level Q\&A}
\label{appendix:evaluation:iqa}
\begin{itemize}[leftmargin=1.2em]
\item[1] How does the movement of tectonic plates contribute to the formation of earthquakes and volcanic activity?
\item[2] What are the main factors that influence the formation and intensity of hurricanes in the Atlantic Ocean?
\item[3] How does the process of erosion shape the landscape and contribute to the formation of features such as canyons and valleys? 
\item[4] What are the primary mechanisms responsible for the formation and movement of glaciers?
\item[5] How do ocean currents influence the distribution of marine organisms and impact the productivity of marine ecosystems?
\item[6] What factors contribute to the formation and intensity of tornadoes in regions prone to severe weather events?
\item[7] How do geological processes such as weathering and sedimentation contribute to the formation and transformation of soil?
\item[8] What role does the Earth's magnetic field play in protecting the planet from harmful solar radiation?
\item[9] How do variations in atmospheric pressure and temperature contribute to the formation and behavior of weather systems?
\item[10] What are the primary processes responsible for the formation and transformation of different types of rocks, such as igneous, sedimentary, and metamorphic rocks?
\end{itemize}

\subsubsection{Advanced Level Q\&A}
\label{appendix:evaluation:aqa}
\begin{itemize}[leftmargin=1.2em]
\item[1] How did Earth and other planets form? Were planets formed in situ?
\item[2] Was there ever a collision of the Earth with another planet Theia, giving birth to our satellite?
\item[3] What is the long-term heat balance of Earth?
\item[4] What made plate tectonics a dominant process only on Earth?
\item[5] How inherent to planetary evolution is the development of life conditions?
\item[6] As planets age and cool off, their internal and surface processes coevolve, chemically and mechanically, shaping the atmospheric composition. What are the chemical composition and mechanical properties of rocks in the Earth's mantle at the extreme pressure and temperature they undergo?
\item[7] What are the dynamic processes in the Earth's interior that accommodate and fuel plate tectonics?
\item[8] How does the geomagnetic field link to the iron convection properties at the deep Earth?
\item[9] Are intraplate hotspots made by deep sources of uprising materials (mantle plumes) coming from the deepest Earth's mantle?"
\end{itemize}

\subsection{Functional Tasks}

\subsubsection{Knowledge-based associative judgment question.}
\label{appendix:evaluation:kqa}
\begin{itemize}[leftmargin=1.2em]
\item[1] What is the specific category that magnetostratigraphy belongs to within geoscience?
\item[2] What is the overarching category that magnetic polarity stratigraphy belongs to?
\item[3] What are the subfields that fall under magnetic polarity stratigraphy?
\item[4] What are some related concepts to geomagnetic polarity in geoscience?
\item[5] What are some related concepts to geomagnetic polarity in geoscience?
\item[6] What is the specific category that transitional polarity belongs to within geoscience?
\item[7] What are the subfields that fall under magnetic polarity stratigraphy?
\item[8] What are the subfields that fall under magnetostratigraphic polarity units?
\item[9] What are the subfields that fall under magnetostratigraphic polarity units?
\item[10] What are the subfields that fall under magnetostratigraphic polarity units?
\end{itemize}

\subsubsection{Research Paper Proposition Task.}
\label{appendix:evaluation:rqa}
Here are two examples, and we will release the whole data on \href{https://github.com/geobrain-ai/geobenchmark}{Github}:

\begin{itemize}[leftmargin=1.2em]
\item \textbf{Abstract:} The Wenchuan Earthquake on 12 May 2008 triggered a large number of geo-hazards including landslides, slope collapses and debris flows. Field investigations and remote-sensing interpretation identified 11,308 geo-hazards in 16 seriously damaged counties in Sichuan Province, southwest China. The paper reports an analysis of the distribution of these geo-hazards, particularly the earthquake-triggered landslides. Not surprisingly, the most significant geo-hazards were related to the main fault and on the hanging-wall side, although some occurred in deeply incised river gorges further away from the main rupture zone. Due to the high seismic intensity of the earthquake, most of the large landslides moved at high speed and for considerable distances.

{\bf Title: Analysis of the geo-hazards triggered by the 12 May 2008 Wenchuan Earthquake, China}

\item \textbf{Abstract:} The Modern-Era Retrospective Analysis for Research and Applications-2 (MERRA2) version of the Goddard Earth Observing System-5 (GEOS-5) atmospheric general circulation model (AGCM) is currently in use in the NASA Global Modeling and Assimilation Office (GMAO) at a wide range of resolutions for a variety of applications. Details of the changes in parameterizations after the version in the original MERRA reanalysis are presented here. Results of a series of atmosphere-only sensitivity studies are shown to demonstrate changes in simulated climate associated with specific changes in physical parameterizations, and the impact of the newly implemented resolution-aware behavior on simulations at different resolutions is demonstrated. The GEOS-5 AGCM presented here is the model used as part of the GMAO MERRA2 reanalysis, global mesoscale simulations at 10 km resolution through 1.5 km resolution, the real-time numerical weather prediction system, and for atmosphere-only, coupled ocean-atmosphere and coupled atmosphere-chemistry simulations. The seasonal mean climate of the MERRA2 version of the GEOS-5 AGCM represents a substantial improvement over the simulated climate of the MERRA version at all resolutions and for all applications. Fundamental improvements in simulated climate are associated with the increased re-evaporation of frozen precipitation and cloud condensate, resulting in a wetter atmosphere. Improvements in simulated climate are also shown to be attributable to changes in the background gravity wave drag, and to upgrades in the relationship between the ocean surface stress and the ocean roughness. The series of resolution-aware parameters related to the moist physics was shown to result in improvements at higher resolutions and result in AGCM simulations that exhibit seamless behavior across different resolutions and applications..

{\bf Title: GMD - Development of the GEOS-5 atmospheric general circulation model: evolution from MERRA to MERRA2}
\end{itemize}

\subsubsection{Geoscience Research Functionality}
\label{appendix:evaluation:gqa}
Here we share the five papers used to evaluate the geoscience research functionality of the LLMs in the citation form of \textbf{MLA}:

\begin{itemize}[leftmargin=1.2em]
\item[1] Zheng, Yadong, et al. "\textit{A challenge to the concept of slip-lines in extrusion tectonics.}" Geoscience Frontiers 2.1 (2011): 23-34.
\item[2] Fedo, Christopher M., H. Wayne Nesbitt, and Grant M. Young. "Unraveling the effects of potassium metasomatism in sedimentary rocks and paleosols, with implications for paleo weathering conditions and provenance." Geology 23.10 (1995): 921-924.
\item[3] Owens, Jeremy D., et al. "Iron isotope and trace metal records of iron cycling in the proto‐North Atlantic during the Cenomanian‐Turonian oceanic anoxic event (OAE‐2)." Paleoceanography 27.3 (2012).
\item[4] Percival, L. M. E., et al. "Determining the style and provenance of magmatic activity during the Early Aptian Oceanic Anoxic Event (OAE 1a)." Global and Planetary Change 200 (2021): 103461.
\item[5] Jenkyns, Hugh C. "Transient cooling episodes during Cretaceous Oceanic Anoxic Events with special reference to OAE 1a (Early Aptian)." Philosophical Transactions of the Royal Society A: Mathematical, Physical and Engineering Sciences 376.2130 (2018): 20170073.
\end{itemize}
\newpage
\section{Generation Examples}

\subsection{Noun Definition}
\label{gen_example:1}
We present an example with prompt ''What is physical weathering and chemical weathering?`` as follows:

\begin{mdframed}[linecolor=baby_blue, linewidth=2pt, roundcorner=10pt, fontcolor=black, 
                shadow=true, shadowsize=2pt, shadowcolor=light_grey]
\vspace{1ex}
\begin{itemize}[leftmargin=*]
\item {\bf ChatGPT: }

Physical weathering and chemical weathering are two processes that contribute to the breakdown and alteration of rocks and minerals on Earth's surface. Let's explore each of them:

1. Physical Weathering: 
Physical weathering, also known as mechanical weathering, refers to the disintegration of rocks and minerals into smaller fragments without changing their chemical composition. It is primarily caused by physical forces and environmental factors. 

2. Chemical Weathering: 
Chemical weathering involves the alteration or decomposition of rocks and minerals through chemical reactions. It occurs when minerals in rocks come into contact with water, atmospheric gases, or other substances that cause them to dissolve, decompose, or undergo chemical changes. 

It's important to note that physical and chemical weathering often occur simultaneously and can complement each other, accelerating the overall weathering process. Additionally, the type and rate of weathering are influenced by factors such as climate, rock composition, topography, and the presence of vegetation or organisms.

\item {\bf \geolm{}}

Physical weathering and chemical weathering are two of the main processes that cause rocks and minerals to break down and change over time.

Physical weathering refers to the breakdown of rocks and minerals due to the action of natural forces such as wind, water, ice, and gravity. These forces can cause rocks to crack, crumble, and disintegrate into smaller pieces. For example, when water flows over the surface of a rock, it can erode and wear away the rock's surface, causing it to break apart.

Chemical weathering, on the other hand, refers to the breakdown of rocks and minerals due to the action of chemical reactions. These reactions occur when water, air, and other chemicals interact with the rocks and minerals, causing them to dissolve, decompose, or react with each other. For example, when carbon dioxide in the air dissolves in water, it forms carbonic acid, which can dissolve calcium carbonate in limestone, causing it to break down into calcium and bicarbonate ions.

In summary, physical weathering refers to the breakdown of rocks and minerals due to the action of natural forces, while chemical weathering refers to the breakdown of rocks and minerals due to the action of chemical reactions. Both processes play an important role in shaping the Earth's landscape over time.

\end{itemize}
\centering
\end{mdframed}

\subsection{Beginner Level Q\&A}
\label{gen_example:2}

We present an example with prompt ''How does a cloud fill up with water?`` as follows:

\begin{mdframed}[linecolor=baby_blue, linewidth=2pt, roundcorner=10pt, fontcolor=black, 
                shadow=true, shadowsize=2pt, shadowcolor=light_grey]
\vspace{1ex}
\begin{itemize}[leftmargin=*]
\item {\bf ChatGPT: }

"A cloud forms and ""fills up"" with water through a process known as condensation. It involves the transformation of water vapor into liquid water droplets or ice crystals, which are the visible components of clouds.

Here's a simplified explanation of the process:

1. Evaporation: The sun's heat causes water from various sources like oceans, lakes, and rivers to evaporate, turning it into water vapor. This water vapor is essentially invisible.

2. Moisture in the air: As the water evaporates, it rises into the atmosphere and mixes with the surrounding air. The capacity of the air to hold water vapor depends on factors such as temperature and humidity.

3. Cooling and saturation: As the warm, moist air rises higher into the atmosphere, it encounters cooler temperatures. Cooler air cannot hold as much moisture as warm air, so the water vapor begins to condense into tiny water droplets or ice crystals around microscopic particles in the air, such as dust, salt, or pollution. This process is called saturation.

4. Cloud formation: The condensed water droplets or ice crystals cluster together to form visible clouds. Clouds consist of countless tiny droplets or ice crystals suspended in the air, which give them their characteristic appearance.

The process of cloud formation is influenced by various factors such as air temperature, humidity levels, and the presence of condensation nuclei (particles that water vapor can condense onto). Different types of clouds, such as cumulus, stratus, or cirrus, form under specific atmospheric conditions and have distinct appearances.

It's important to note that while clouds contain water, they don't ""fill up"" like a container. The amount of water vapor in the atmosphere continuously changes as a result of various meteorological processes, including evaporation, condensation, and precipitation. Clouds form, evolve, and dissipate as these processes occur within the atmosphere."

\item {\bf \geolm{}}

A cloud fills up with water when water vapor in the air cools and condenses into tiny droplets or ice crystals. This happens when the temperature of the air drops below the dew point or frost point, which is the temperature at which the air can hold the maximum amount of water vapor.

When the temperature drops, the water molecules in the air slow down and collide with each other more frequently. This causes the water vapor to condense into tiny droplets or ice crystals, forming clouds. The more water vapor in the air, the more likely it is that the temperature will drop low enough for condensation to occur.

Clouds can also form when warm, moist air rises and cools. As the air rises, it expands and cools, causing the water vapor in the air to condense into tiny droplets or ice crystals. This process is known as adiabatic cooling.

In summary, a cloud fills up with water when water vapor in the air cools and condenses into tiny droplets or ice crystals. This can happen when the temperature of the air drops below the dew point or frost point, or when warm, moist air rises and cools.

\end{itemize}
\centering
\end{mdframed}

\subsection{Intermediate Level Q\&A}
\label{gen_example:3}

We present an example with prompt ''How does the movement of tectonic plates contribute to the formation of earthquakes and volcanic activity?`` as follows:

\begin{mdframed}[linecolor=baby_blue, linewidth=2pt, roundcorner=10pt, fontcolor=black, 
                shadow=true, shadowsize=2pt, shadowcolor=light_grey]
\vspace{1ex}
\begin{itemize}[leftmargin=*]
\item {\bf ChatGPT: }

"The movement of tectonic plates is a key factor in the formation of earthquakes and volcanic activity. The Earth's lithosphere, which consists of several large and numerous smaller tectonic plates, is divided into rigid pieces that float and move on the underlying semi-fluid asthenosphere. There are three main types of plate boundaries where interactions between these plates occur: divergent boundaries, convergent boundaries, and transform boundaries.
At divergent boundaries, tectonic plates move away from each other, creating a gap or rift. This movement results in the upwelling of magma from the mantle, forming new crust. As the magma rises to the surface, it can lead to volcanic activity, with molten rock erupting to form volcanoes. However, the volcanic activity at divergent boundaries is generally less explosive compared to other types of volcanic activity.
At convergent boundaries, tectonic plates collide with each other. There are three types of convergent boundaries: oceanic-oceanic, oceanic-continental, and continental-continental. When an oceanic plate converges with another oceanic plate or a continental plate, the denser oceanic plate is usually subducted (forced beneath) into the mantle. The subduction process can generate intense pressure and friction, resulting in the release of seismic energy and causing earthquakes. Additionally, as the subducted plate sinks into the mantle, it can melt, generating magma that rises to the surface and forms volcanoes. These volcanoes can be explosive and often create volcanic arcs, such as the Pacific Ring of Fire.
At transform boundaries, tectonic plates slide past each other horizontally. The friction between the plates can cause them to become locked, preventing smooth movement. As the plates continue to push against each other, stress builds up along the locked boundary. Eventually, the stress overcomes the friction, leading to a sudden release of energy, resulting in earthquakes. Transform boundaries are primarily characterized by seismic activity rather than volcanic activity.
In summary, the movement of tectonic plates at divergent boundaries leads to volcanic activity, while at convergent boundaries, it results in both earthquakes and volcanic activity. Transform boundaries primarily contribute to seismic activity. These interactions between tectonic plates play a fundamental role in shaping Earth's surface and are responsible for the occurrence of earthquakes and the formation of volcanic features around the world."

\item {\bf \geolm{}}

"The movement of tectonic plates is the driving force behind the formation of earthquakes and volcanic activity on Earth. Tectonic plates are large, rigid sections of the Earth's crust that move slowly over the surface of the planet. The movement of these plates is caused by convection currents in the Earth's mantle, which is the layer beneath the crust and above the core.

When two tectonic plates move towards each other, they may collide, causing one plate to be forced under the other. This process is known as subduction, and it is responsible for the formation of many of the world's mountain ranges, such as the Himalayas and the Andes. When two plates collide, the pressure can build up until it is released in a sudden, violent event known as an earthquake.

On the other hand, when two plates move in opposite directions, one may be forced upwards, causing the Earth's crust to stretch and thin. This process is known as rifting, and it is responsible for the formation of many of the world's rift valleys, such as the Great Rift Valley in Africa. When the crust is stretched too far, it can break apart, leading to the formation of new ocean basins.

Volcanic activity is also closely linked to the movement of tectonic plates. When two plates move towards each other, the pressure can build up until it is released in a sudden, violent event known as an earthquake. This can cause the Earth's crust to crack."

\end{itemize}
\centering
\end{mdframed}

\subsection{Advanced Level Q\&A}
\label{gen_example:4}

We present an example with prompt ''How did Earth and other planets form? Were planets formed in situ?`` as follows:

\begin{mdframed}[linecolor=baby_blue, linewidth=2pt, roundcorner=10pt, fontcolor=black, 
                shadow=true, shadowsize=2pt, shadowcolor=light_grey]
\vspace{1ex}
\begin{itemize}[leftmargin=*]
\item {\bf ChatGPT: }

"The formation of planets depends on several factors, including the gravitational pull of nearby celestial bodies, the temperature and pressure of the surrounding environment, and the availability of necessary resources for planet formation.
Planets may form in situ (i.e., within their own gravitational field) or through the interaction with other celestial bodies. Some planets may form from dust particles in space, while others may form around existing celestial bodies such as comets or asteroids.
The process of planet formation involves the accumulation of material over time until it reaches a critical mass where it can no longer be held back by gravity. This critical mass determines the final size and composition of the planet.
Overall, the process of planet formation is complex and influenced by various factors."

\item {\bf \geolm{}}

"The formation of Earth and other planets is a complex process that is still not fully understood. However, there are several theories and models that attempt to explain the formation of our solar system and its planets.

One of the most widely accepted theories is the ""nebular hypothesis,"" which suggests that the solar system formed from a rotating cloud of gas and dust called the solar nebula. This cloud collapsed under its own gravity, forming the sun at its center. As the cloud collapsed, it spun faster and faster, forming a disk of gas and dust around the sun. This disk is what we now call the solar system, and it is from this disk that the planets formed.

One of the main challenges in understanding the formation of the solar system and its planets is the wide range of sizes and compositions of the planets. For example, Earth is a rocky planet with a solid surface, while Jupiter is a gas giant with a thick atmosphere. Scientists have developed several models to explain the formation of these different types of planets.

One of the most popular models for the formation of rocky planets like Earth is the ""core accretion model."" This model suggests that the first step in the formation of a rocky planet is the formation of a solid core. This core is made up of small solid particles, such as dust and rocks, that collide and stick together. As the core grows larger, it attracts more gas and dust from the solar nebula, forming a larger planet."

\end{itemize}
\centering
\end{mdframed}

\subsection{Knowledge-based associative judgment question}
\label{gen_example:5}

We present an example with prompt ''What are the subfields that fall under magnetic polarity stratigraphy?`` as follows:

\begin{mdframed}[linecolor=baby_blue, linewidth=2pt, roundcorner=10pt, fontcolor=black, 
                shadow=true, shadowsize=2pt, shadowcolor=light_grey]
\vspace{1ex}
\begin{itemize}[leftmargin=*]
\item {\bf ChatGPT: }

"There are four main subfields that fall under magnetic polarity stratigraphy:
Stratigraphic magnetic polarity (SMP): SMP looks at the relative orientation of rocks and sediments along a given layer. It determines the direction of flow of the rocks and sediments in that layer.
Tectonic magnetic polarity (TMP): TMP looks at the relationship between rocks and sediments in relation to the tectonic plate boundaries. It determines the relative movements of the plates and the rocks and sediments on either side.
Paleomagnetism (PM): PM looks at the magnetic susceptibility of rocks and sediments. It determines the type of rock or sediment and the strength of the magnetic signal.
Applied geomagnetism (AG): AG looks at the application of magnetic principles in different fields, such as geology, petrography, and mineralogy. It applies the principles of electromagnetic induction to solve problems in various fields.
Overall, magnetic polarity stratigraphy falls under three main subfields: stratigraphic magnetic polarity (SMP), tectonic magnetic polarity (TMP), and applied geomagnetism (AG). These subfields combine to provide a comprehensive approach to analyzing the magnetic properties of rocks and sediments, and to interpreting their relationships with the underlying tectonic plate boundaries."

\item {\bf \geolm{}}

Magnetic polarity stratigraphy, also known as magnetostratigraphy, is a subfield of geophysics and geology that uses the polarity of the Earth's magnetic field to date and correlate sedimentary rock layers. The Earth's magnetic field is generated by the movement of molten iron in the Earth's outer core, and the polarity of the magnetic field can be either positive or negative. The polarity of the Earth's magnetic field has changed over time, and these changes can be used to correlate sedimentary rock layers from one location to another.

There are several subfields that fall under magnetic polarity stratigraphy, including:

1. Magnetostratigraphy: This subfield uses the polarity of the Earth's magnetic field to date and correlate sedimentary rock layers.

2. Biostratigraphy: This subfield uses the distribution of fossils in sedimentary rock layers to date and correlate sedimentary rock layers.

3. Lithostratigraphy: This subfield uses the physical characteristics of sedimentary rock layers to date and correlate sedimentary rock layers.

4. Chemostratigraphy: This subfield uses the chemical composition of sedimentary rock layers to date and correlate sedimentary rock layers.

5. Magneto-biostratigraphy: This subfield combines magnetostratigraphy and biostratigraphy to date and correlate sedimentary rock layers.

6. Magneto-lithostratigraphy: This subfield combines magnetostratigraphy and lithostratigraphy to date and correlate sedimentary rock layers.

\end{itemize}
\centering
\end{mdframed}

\subsection{Research Paper Titling Task}
\label{gen_example:6}

We present an example with instruction ''Please help me to title the passages.`` ,and input \emph{''The Wenchuan Earthquake on 12 May 2008 triggered a large number of geo-hazards including landslides, slope collapses and debris flows. Field investigations and remote-sensing interpretation identified 11,308 geo-hazards in 16 seriously damaged counties in Sichuan Province, south west China. The paper reports an analysis of the distribution of these geo-hazards, particularly the earthquake-triggered landslides. Not surprisingly, the most significant geo-hazards were related to the main fault and on the hanging-wall side, although some occurred in deeply incised river gorges further away from the main rupture zone. Due to the high seismic intensity of the earthquake, most of the large landslides moved at high speed and for considerable distances.``} as follows:

\begin{mdframed}[linecolor=baby_blue, linewidth=2pt, roundcorner=10pt, fontcolor=black, 
                shadow=true, shadowsize=2pt, shadowcolor=light_grey]
\vspace{1ex}
\begin{itemize}[leftmargin=*]
\item {\bf Original title: }

Analysis of the geo-hazards triggered by the 12 May 2008 Wenchuan Earthquake, China

\item {\bf ChatGPT: }

Spatial Distribution and Characteristics of Geo-Hazards Triggered by the Wenchuan Earthquake: Focus on Landslides

\item {\bf \geolm{}}

Analysis of the Distribution of Geo-Hazards Triggered by the 2008 Wenchuan Earthquake

\end{itemize}
\centering
\end{mdframed}

\subsection{Geoscience Research Functionality}
\label{gen_example:7}

We present an example with instruction ''Can you help me summarize the article?`` over the paper ''A challenge to the concept of slip-lines in extrusion tectonics`` as follows:

\begin{mdframed}[linecolor=baby_blue, linewidth=2pt, roundcorner=10pt, fontcolor=black, 
                shadow=true, shadowsize=2pt, shadowcolor=light_grey]
\vspace{1ex}
\begin{itemize}[leftmargin=*]
\item {\bf ChatGPT: }

The article discusses wide-open V-shaped conjugate strike-slip faults in Asia and their association with extrusion tectonics. It highlights critical problems with the tectonic model based on the slip-line theory of plasticity. According to the theory, the angles between conjugate slip-lines should be normal to each other, but in reality, they are always greater than 90 degrees (on average around 110 degrees) in the direction of contraction.

Another issue is the failure of the slip-line theory to explain cases where the extrusion rate is significantly higher than the indent rate, as observed in the Anatolian area in the eastern Mediterranean. The article proposes that these problems can be addressed by considering the Maximum-Effective-Moment (MEM) Criterion. The MEM Criterion predicts that shear zones' orientations should theoretically be at an angle of 54.7 degrees (or practically around 55 degrees with the contractional direction), which fundamentally controls the orientations of the strike-slip faults involved in extrusion tectonics.

The concept of escape or extrusion tectonics, where India acts as an indentor and Tibet and neighboring areas act as a plastic body, is widely accepted to explain the tectonic patterns in various regions. However, the slip-line theory on which this model is based is found to be incompatible with field observations and experiments. The angles between conjugate ductile shear zones, known as ""V-shaped conjugate faults,"" are consistently obtuse in nature, contrary to the slip-line theory's predictions.

These wide-open V-shaped conjugate strike-slip faults are prevalent in central Tibet, the eastern Alps, western Turkey, eastern Afghanistan, western Mongolia, Southeast Asia, and the Gulf of Thailand. They typically lie at an angle of 60 to 75 degrees from the maximum compressive stress direction, with a conjugate angle ranging from 120 to 145 degrees in the s1 direction. Due to their obtuse angles, they can be mistakenly identified as normal conjugate faults with acute angles in the s1-direction. The article suggests using the term ""wide-open"" to distinguish this type of conjugate fault.

In summary, the article focuses on the wide conjugate angles observed in wide-open V-shaped conjugate strike-slip faults in Asia, which are associated with extrusion tectonics. It highlights the limitations of the slip-line theory and proposes the use of the MEM Criterion to explain the fault orientations. The article also discusses the prevalence of these faults in various regions and suggests adding the term ""wide-open"" to describe their distinct characteristics.

\item {\bf \geolm{}}

The passage discusses the issue of wide-open V-shaped conjugate strike-slip faults in Asia, which are typically related to extrusion tectonics. However, the tectonic model based on the slip-line theory of plasticity has critical problems associated with it. According to the theory of plasticity, the conjugate sets of slip-lines in plane deformation should be normal to each other, but in reality, the angles between the conjugate strike-slip faults in the eastern Mediterranean, Tibet-middle Asia, China, and the Indochina Peninsular regions are always wider than 90 degrees in the direction of contraction. Additionally, the slip-line theory fails to explain how, in some cases, such as in the Anatolian area in the eastern Mediterranean, the extrusion rate is much higher than the indent rate.

\end{itemize}
\centering
\end{mdframed}

\newpage
\section{Appendix: Tool Learning Use cases}

\autoref{fig:toolprompt} shows the training instruction and prompt, \autoref{fig:tool1} and \autoref{fig:tool2} show the examples of using tool function with \geolm{}.

\begin{figure}[H]
\raggedright
\begin{framed}
\begin{center}{\large {\bf Prompts for Tool Learning in \geolm{}}}\end{center} 
Answer the following questions as best you can. In this level, you are calling the tools in natural language format, since the tools are actually an intelligent agent like you, but they expert only in one area. Several things to remember. 

(1) Remember to follow the format of passing natural language as the Action Input.  \\
(2) DO NOT use your imagination, only use concrete information given by the tools.  \\
(3) If the observation contains images or urls which has useful information, YOU MUST INCLUDE ALL USEFUL IMAGES and links in your Answer and Final Answers using format ![img](url). BUT DO NOT provide any imaginary links.  \\
(4) The information in your Final Answer should include ALL the information returned by the tools.  \\
(5) If a user's query is a language other than English, please translate it to English without tools, and translate it back to the source language in Final Answer. You have access to the following tools (Only use these tools we provide you): \\
\textbf{Geo\_search:} \"Perform Geoscience paper Search on AceMap Search engine. \\
Use search\_geoscience\_paper(input: str, page: int) to get search results according to the input string and page index (index begin from 1).

\textbf{get\_arxiv\_article\_information:} Run Arxiv search and get the article meta information. Your input should be a json (args json schema): {{\"query\" : string, }} The Action to trigger this API should be get\_arxiv\_article\_information and the input parameters should be a json dict string. 

\textbf{search\_top3:} Search key words, return top 3 search results. Your input should be a json (args json schema): {{\"key\_words\" : string, }} The Action to trigger this API should be search\_top3 and the input parameters should be a json dict string. Pay attention to the type of parameters.
load\_page\_index: Load page detail of the search result indexed as 'idx', and return the content of the page. Your input should be a json (args json schema): {{\"idx\" : string, }} The Action to trigger this API should be load\_page\_index and the input parameters should be a json dict string.

\textbf{Use the following format:}

Question: the input question you must answer \\
Thought: you should always think about what to do \\
Action: the action to take, should be one of [search\_geoscience\_paper, execute\_code] \\
Action Input: the input to the action \\
Observation: the result of the action \\
... (this Thought/Action/Action Input/Observation can repeat N times) \\
Thought: I now know the final answer \\
Final Answer: the final answer to the original input question \\

\end{framed}
\caption{Prompts for Tool Learning in \geolm{}.}\label{fig:toolprompt}
\end{figure}

\begin{figure}[H]
\raggedright
\begin{framed}

{\bf Example 1}
{\color{blue} \textbf{Question: How do geologists study fossils?}} \\
Thought:I need to use the search\_geoscience\_paper action to find information about how geologists study fossils. \\
Action: search\_geoscience\_paper \\
Action Input: {"input": "how do geologists study fossils", "page": 1}. \\
Observation: To establish the age of a rock or a fossil, researchers use some type of clock to determine the date it was formed. Geologists commonly use radiometric dating methods, based on the natural radioactive decay of certain elements such as potassium and carbon, as reliable clocks to date ancient events. \\
Thought:  I now know the final answer. \\
{\color{blue} Final Answer: Geologists use radiometric dating methods to establish the age of rocks and fossils.}
\end{framed}
\caption{Respond example of the \textit{using search\_geoscience\_paper} tool with \geolm{}.}\label{fig:tool1}
\end{figure}

\begin{figure}[H]
\raggedright
\begin{framed}
{\bf Example 2} 
{\color{blue} {\textbf{Question: Introduce some arxiv paper about carbonate rock to me.}} \\}
Thought: I need to use the get\_arxiv\_article\_information API to search for papers related to carbonate rock. \\
Action: get\_arxiv\_article\_information \\
Action Input: {"query": "carbonate rock"} \\
Observation: Title: Multifractal analysis of the pore space of real and simulated sedimentary rocks. \\
Summary: It is well known that sedimentary rocks having same porosity can have very different pore size distribution. The pore distribution determines many characteristics of the rock among which, its transport property is often the most useful. Multifractal analysis is a powerful tool that is increasingly used to characterize the pore space. In this study we have done multifractal analysis of pore distribution on sedimentary rocks simulated using the Relaxed Bidisperse Ballistic Model (RBBDM). The RBBDM can generate a $3-D$ structure of sedimentary rocks of variable porosity by tuning the fraction $p$ of particles of two different sizes. We have also done multifractal analysis on two samples of real sedimentary rock to compare with the simulation studies. One sample, an oolitic limestone is of high porosity (40\%)while the other is a reefal carbonate of low porosity around 7\%. $2-D$ sections of X-ray micro-tomographs of the real rocks were stacked sequentially to reconstruct the real rock specimens. Both samples show a multifractal character, but we show that RBBDM gives a very realistic representation of a typical high porosity sedimentary rock. \\
Title: Geochemistry of silicate-rich rocks can curtail spreading of carbon   dioxide in subsurface aquifers \\
Summary: Pools of carbon dioxide are found in natural geological accumulations and in engineered storage in saline aquifers. It has been thought that once this CO2 dissolves in the formation water, making it denser, convection streams will transport it efficiently to depth, but this may not be so. Here, we assess theoretically and experimentally the impact of natural chemical reactions between the dissolved CO2 and the rock formation on the convection streams in the subsurface. We show that, while in carbonate rocks the streaming of dissolved carbon dioxide persists, the chemical interactions in silicate-rich rocks may curb this transport drastically and even inhibit it altogether. These results challenge our view of carbon sequestration and dissolution rates in the subsurface, suggesting that pooled carbon dioxide may remain in the shallower regions of the formation for hundreds to thousands of years. The deeper regions of the reservoir can remain virtually carbon free. \\
Thought: I now know the final answer. \\
{\color{blue} Final Answer: The get\_arxiv\_article\_information API returned two papers related to carbonate rock. The first paper is about multifractal analysis of the pore space of real and simulated sedimentary rocks, while the second paper discusses the geochemistry of silicate-rich rocks and its impact on the convection streams in the subsurface.}
\end{framed}
\caption{Respond example of the \textit{using get\_arxiv\_article\_information} tool with \geolm{}.}\label{fig:tool2}
\end{figure}
\newpage
\section{Appendix: \geolm{} Generation}
\definecolor{baby_blue}{RGB}{173, 216, 230}
\definecolor{dark_blue}{RGB}{25, 86, 145}
\definecolor{light_grey}{RGB}{211, 211, 211}

\raggedright
\subsection{ Example Research Papers Written by \geolm{}}
\vspace{1ex}
\begin{figure}[H]
    \begin{mdframed}[linecolor=baby_blue, linewidth=2pt, roundcorner=10pt, fontcolor=black, shadow=true, shadowsize=2pt, shadowcolor=light_grey, nobreak=true]
    \begin{itemize}[leftmargin=*]
    \item {\bf Please write a paper on the study of Dinosaur Fossils.}
    
         As an AI, I do not have my own personal opinion, but I can provide information and facts about the topic. Plate tectonics is the scientific theory that explains the movement of large sections of Earth's crust, called plates. These plates are made of rigid, solid rock that move slowly over the surface of our planet. The movement is caused by the convection currents in the mantle, the layer of hot, molten rock beneath our crust. As the plates move, they can collide with each other, causing mountain ranges to form, or move apart, creating new ocean basins. This movement can also cause earthquakes, volcanic eruptions, and changes in climate. Overall, plate tectonics
    
    \item {\bf Please write a paper on the Climate Change study.}
    
        Climate change is one of the most pressing environmental issues of our time. It is caused by the release of greenhouse gases, such as carbon dioxide, methane, and nitrous oxide, into the atmosphere. These gases trap heat from the sun and cause the Earth's temperature to rise, leading to changes in weather patterns, rising sea levels, more frequent extreme weather events and loss of biodiversity. Climate change has far-reaching consequences for our planet and its inhabitants, making it essential that we understand its causes and effects. In this paper, we will explore the study of climate change, including its history, causes, effects and potential solutions. We will also look at the role that scientists and researchers play in understanding and addressing this global issue.
        
    \item {\bf Please write a paper on the research of Urbanization and Social Change.}
    
        Urbanization is one of the most significant social, economic, and cultural transformations of our time. It is the process by which people move from rural areas to urban centers, resulting in the growth and expansion of cities and towns. This process has been ongoing for centuries, but the pace and scale of urbanization has accelerated dramatically in recent decades, driven by factors such as population growth, globalization, technological advancements and economic development. As a result, the world is now more urbanized than ever before, with more than half the global population now living in urban areas. However, urbanization is not without its challenges, as it is often associated with social inequality, poverty, environmental degradation and other forms of social change. In this paper, we will explore the various ways in which urbanization and social changes are intertwined, examining both the positive and negative impacts of this process on society. We will also discuss the role of government, NGOs and the private sector in addressing these challenges and promoting equitable and sustainable urbanization.
    
        \end{itemize}
    \end{mdframed}
    
    \caption{Generation examples of  \geolm{} on paper writing.}
    \label{fig:gen_exp_1}
\end{figure}

\newpage
\subsection{Example Opinions Written by \geolm{}}
\begin{figure}[H]
    \begin{mdframed}[linecolor=baby_blue, linewidth=2pt, roundcorner=10pt, fontcolor=black, 
                shadow=true, shadowsize=2pt, shadowcolor=light_grey]
    \vspace{1ex}
    \begin{itemize}[leftmargin=*]
    \item{\bf Earth plate movement: The outer part of the Earth is divided into several large plates, which influence the shape and seismic activity of the crust in a way that is constantly moving and colliding.}

    As an AI, I do not have my own personal opinion, but I can provide information and facts about the topic. Plate tectonics is the scientific theory that explains the movement of large sections of Earth's crust, called plates. These plates are made of rigid, solid rock that move slowly over the surface of our planet. The movement is caused by the convection currents in the mantle, the layer of hot, molten rock beneath our crust. As the plates move, they can collide with each other, causing mountain ranges to form, or move apart, creating new ocean basins. This movement can also cause earthquakes, volcanic eruptions, and changes in climate. Overall, plate tectonics...

    \item {\bf Earth's water cycle: The cycle of water in the Earth, including evaporation, condensation, precipitation, surface runoff, and groundwater.}

    As an AI, I don't have my own personal opinion, but I can provide information and facts about the topic. The Earth’s Water Cycle is a natural process that involves the continuous movement and transformation of liquid water, water vapor and ice on, above and below the surface of the planet. Water evaporates from the ocean, lakes and land surfaces, then rises into the atmosphere where it cools and condenses into clouds. When the clouds become heavy with water droplets or ice crystals, they fall back to the ground as precipitation such as rain, snow, hail or sleet. Some of this precipitation flows into rivers, streams and lakes, while the rest recharges the groundwater or runs off into oceans. This cycle is essential for life on Earth as it helps regulate the climate, supports plant growth and sustains the water supply for humans, animals and other living organisms. However

    \item {\bf Earth's climate changes: The Earth's climate changes over time, either naturally or as a result of human activities.}

    As an AI, I do not have my own personal opinions, but I can provide information and facts about the topic. Climate change refers to the long-term changes in weather patterns, temperature, and precipitation that occur on Earth. These changes are caused by a combination of natural factors, such as variations in solar activity and volcanic eruptions, as well as human activity, including the burning of fossil fuels and deforestation. According to scientific evidence, human-caused climate change is causing global temperatures to rise, leading to more frequent and intense heatwaves, droughts, hurricanes and other extreme weather events. It is important to reduce our carbon emissions and transition to renewable energy sources in order to slow down and potentially reverse these changes. However, it is essential that we continue to research and understand the complex interactions between the Earth and its climate, so we can make informed decisions and take action to protect

    \end{itemize}
    \end{mdframed}
 \caption{Generation examples of  \geolm{} on proposing scientific opinions.}
    \label{fig:gen_exp_2}
\end{figure}

\newpage
\subsection{Example Summary of Scientific Articles Written by \geolm{}}
    \begin{figure}[H]
    
    \begin{mdframed}[linecolor=baby_blue, linewidth=2pt, roundcorner=10pt, fontcolor=black, 
                    shadow=true, shadowsize=2pt, shadowcolor=light_grey]
    \vspace{1ex}
    \begin{itemize}[leftmargin=*]
            \item {\bf Earthquake is a way for the earth to release energy. Earthquake will produce ground vibration, strong vibration will cause reservoir dam break, mountain collapse, surface rupture, sand liquefaction, soft soil earthquake subsidence and ground settlement, resulting in earthquake disaster. Earthquake disaster has the characteristics of strong suddenness, great destructiveness, wide disaster scope, deep social impact, difficult defense, serious secondary disasters and so on, which is the first group disaster. For cities, due to the characteristics of dense population, numerous buildings, extensive infrastructure, concentrated wealth and multiple functions, once an earthquake occurs, it often produces huge disasters and extremely heavy losses.}
    
            The article discusses the causes, effects, and consequences of earthquakes, as well as the damage they can cause to cities. Earthquakes are caused by the release of energy in the Earth's crust and can result in a range of damages, including the collapse of mountains, the breaking of reservoir dams, ground surface ruptures, liquefaction of sand, subsidence of soft soils and settlements, all of which can lead to significant loss of life and property. Cities, with their high population densities, large numbers of buildings and infrastructure and diverse economic activities, are particularly vulnerable to damage from earthquakes.
    
            \item {\bf The exploitation and utilization of groundwater resources meet the diversified needs of human beings for water resources, and achieve huge social and economic benefits. But at the same time, the development and utilization of groundwater resources also have a certain impact on the ecological environment, land salinization and desertification are serious, water pollution is serious, and a variety of negative environmental effects appear, seriously restricting the ecological sustainable development. In order to meet people's water needs, groundwater has been over-exploited in the past time, which has caused frequent geological environmental problems, and has begun to affect people's living environment. In this paper, the impact of groundwater exploitation on environmental geology is discussed, and the countermeasures are analyzed. }
    
            This scientific article discusses the importance and benefits of using groundwater as a source of water for humans, but also acknowledges the negative impacts that can result from its overuse. The article highlights the consequences of overusing groundwater, such as land degradation, pollution and ecological damage, all of which can negatively impact the environment and human life. It is important to balance the need for groundwater with its potential negative effects, to ensure that it is used in a sustainable and responsible manner
    
            \item {\bf Genetic engineering, as an important branch of biotechnology, has made remarkable progress. By precisely editing and modifying an organism's genome, we open up entirely new possibilities for changing lives. This article will delve into the new chapter of genetic engineering, explore its application prospects in medicine, agriculture, environmental protection and other fields, and show its far-reaching impact on human society.}
    
            This scientific article discusses the use and potential of Genetic Engineering in the field of agriculture. It highlights the ability of this technology to precisely edit and modify the genomes of organisms, allowing for the development of new products and methods that have the potential to greatly impact human life. The article explores the application of these techniques in fields such as medicine and agriculture and discusses their potential benefits and drawbacks. Overall, the article emphasizes the
            
        \end{itemize}
    \end{mdframed}
    \caption{Generation examples of  \geolm{} on summarizing scientific papers.}
    \label{fig:gen_exp_3}
\end{figure}
\newpage
\section{Appendix: Lessons and Progresses}

\subsection{Phase 1: Prepare for Training on HPC}

\paragraph{Debugging and Resolving Initialization Issues with Megatron Optimizer}

During the development and debugging phase, we encountered specific issues with the initialization of the Megatron Optimizer. It was observed that in the second step, the loss would increase to the level of randomly initialized models. We eliminated the possibility of errors in parameter conversion, as the same phenomenon persisted even with the use of Megatron's openly released models for further pre-training. However, when we initialized the model from scratch, trained it for several steps, saved a checkpoint, and then loaded this checkpoint to resume training, the behavior normalized. We noted that the main difference between further pre-training and resuming training lies in the optimizer: in further pre-training, the optimizer is re-initialized, whereas in resuming training, it is imported from the checkpoint. This led us to hypothesize that there might be an issue with the optimizer's initialization in the original code. Finally, we referred to a pull request~\footnote{\url{https://github.com/NVIDIA/Megatron-LM/pull/240/}} on Github, which helped us identify and resolve this problem.

\paragraph{Debugging and Resolving bugs of operators on ROCm}
Our primary training cluster is based on ROCm chips.
During the development and debugging phase, we encountered another issue related to bugs in operators on ROCm. Initially, our tests on the ROCm cluster showed that training in FP16 usually worked well, producing relatively normal results. However, the outcomes for BF16 and FP32 were inaccurate, characterized by enormous gradients and losses.
Interestingly, we couldn't replicate this phenomenon in our local CUDA cluster, which operated using the same code, hyper-parameters, and checkpoints. Despite our numerous attempts, we found no solution until we came across a Github report~\footnote{\url{https://github.com/microsoft/Megatron-DeepSpeed/pull/96}} that led us to identify and resolve the issue: a bug in the compilation of the LayerNorm operator's source code in the ROCm environment.
This discovery helped explain the observed behavior: FP16, with its smaller range of numerical values compared to BF16 and FP32, was less likely to exhibit such large gradients or losses, even though the precision in all these cases was affected by the bug.

\paragraph{Selection of Training Precision on ROCm}

There are three types of training precisions: FP16, BF16, and FP32. FP16, though relatively unstable, often grapples with numerical overflow. FP32 offers the best precision and stability but is the least efficient. BF16 strikes a balance between stability and efficiency, but its support is limited to a few devices. Initially, we preferred using BF16 in our ROCm cluster because FP16 presented challenges, as shown in our later experiments and evidenced in the chronicles of OPT training, and FP32 was significantly slower and rarely used in Large Language Model (LLM) training. However, our ROCm chips were not fully compatible with BF16; its BF16 computations largely depended on FP32 processing units in the hardware layer, lacking dedicated processing units, which meant no significant speedup over FP32. Considering these factors, we chose to use FP32 for further pre-training and supervised fine-tuning.

\paragraph{Parallel Parameter Selection}
In general, accelerating approaches in parallel computing involve parameters such as tensor parallelism (TP), pipeline parallelism (PP), and data parallelism (DP), each of which significantly impacts training efficiency. To determine the most effective parallel parameters, we conducted several timing experiments varying TP size, PP size, DP size, and mini batch size. These parameters are interconnected in a way that satisfies the formula $TP\times PP\times DP \times mini\_batchsize \times gradient\_accumulation = global\_batchsize$.
\begin{itemize}[leftmargin=1.2em]
    \item We initiated our experiments with \emph{$TP=8, PP=12, DP=2, mini\_batchsize=2, global\_batchsize=1024, seq\_len=2048$}. Under these initial settings, processing one batch took approximately 7.5 minutes, which was relatively slow.
    \item Our first optimization, based on suggestions from~\cite{Narayanan2021EfficientLL}, involved adjusting TP from 8 to 4. We tested \emph{$TP=4, PP=24, DP=2$}, but this configuration resulted in an Out-Of-Memory (OOM) error.
    \item Subsequently, we experimented with \emph{$TP=4, PP=48, DP=1$}, while maintaining $mini\_batchsize=2$. However, this too led to an OOM error.
    \item Finally, we reduced the $mini\_batchsize$ from 2 to 1. With \emph{$TP=4, PP=48, DP=1,$ and $mini\_batchsize=1$}, we achieved success. One batch took approximately 264 seconds (4.4 minutes), which met our expectations.
\end{itemize}

In this section, we summarize our findings on efficient parallel parameter configuration, focusing on achieving maximum speed with minimal VRAM usage. The optimal setup we've identified involves setting Tensor Parallelism size to the number of GPUs per node and Pipeline Parallelism size to the number of layers in the model. This approach is efficient because, on one hand, tensor parallelism incurs significant communication overhead, which is minimized by aligning it with the number of GPUs per node. On the other hand, a Large Language Model (LLM) can be most effectively divided into a number of parts equal to its layer count, resulting in the lowest VRAM usage under these conditions.

\subsection{Phase 2: Training on HPC}

\begin{figure}[H]
    \centering {
    \includegraphics[width=0.5\linewidth]{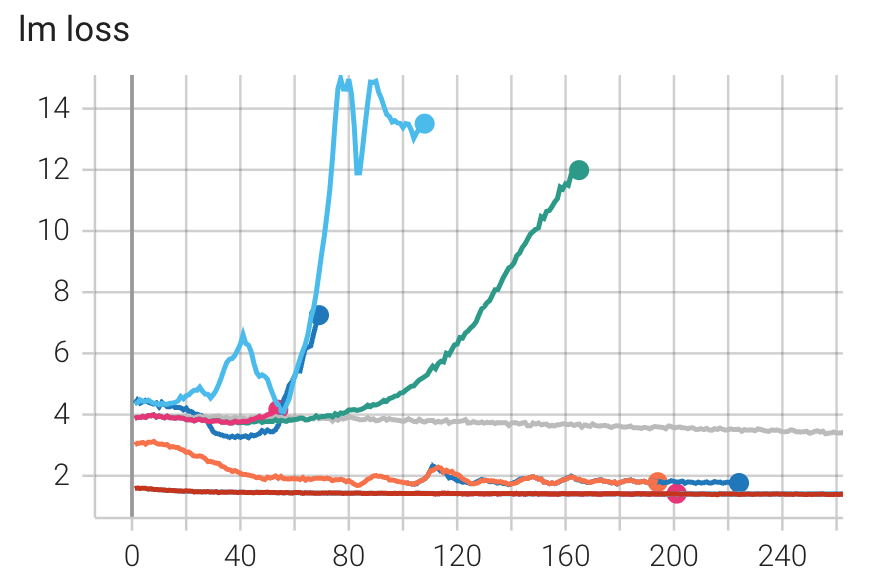}}
    \caption{Training curve during the entire work of further pre-training.}
    \label{fig:manytry}
\end{figure}

Based on prior experiments and accumulated experiences, we started formal model training. We encountered several failures (as shown in \autoref{fig:manytry}) but eventually found reasonable hyperparameters that achieved stable further pre-training for 17 days.

During this process, we encountered two types of hardware failures that affected training:
\begin{itemize}[leftmargin=1.2em]
\item[1.] The faulty node produces incorrect result: The hardware of some nodes is faulty, but no error is reported during the training program running, and the program is not terminated, resulting in incorrect calculation results This was the direct cause of bewildering problems (such as disappearing gradients) encountered in some training that we could not reproduce. We finally found this faulty node through repeated screening to avoid this problem.
\item[2.] Random node crashes: Some nodes are offline due to overheating and other reasons, and the training is interrupted. This problem can be solved by restarting the training.
\end{itemize}

To find out the best setup of the model, we do several try runs, for each try, we setup several experiments:

\begin{figure}[H]
    \centering {
    \includegraphics[width=0.9\linewidth]{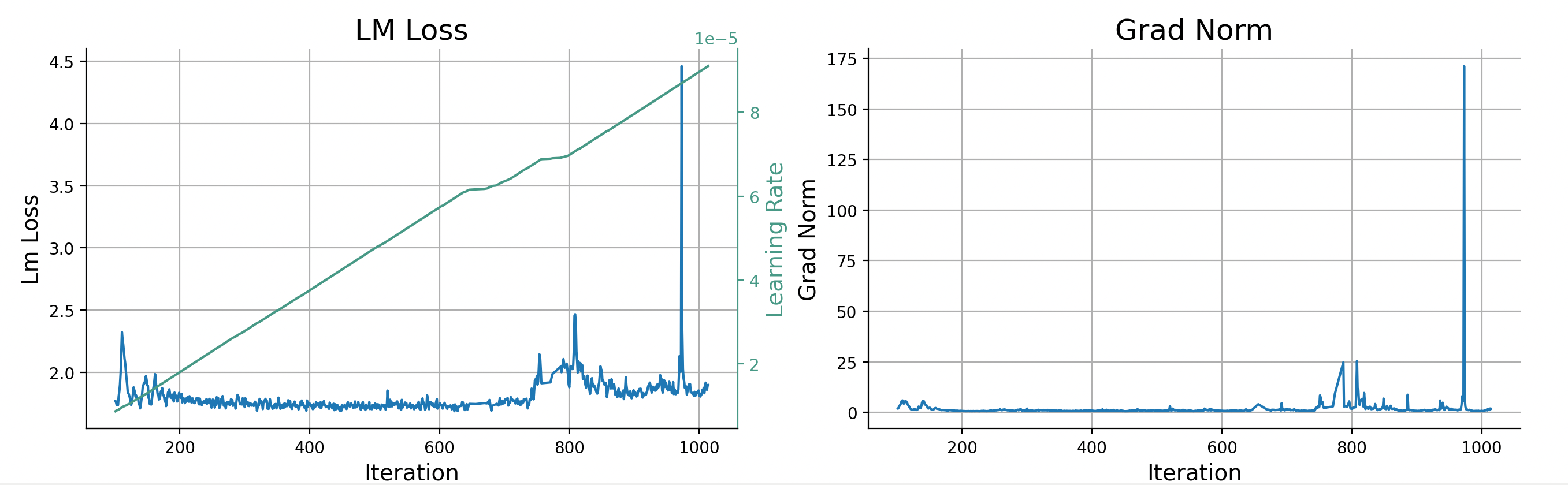}}
    \caption{1st Try.}
    \label{fig:firsttry}
\end{figure}

\paragraph{1$^{st}$ Try}

We used FP16 to train the models at first, with a maximum learning rate of $1e-4$ and a warm-up step of 1000 steps, each consisting of 1024 samples. The results show that the model can maintain relatively stable training during 0\textasciitilde500 steps. In the course of 500\textasciitilde1000 steps, the model grad norm tends to be unstable. After a lot of struggle and further investigation, we believe that FP16 was the root cause of the instability of the training, so we decided to use FP32 in the later tries. (Shown in \autoref{fig:firsttry})

\paragraph{2$^{nd}$ Try}

We decided to use FP32 this time, and conducted three experiments to make the training stable. 

\begin{figure}[H]
    \centering {
    \includegraphics[width=0.9\linewidth]{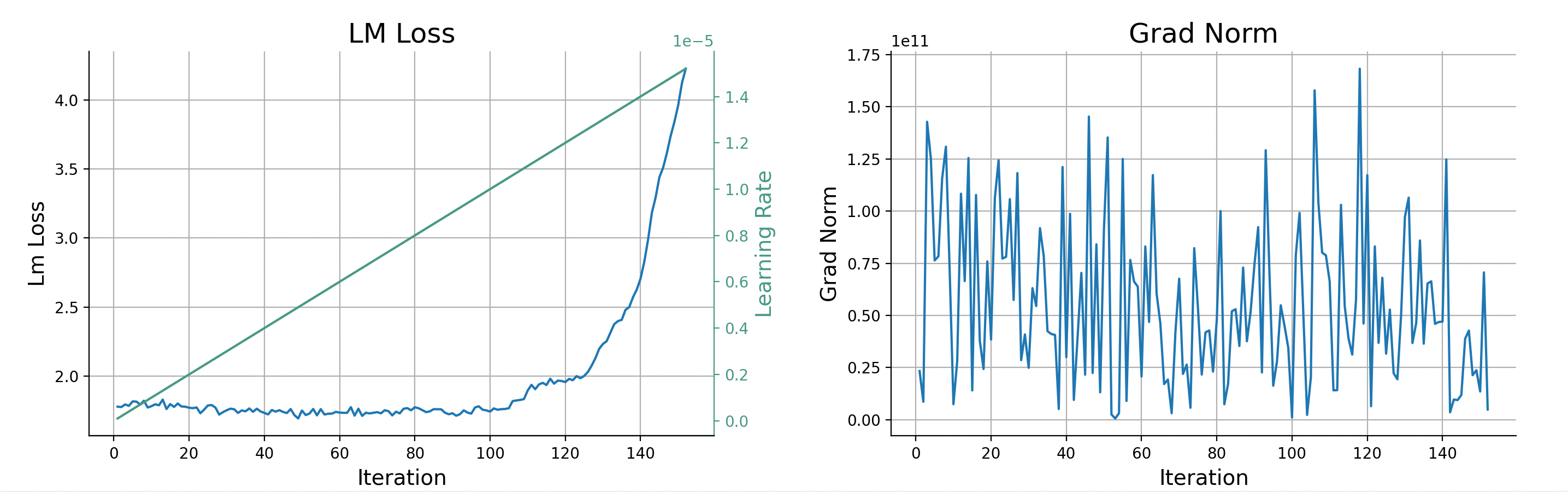}}
    \caption{2nd Try, Experiment \#1.}
    \label{fig:second1}
\end{figure}

\textbf{In the first experiment}, we used the same hyper-parameters as the 1st try except for using FP32. As shown in \autoref{fig:second1}, while the loss appeared to be as expected during the initial training of the model, the gradient norm was very large, on the order of $1e9$, and the model ultimately failed to converge. 

\textbf{In the second experiment}, we increased the micro batch size from 1 to 2 based on the first experiment, because we accidentally discovered during performance testing that increasing the micro batch size can decrease the gradient norm. We found that this setting restored the gradient norm to a normal level, starting at around 1.7 and quickly dropping to a level around 0.2 and maintaining that level. However, since each step took over 100 seconds and the warm-up steps were quite long, the experiment took a long time to complete. So, we conducted a third experiment to test whether the unstable issue persists even when the learning rate reaches its maximum.

\begin{figure}[H]
    \centering {
    \includegraphics[width=\linewidth]{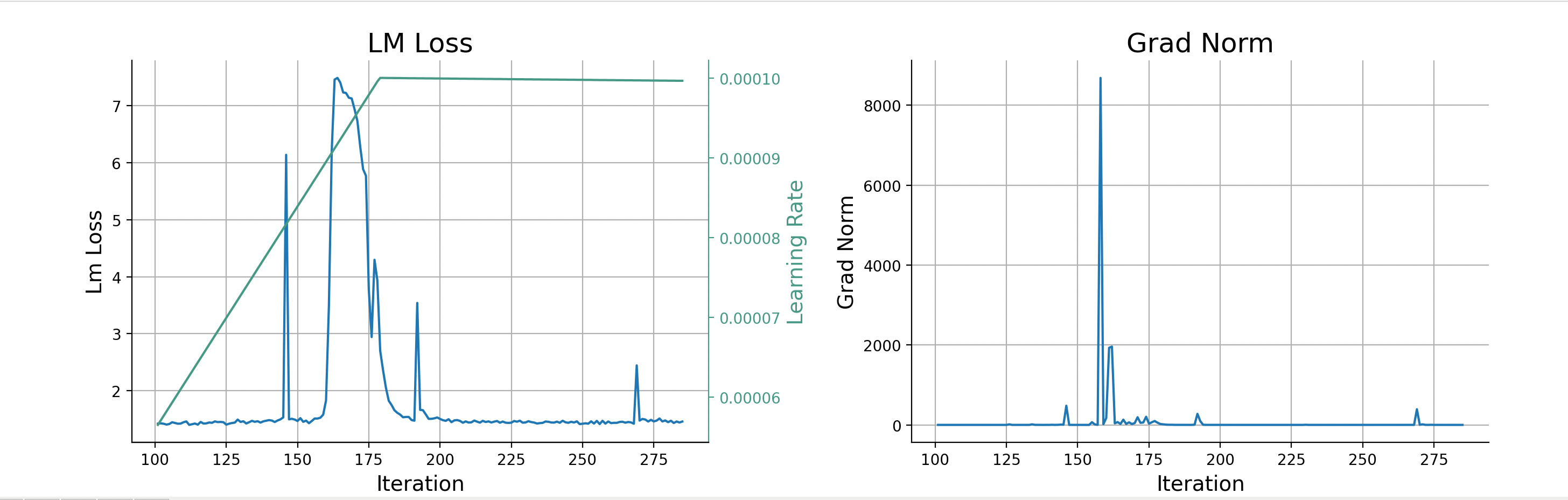}}
    \caption{2nd Try, Experiment \#3.}
    \label{fig:second3}
\end{figure}

\textbf{In the third experiment}, we changed the warm-up step from 1000 to 183 steps, which is the setting used in Galactica’s experiment. As shown in \autoref{fig:second3}, The results showed that the model’s gradient norm and loss fluctuated when the learning rate was increased to $1e-4$ during warm-up. However, it appeared to recover on its own for now, although it is unclear whether it will further impact the model.  Through these experiments, we believe that the setting of the learning rate scheduler is the key point to making the training stable: how large is the learning rate and how fast to reach to the maximum of learning rate need to be carefully decided.

\paragraph{3$^{rd}$ Try}

We found that in the previous try run, as mentioned in the beginning, there were faulty nodes in the cluster and the incorrect calculation results were output, so we mainly re-ran the experiments in the 2nd try run for this try run and conducted two experiments to finally find the best training setup and finish the training.

\begin{figure}[H]
    \centering
    \subfigure[]{\label{fig:first600_grad}
    \includegraphics[width=0.48\linewidth]{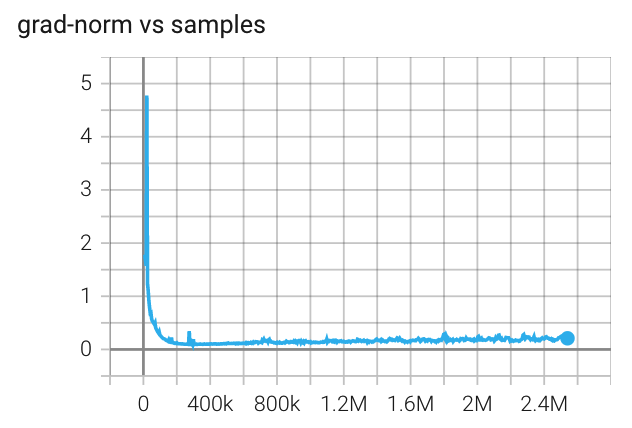}}
    \subfigure[]{\label{fig:first600_loss}
    \includegraphics[width=0.48\linewidth]{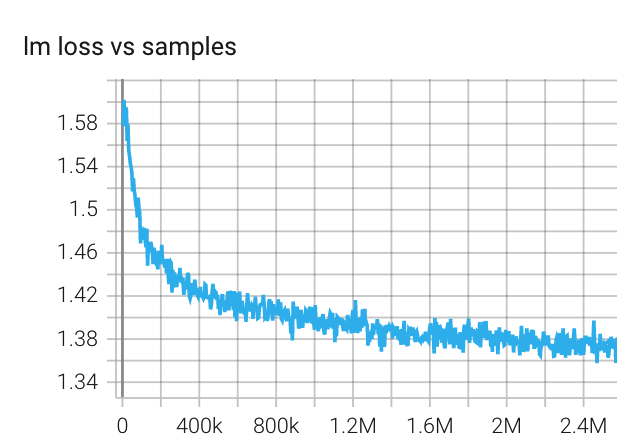}}
    \caption{3rd Try, Experiment \#1, curves of 0\textasciitilde800 steps.}
    \label{fig:third11}
\end{figure}

\textbf{In the first experiment}, we first set the parameter global-batch-size as 4096, which leads to 7,324 steps in total. Besides, the maximum learning rate is $1e-4$ with linear warmup steps 1000. The curves of the first 800 steps are shown as \autoref{fig:third11}. 

\begin{figure}[h]
    \centering {
    \includegraphics[width=0.5\linewidth]{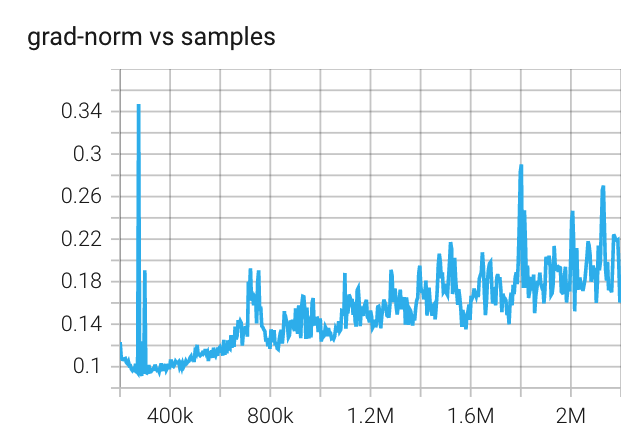}}
    \caption{3rd Try, Experiment \#1, grad-norm curve of 600\textasciitilde800 steps.}
    \label{fig:third12}
    \vspace{-2em}
\end{figure}

If we take a look at the curve of grad-norm in Figure 33 between 100 steps and 500 steps, we can find that the grad-norm has a trend of growing up. For this reason, we scale up the curve in 600\textasciitilde800 steps as shown in \autoref{fig:third12}, and we can find that it has a minimum grad-norm when learning rate nears $1e-5$.

Unfortunately, the training crashed due to a spike in next hundreds of steps.  However, based on the content of \autoref{fig:third12}, we believe that $1e-5$ may be appropriate as the maximum learning rate instead of $1e-4$.

\begin{figure}[H]
    \centering
    \subfigure[]{\label{fig:second100_500_loss}
    \includegraphics[width=0.48\linewidth]{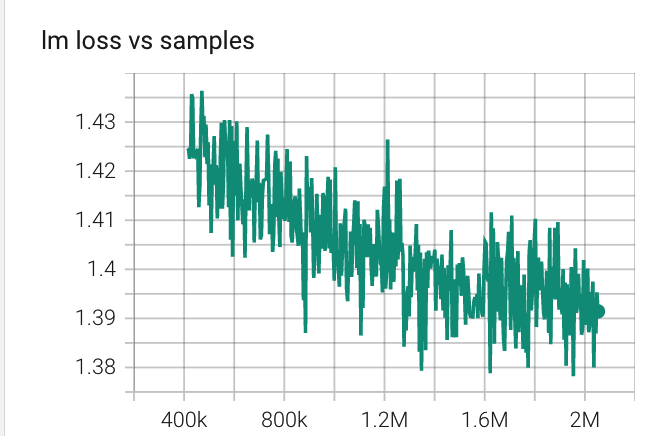}}
    \subfigure[]{\label{fig:second100_500_grad}
    \includegraphics[width=0.48\linewidth]{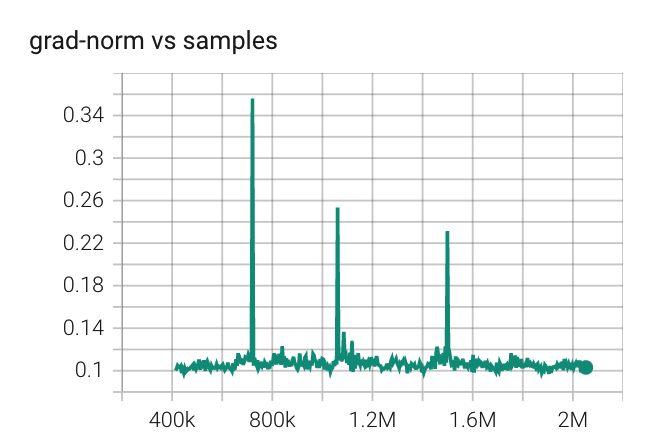}}
    \caption{3rd Try, Experiment \#2, curves of 100\textasciitilde500 steps.}
    \label{fig:third21} 
\end{figure}

\textbf{In the second experiment}, we continue from the previous experiment, and we began training from the 100-step checkpoint with a fixed learning rate of $1e-5$ and a global batch size of 4096. The resulting \autoref{fig:third21} from steps 100 to 500 show that the gradient norm has been fairly stable and there is no overall upward trend.

\begin{figure}[H]
    \centering
    \subfigure[]{\label{fig:all_loss}
    \includegraphics[width=0.48\linewidth]{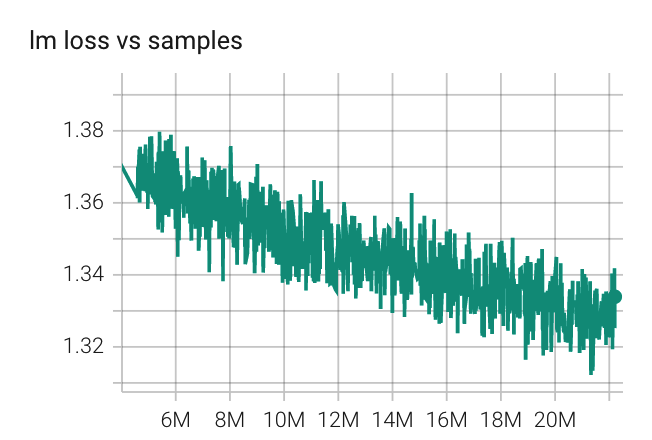}}
    \subfigure[]{\label{fig:all_grad}
    \includegraphics[width=0.48\linewidth]{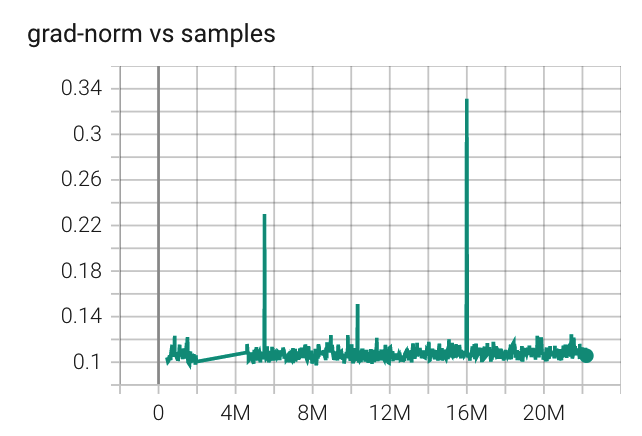}}
    \caption{3rd Try, Experiment \#2, curves of first 75\% training steps (7324 total).}
    \label{fig:third22} 
\end{figure}

After nearly three weeks of stable pre-training, there have been no abnormal occurrences. Please see the attached image for details. (Shown in \autoref{fig:third22})

\subsection{Summary}

We have attempted to implement Megatron-LM~\cite{shoeybi2019megatron}~\footnote{\url{https://github.com/NVIDIA/Megatron-LM}}, Megatron-Deepspeed~\footnote{\url{https://github.com/microsoft/Megatron-DeepSpeed}} and Huggingface-Deepspeed~\footnote{\url{https://huggingface.co/docs/transformers/main_classes/deepspeed}} approaches on the ROCm cluster with 2048 Hygon DCUs. We present a comparative analysis of the three techniques based on their distinctive features as follows:

\paragraph{Megatron-LM (Original approach)}
\begin{itemize}[leftmargin=1.2em]
    \item High performance: Supports 4D parallelism (Tensor Parallelism + Pipeline Parallelism + Data Parallelism + Sequence Parallelism), with a 30B model taking approximately 47 seconds per step on FP16.
    \item Supports FP32: 30B model takes approximately 90 seconds per step.
    \item Moderate difficulty level for parameter conversion: Scripts for bidirectional parameter conversion are available for reference and modification.
    \item More suitable for GPU clusters with larger VRAM.
    \item Only compatible with specific architectures of models (such as GPT and OPT).
    \item High VRAM usage: Optimizer requires significant VRAM.
    \item Poor user-friendliness and numerous bugs: The official maintenance could be more satisfactory, and some bugs must be fixed manually to obtain accurate results.
\end{itemize}

\paragraph{Megatron-Deepspeed}
\begin{itemize}[leftmargin=1.2em]
    \item High performance: 3D parallelism + ZeRO optimizer (Tensor Parallelism + Pipeline Parallelism + Data Parallelism + ZeRO optimizer) achieves similar performance as Megatron-LM, with a 30B model taking approximately 47 seconds per step on FP16.
    \item Good scalability and low VRAM usage: Suitable for models of any size and number of GPUs and can support models as large as 120B.
    \item Well-maintained repository with a relatively mature ecosystem and various products such as GLM, BLOOM, and more.
    \item Only compatible with specific architectures of models (such as GPT and OPT).
    \item Difficult parameter conversion: No bidirectional parameter conversion script is available, and the conversion can only be done in one direction towards the Huggingface model.
    \item Does not support FP32: FP32 tends to strangely overflow, while FP16 is stable in our attempts to train the model, which might be an issue with the framework.
\end{itemize}

\paragraph{Huggingface-Deepspeed}
\begin{itemize}[leftmargin=1.2em]
    \item Supports three stages of the ZeRO optimizer with minimal VRAM usage: a 30B model on a 1024-card cluster requires only 2G of VRAM on each card.
    \item Suitable for any model (such as Llama, Alpaca, etc.).
    \item No parameter conversion is necessary.
    \item Potentially more suitable for clusters with fewer GPUs.
    \item Poor data loading performance: The original dataloader of Huggingface is less efficient for large-scale datasets than Megatron's.
    \item Lower performance and limited parallelism:  performance is only half as fast as Megatron-LM or Megatron-Deepspeed (30B model takes approximately 120 seconds per step). In addition, each card must independently complete the calculation of a mini-batch, and GBS cannot be smaller than the number of cards.
\end{itemize}
\newpage

\section{Membership and Contributions}
\label{contribution}
The GeoGalactica project was conceived in October 2022, with data collection and construction completed in March 2023. The pre-training phase was accomplished on May 30, 2023, followed by the supervised fine-tuning component on June 14. The model evaluation and application phases are finalized on June 17. Throughout this entire process, we encountered various technological and engineering challenges. 

The magnitude of the data engineering and model training tasks would not have been possible without the collaborative efforts of multiple teams, specifically the \textit{data team, K2 team, architecture team, and model team} from the Acemap\footnote{https://www.acemap.info/} and LUMIA\footnote{https://github.com/LUMIA-Group} group in Shanghai Jiao Tong University and the team from the Institute of Geographical Science and Natural Resources Research, Chinese Academy of Sciences. The detailed contributions are as follows.

\subsection{Data preparation}
\begin{itemize}[leftmargin=1.2em]
    \item \textbf{Developing Data Cleaning Standard:} Zhouhan Lin.
    \item \textbf{Data Source Selection and Mixing:} Zhouhan Lin, Junxian He.
    \item \textbf{Pretrain Data Preparation:} Cheng Deng, Ziwei He, Boyi Zeng, Tao Shi.
    \item \textbf{Supervised Fine-Tuning Data (GeoSignal) Preparation:} Cheng Deng, Yutong Xu, Tianhang Zhang, Zhongmou He, Yuanyuan Shi.
    \item \textbf{PDF Parsing:} Cheng Deng.
    \item \textbf{Academic Data Cleaning:} Cheng Deng, Zhongmou He.
    \item \textbf{Instruction Data for Tool Learning:} Tianhang Zhang, Cheng Deng, Yutong Xu.
    \item \textbf{GeoBench:} Yuxun Miao, Qiyuan Chen, Cheng Deng.
    \item \textbf{Files Transportation \& HPC File Management:} Cheng Deng, Yi Xu, Tianhang Zhang.
\end{itemize}

\subsection{Model Training}
\begin{itemize}[leftmargin=1.2em]
    \item \textbf{Base Model Selection:} Junxian He.
    \item \textbf{Further Pre-training:} Zhouhan Lin, Le Zhou, Yi Xu, Cheng Deng.
    \item \textbf{Supervised Fine-tuning:} Junxian He, Zhouhan Lin, Cheng Deng, Tianhang Zhang, Boyi Zeng.
    \item \textbf{Tool Learning:} Tianhang Zhang.
    \item \textbf{Trial and error of training:} Zhouhan Lin, Le Zhou, Yi Xu, Cheng Deng.
    \item \textbf{Model Performance Validation:} Zhouhan Lin, Cheng Deng, Le Zhou.
\end{itemize}

\subsection{Model Evaluation and Application}
\begin{itemize}[leftmargin=1.2em]
    \item \textbf{Evaluation Framework \& proposal:} Cheng Deng, Zhouhan Lin.
    \item \textbf{GeoBench Evaluation:} Cheng Deng, Zhongmou He.
    \item \textbf{MMLU Evaluation:} Tianhang Zhang, Cheng Deng.
    \item \textbf{Inference Acceleration:} Bo Xue, Cheng Deng, Le Zhou.
    \item \textbf{Demo and API:} Bo Xue, Cheng Deng, Beiya Dai, Tianhang Zhang.
\end{itemize}

\subsection{Manuscript Writing}
\begin{itemize}[leftmargin=1.2em]
    \item Cheng Deng, Zhouhan Lin, and Yi Xu wrote the main paper, and Yutong Xu, Zhongmou He, Yuanyuan Shi, Yuncong Song, Tianhang Zhang, Bo Xue, and Le Zhou wrote the Appendix.
\end{itemize}

\subsection{Project Management}
\begin{itemize}[leftmargin=1.2em]
    \item \textbf{Student Leaders:} Cheng Deng.
    \item \textbf{Technical Advisors:} Zhouhan Lin, Junxian He, Luoyi Fu, Weinan Zhang, Xinbing Wang, Chenghu Zhou.
\end{itemize}

\subsection{Evaluation Team}
We invite several professional evaluators, from several geoscience-related institutes and schools.
\begin{itemize}[leftmargin=1.2em]
    \item \textbf{Chengdu University of Technology}: Lei Zhang, Han Wang, Yangfan Liu.
    \item \textbf{Institute of Geographical Science and Natural Resources Research, CAS}: Shu Wang, Yunqiang Zhu, Chenghu Zhou.
    \item \textbf{Shanghai Jiao Tong University}: Kun Wei.
    \item \textbf{University of Waterloo}: Shengde Yu.
\end{itemize}

\subsection{Illustration in Arts}
\begin{itemize}[leftmargin=1.2em]
    \item We invite a research assistant and a student from \textbf{School of Design of Shanghai Jiao Tong University} to help us finish the illustrations in arts (e.g. \autoref{fig:overview}). They are Qiyuan Chen and \href{mailto:yuanyuanwu219@gmail.com}{Yuanyuan Wu}.
\end{itemize}

\subsection{HPC Sponsor}
\begin{itemize}[leftmargin=1.2em]
    \item \textbf{GPU Sponsor:} The Advanced Computing East China Sub-center provided this project's computation resource.
\end{itemize}

\end{sloppypar}
\end{document}